\documentclass[conference]{IEEEtran}
\IEEEoverridecommandlockouts
\usepackage{cite}
\usepackage{amsmath,amssymb,amsfonts}
\usepackage{algorithmic}
\usepackage{graphicx}
\usepackage{textcomp}
\usepackage{xcolor}
\def\BibTeX{{\rm B\kern-.05em{\sc i\kern-.025em b}\kern-.08em
		T\kern-.1667em\lower.7ex\hbox{E}\kern-.125emX}}

\usepackage[ruled,vlined,linesnumbered]{algorithm2e}
\SetKwRepeat{Do}{do}{while}
\usepackage{subfig}
\usepackage[nottoc]{tocbibind}

\usepackage[normalem]{ulem}

\usepackage{listings}

\begin{document}

\title{One-Class Model for Fabric Defect Detection\\
}

\author{\IEEEauthorblockN{Hao Zhou, Yixin Chen, David Troendle, Byunghyun Jang}
\textit{Computer and Information Science} \\
\textit{University of Mississippi}\\
University, MS USA \\ hzhou3@go.olemiss.edu, \{ychen, david, bjang\}@cs.olemiss.edu}

\maketitle

\begin{abstract}
An automated and accurate fabric defect inspection system is in high demand as a replacement for slow, inconsistent, error-prone, and expensive human operators in the textile industry. Previous efforts focused on certain types of fabrics or defects, which is not an ideal solution. In this paper, we propose a novel one-class model that is capable of detecting various defects on different fabric types. Our model takes advantage of a well-designed Gabor filter bank to analyze fabric texture. We then leverage an advanced deep learning algorithm, autoencoder, to learn general feature representations from the outputs of the Gabor filter bank. Lastly, we develop a nearest neighbor density estimator to locate potential defects and draw them on the fabric images. We demonstrate the effectiveness and robustness of the proposed model by testing it on various types of fabrics such as plain, patterned, and rotated fabrics. Our model also achieves a true positive rate (a.k.a recall) value of 0.895 with no false alarms on our dataset based upon the Standard Fabric Defect Glossary.

\end{abstract}


\section{Introduction\label{sec:intro}}

Fabric inspection plays a critical quality control role in the textile industry. Trained human inspectors conduct quality inspections to find any potential fabric defects. However, due to the inherent limitations of human labor such as eye fatigue and distraction, the process is considered time-consuming, inconsistent, error-prone, and expensive. Thus, in order to improve this important process, an automated and accurate inspection system is highly desired. However, developing such an automated inspection system is technically challenging mainly due to the following two challenges.

\textbf{\textit{Challenge 1}}: Defects in fabrics vary in their size and shape. Further, fabric defects are usually caused by machine malfunctions such as weaving and painting issues, and they are relatively rare in practice, implying that the number of error-free fabrics (referring as non-defective, positive, or normal) is significantly larger than the number of defective fabrics (referring as defective or negative).

To deal with many different types of defects, researchers have built detection systems based on multi-class classification~\cite{zhang2018yarn}\cite{zhou2020exploring}. However, such systems do not work well beyond the pre-defined classes and suffer from inherent unbalanced dataset issues (e.g., the number of instances of some defects is larger than that of other defects).

\textbf{\textit{Challenge 2}}: In addition to the variations in defects, background textures also vary in their patterns, based on different weaving methods and painting patterns (e.g., plain, patterned, etc). They are categorized into over 70 kinds~\cite{ngan2011automated}.
Building a system that analyzes all kinds of fabrics is a challenging task. 

Some researchers took advantage of Gabor filters to analyze specific types of fabrics with optimized filter parameters~\cite{tong2016differential}\cite{mak2012fabric}. 
However, different optimization approaches and objectives yield different results, making it difficult to apply in the real world. More importantly, an optimized Gabor filter only works for one or a few types of fabrics.
In other words, covering a wide range of fabrics requires an optimized Gabor filter bank.


To address the aforementioned challenges, this paper proposes a novel one-class model for fabric inspection. The model is based on one-class classification (OCC) and Gabor filters. For challenge 1, we leverage one-class classification, which is helpful when negative samples are absent or not well-defined. In OCC, an efficient classifier is able to define a classification boundary with only the knowledge of the positive class~\cite{khan_madden_2014}.
Therefore, OCC simplifies the problem to focusing only on positive fabrics. This eliminates the need to collect, deal with various fabric defects and worry about imbalanced datasets or lack of negative samples.
For challenge 2, we carefully design a Gabor filter bank with different orientations and bandwidths to analyze different fabric textures. By doing so, we ease the development of an optimized Gabor filter design for each fabric texture. In order to define a good classification boundary, we design an autoencoder to maximize representative features from the output of the Gabor filter bank. Finally, we leverage a nearest neighbor density estimator to locate defects and draw detective areas on fabric images.

To validate our proposed one-class model, we extensively test the model from different aspects. We focus on the effectiveness of the model on plain, patterned, and rotated fabric images, and also the advantages of the autoencoder as a feature selector compared against a hand-crafted method, a Principal Component Analysis (PCA) method, and another advanced deep learning algorithm. We also investigate the model's performance when parameters of our model change. The experiments show that our proposed model effectively detects defects on fabric images with various background textures and achieves a true positive rate value of 0.895 with no false alarms on a selected dataset from the Standard Fabric Defect Glossary~\cite{StandardFabricDefectGlossary}.

The rest of the paper is organized as follows. Section~\ref{sec:relatedwork} summarizes efforts on the problem, Second~\ref{sec:design} details the overall design of our model, Section~\ref{sec:evaluation} shows our experiments from different aspects, and Section~\ref{sec:conc} concludes our paper.

\section{Related Works\label{sec:relatedwork}}
\begin{figure*}
	\centering
	\includegraphics[width=\textwidth, keepaspectratio]{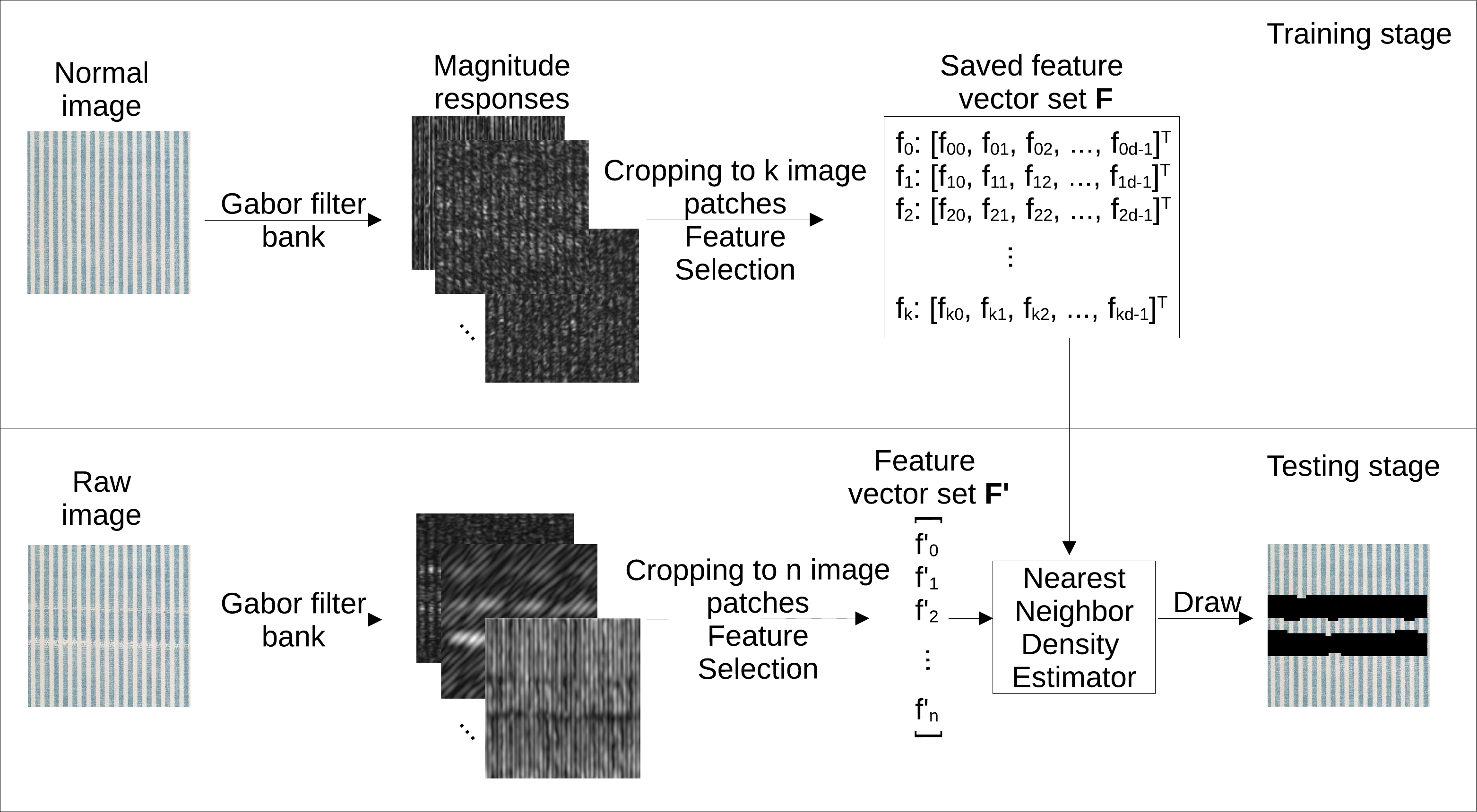} 
	\caption{Overall model design.}
	\label{fig:overall}
\end{figure*}
Researchers have proposed numerous systems and algorithms to automate the fabric defect detection problem. These methods can be generally classified into four types of approaches - statistical, structural, model-based, and spectral. We briefly introduce these approaches with more focus on relevant spectral approaches. We also introduce several advanced deep learning algorithms for the problem toward the end of this section.

Statistical approaches~\cite{chetverikov2002finding}~\cite{latif2000efficient} usually leverage first- (e.g., mean and standard deviation) or second-order statistics (e.g., correlation method). These statistics are used to represent the color information of fabrics. However, statistical information extracted only from raw images (color or grey-level images) is not enough to fully represent fabric features, not being able to distinguish between fabric features and defects, especially in patterned cases. Secondly, the structural approach~\cite{chetverikov2000pattern} works well only on plain fabrics as the patterns of plain fabrics are easier to analyze and retrieve. When encountering complex patterns, the structural approach is not a viable solution. Thirdly, model-based approaches (e.g., Markov random field) usually explore the relation among pixels of fabric images. As with the statistical approaches, model-based approaches tend to ignore small defects, which degrades the applicability of the approach. Lastly, spectral approaches based on Fourier, wavelet, or Gabor transforms are superior to other approaches as they can extract texture features by analyzing fabrics in the frequency domain, which in return, are less sensitive to noise compared to ones in the spatial domain (e.g., statistical approach). 

Gabor filters have gained popularity for texture analysis due to their ability to model the human visual cortex cells. In the literature, Gabor filter-based approaches for fabric defect detection are based either on optimized Gabor filters or on a set of Gabor filters called a Gabor filter bank. Tong \textit{et al.}~\cite{tong2016differential} optimize a few Gabor filters by utilizing composite differential evolution. In return, the algorithm can successfully segment the defects from the fabric images. Mak \textit{et al.}~\cite{mak2012fabric} also optimize Gabor filters by a genetic algorithm. It should be noted, however, optimized filters work only on certain types of fabrics.
Kumar \textit{et al.}~\cite{kumar2002defect} propose a Gabor filter bank with different scales and orientations. The Gabor filter bank covers multi-resolution and can detect the features of defects by fusing outputs of Gabor filters. Even though a Gabor filter bank deals with different types of fabrics, simply fusing the outputs of filters may intensify noise in normal fabrics since each filter extracts features from one pair of scales and orientations. Unfortunately, the outputs of a Gabor filter bank are significantly bigger (e.g., the dimension of feature vectors is large). Bissi \textit{et al.}~\cite{bissi2013automated} perform Principal Component Analysis (PCA) on the outputs of a Gabor filter bank which greatly reduces the dimension of feature vectors. However, during the process, information loss can occur if the number of principal components is not selected carefully.

Researchers have also tried to solve the fabric defect detection problem by utilizing advanced deep learning algorithms. Some treat the problem as an object detection problem. Zhou \textit{et al.}~\cite{zhou2020exploring} build a system by incorporating several techniques to the vanilla Faster RCNN to locate and classify defects. Zhang \textit{et al.}~\cite{zhang2018yarn} build a detection system based on YOLO by extensively comparing among different YOLO frameworks. These algorithms are supervised, which implies that in order to obtain a well-trained model, a balanced and large dataset is a must. However, this is difficult to achieve in practice. Moreover, defect variations in terms of size and shape, and texture background make these algorithms vulnerable.

\section{Design\label{sec:design}}

In this section, we overview the design of our proposed model for the fabric defect detection problem. As shown in Fig.~\ref{fig:overall}, the model has two stages - training and testing stages. The model in the training stage takes normal images (i.e., images without defective areas) as input, and applies a Gabor filter bank with different orientations and bandwidths. The filtered images then are cropped into small image patches (e.g., $k$ image patches) and a technique of feature selection is applied on these images patches. Each feature vector is D-dimension and the generated feature vectors are saved for the testing stage. In the testing stage, the model takes a raw image (either a normal or a defective image) as input, applies the same Gabor filter bank to the image, and crops the filtered images into image patches. After the same feature selection, a density estimator takes the saved feature vector and the new feature vector to determine an image patch is either normal or defective. Lastly, the determined defective image patches are painted on the raw image. In the rest of this section, we detail our Gabor filter bank, the feature section technique, and our density estimator.

\subsection{Gabor Filter Bank\label{bank}}

Gabor filter is a good model of simple cells in the human visual cortex and is capable of representing textures very well~\cite{jones1987evaluation}. A Gabor filter is the product of a sinusoid and a Gaussian:

\[
g(x,y;\lambda,\theta,\phi,\gamma) = \exp\bigg(-\frac{x'^2+\gamma^2 y'^2}{2\sigma^2}\bigg) \cos\bigg(2\pi\frac{x'}{\lambda} + \phi\bigg),
\]
where
$\lambda$ is the wavelength of the Gabor filter (e.g., the number of cycles/pixel), $\theta$ is the orientation of the Gabor filter (e.g., the angle of the normal to the sinusoid), $\phi$ is the phase (e.g., the offset of the sinusoid), $\gamma$ represents aspect ratio, $\sigma$ is the spatial envelope of the Gaussian and is controlled by the bandwidth (e.g., $\sigma$ = 0.56$\lambda$), \(x' = x\cos\theta + y\sin\theta\), and \(y' = -x\sin\theta + y\cos\theta.\)

In the fabric defect detection problem, the wavelength $\lambda$ and orientation $\theta$ are more important as fabrics can be orientated differently and the width of yarns varies. So, the design challenge is to find a good pair of $\lambda$ and $\theta$. However, due to the varieties of fabrics, it is not possible to find a perfect pair of $\lambda$ and $\theta$ that works well for all types of fabrics. Recently, researchers tried to optimize Gabor filters for each type of fabrics~\cite{tong2016differential}~\cite{li2019fabric}, but it is unrealistic when considering a real-world situation where a type of unseen fabric continuously presents. Without losing generality, we propose a Gabor filter bank with different wavelengths and bandwidths that can cover most fabrics. We have a set of wavelengths \{$\lambda$ : $\lambda$ $=$ $2m$ where $2$ $\leq$ $m$ $\leq$ $6$\} and a set of orientations \{$\theta$ : $\theta$ $=$ $10n$, where $0$ $\leq$ $n$ $\leq$ $18$\}. Thus, our Gabor filter bank $G$ consists of a total of $90$ Gabor filters (e.g., $g_{0}$, $g_{1}$, ..., $g_{89}$).
When filtering the input image $I$ with the Gabor filter bank $G$, the magnitude responses $G'$ can be represented as the following.
\[
\{ G':G'_{i} = I \circledast g_{i} \},
\]
where $0$ $\leq$ $i$ $\leq$ $89$. Note that the filtered images $G'$ have the same size as the image $I$ (512$\times$512 in our dataset).

\subsection{Feature Selection\label{feature}}
Magnitude responses $G'$ are the features filtered by the Gabor filter bank and their sizes are large (e.g., 512$\times$512) since we propose a Gabor filter bank to increase the generality. It is possible that some filters have the same or similar effects when filtering the input image. In this case, the magnitude responses would present redundant features, which decreases the performance and increases testing time in our model. So in order to speed up the model, especially the density estimation process (e.g., determining defective image patches), more general and representative features are needed. To this end, we propose an autoencoder feature learning neural networks method, inspired by the development of deep learning algorithms~\cite{mei2018automatic}.



\begin{figure}
	\centering
	\includegraphics[width=0.8\columnwidth, keepaspectratio]{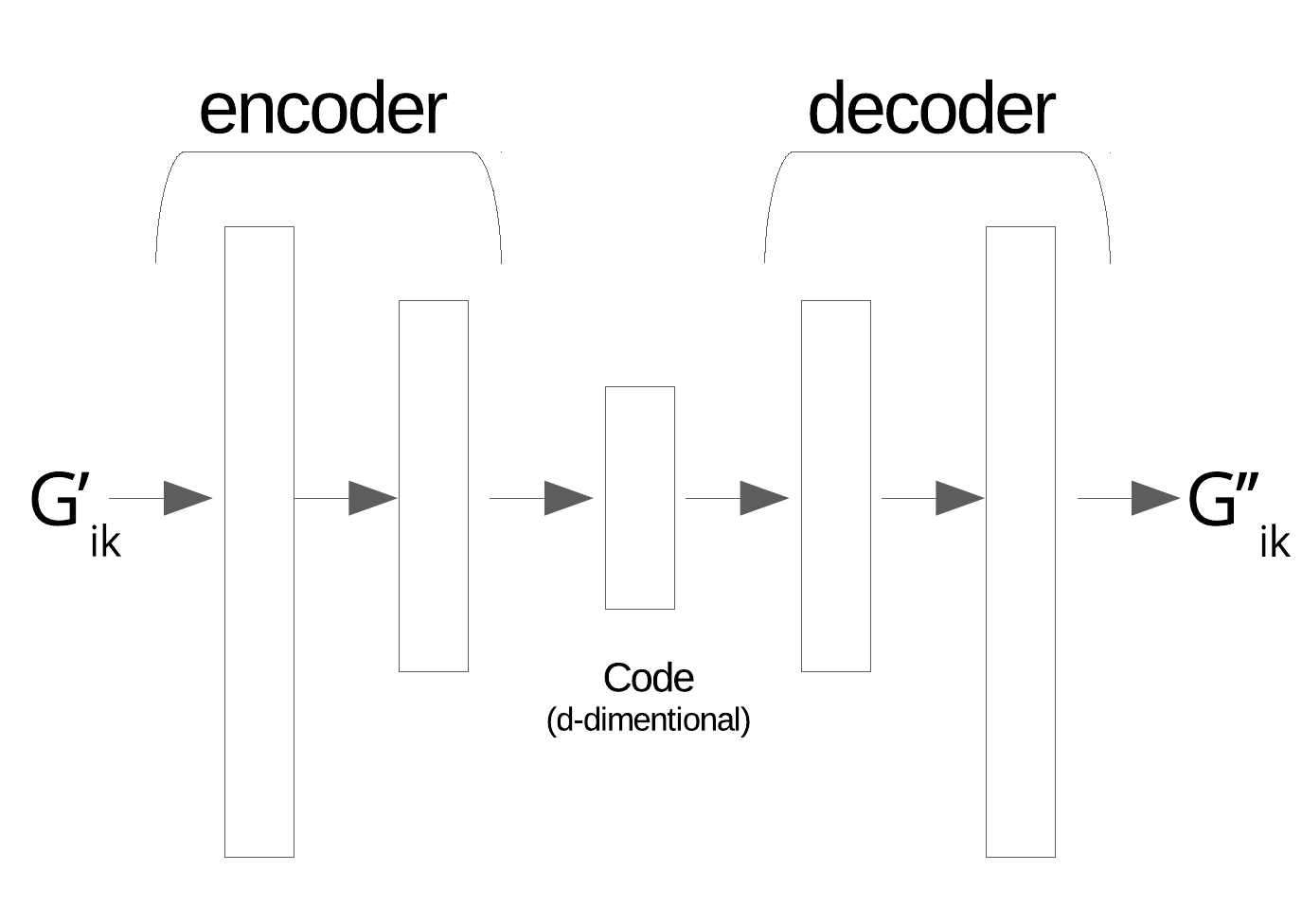} 
	\caption{Structure of autoencoder.}
	\label{fig:encoder}
\end{figure}

In our model, instead of manually determining what statistical data we use or what features are more general, we rely on autoencoder neural networks to find the more representative features from the magnitude responses.
We found that an autoencoder with a simple structure is capable of finding the representative features from the magnitude responses. The structure is shown in Fig.~\ref{fig:encoder}. By trying to reconstruct $G'_{ik}$ and minimizing the distance between $G'_{ik}$ and $G''_{ik}$ over and over again, the autoencoder neural network is able to learn the best code to represent the input response. Note that we train the autoencoder neural network only with the responses from normal fabric images. After training, whenever input is fed to the autoencoder, the code is the representative feature vector of the input.

One may question why we don't apply autoencoders on fabric images directly, or why we don't take advantage of transfer learning~\cite{torrey2010transfer}. One reason we don't directly train autoencoders on fabric images is that we don't have a large and balanced dataset. 
If we train an autoencoder directly on fabric images, the convergence of our autoencoder is an issue. Transfer learning might be a good solution since the previous study~\cite{krizhevsky2012imagenet} shows that the weights of the first layer in convolutional neural networks are indeed quite similar to Gabor filters. However, one reason we don't transfer weights from well-trained models is that the choice of models is limited and some requirements (e.g., input size, number of layers, etc.) must be met. Moreover, to our best knowledge, there are no pre-trained models for the problem of fabric defect detection so that transferring weights from quite different datasets might hurt the performance of our design as well.

\subsection{Nearest Neighbor Density Estimator\label{density}}
By carefully observing fabric images, one can easily find that most defects are different from the fabric's background. This provides us a good opportunity to classify fabric image patches into either defective or normal ones. To this end, we propose a Nearest Neighbor Density Estimator (NNDE). NNDE is able to identify defective fabric patches at the testing stage based on the saved feature vectors from the training stage as in Fig.~\ref{fig:overall}. Given a feature vector $f'_{i}$ (note that a feature vector represents an image patch) and the saved feature vector set $F$, the estimator can be mathematically defined as the following.


\[
s_{i} = \dfrac{Dist(f'_{i}, F)}{Dist(f_{j}, F)}, f'_{i} \in F^\prime, f_{j} \in F,
\]
where $Dist(\cdot)$ returns the distance between the feature vector and its nearest neighbor in $F$ and $s_{i}$ is proportional to the likelihood ratio. Furthermore, a feature vector $f'_{i}$ is classified into either the normal or the defective class based on a threshold $\tau$ as the following.

\[
f'_{i} =
\begin{cases}
	Defective    & \text{if $ s_{i} > \tau $} \\
	Normal & \text{otherwise} \\
\end{cases}
\] 
Note that $\tau$ must be equal to or greater than one. Having a bigger value indicates the model allows more defective patches to be classified as normal patches. Empirically, we set $\tau$ to 1.05.

\section{Results and Evaluation\label{sec:evaluation}}

We conducted experiments to validate the effectiveness of the proposed model. First, we introduce our experiment environment, dataset, and metrics we used for quantitative analysis. Then, we present accuracy results along with output visualization. Next, we present the effectiveness of our model on two different choices of feature selection - hand-crafted~\cite{chetverikov2002finding}, PCA~\cite{beirao2004defect}, and residual neural networks~\cite{he2016deep} methods. We also explore the impact of two important parameters - image patch size and Gabor filter bank parameters on speed and accuracy (conditioned true positive rate). Last, we demonstrate the robustness of our proposed model by testing rotated fabric images.

\subsection{Experiment environment, dataset, and metrics}

We implement the model in Python. The testing environment is configured with AMD A10-6800K APU with a 2.5 GHz frequency. Our dataset\footnote{https://github.com/hzhou3/one-class-dataset} contains fabric images that are collected from the Standard Fabric Defect Glossary (SFDG)~\cite{StandardFabricDefectGlossary} dataset that consists of more complex fabric images than other simple datasets such as the TILDA Textile Texture Database~\cite{TILDATextileTextureDatabase}. Because the SFDG dataset contains a lot of duplicated fabric images, we then select 31 representatives that cover most of cases in the SFDG dataset (i.e., every 512$\times$512 fabric image differs in fabric backgrounds, defects, colors, etc). We hope that our proposed model can solve the defect detection problem on these images so that we show the generosity of our model when compared with other models that work for certain types of fabrics. In the experiments, we use several metrics to quantify the performance of our model, such as true positive rate (TPR), false positive rate (FPR), and receiver operating characteristic curve (ROC curve). TPR is defined as $TP/(TP+FN)$ and FPR is defined as $FP/(TP+FN)$ where $TP$, $TN$, $FP$, $FN$ indicate true positive (defect samples classified as defects), true negative (normal samples classified as normal), false positive (normal samples classified as defects), and false negative (defect samples classified as normal), respectively. By these metrics, we are able to measure the model's ability to distinguish between the classes. In particular, we focus on TPR (recall) when FPR (ratio of false alarm) is equal to 0. We call this metric a conditioned TPR (cTPR). We value this metric since it can represent the real-world situation where a small portion of false alarms results in significant economic losses.

\subsection{Visualized Results}

\begin{figure*}
	\centering
	\begin{tabular}{c c c c c c}
		
		\subfloat{\includegraphics[width=0.15\textwidth, scale=0.7]{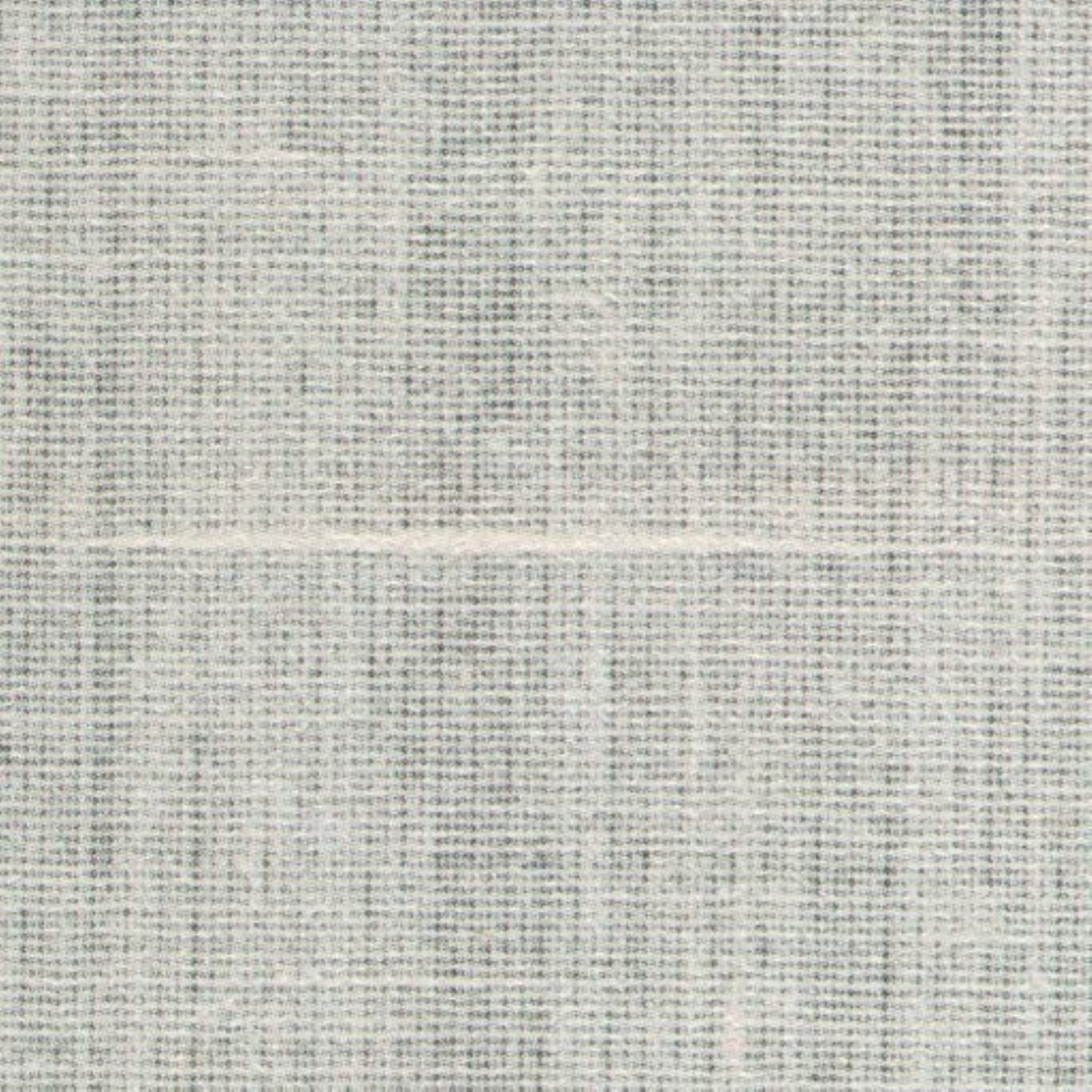}} &
		\subfloat{\includegraphics[width=0.15\textwidth, scale=0.7]{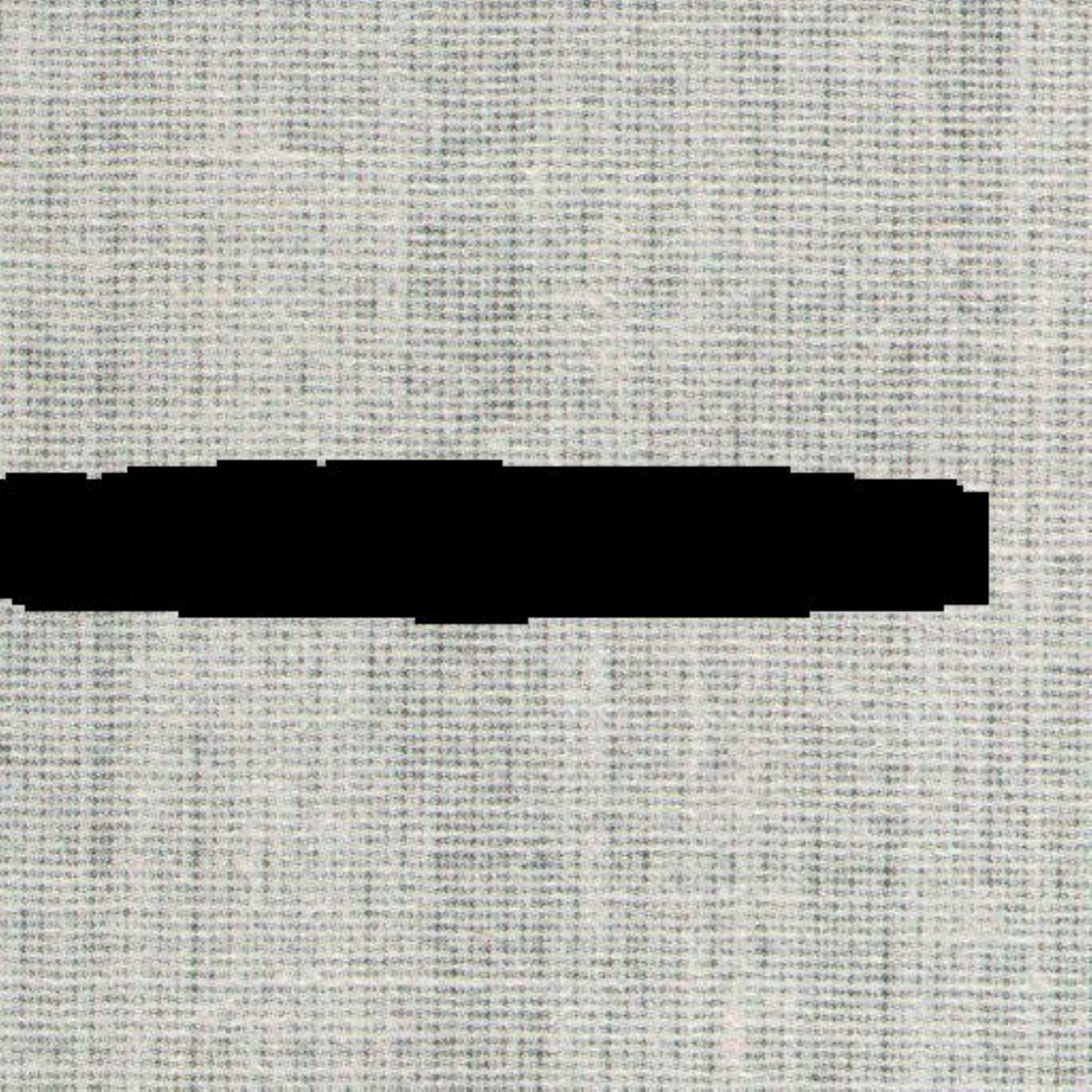}} &
		\subfloat{\includegraphics[width=0.15\textwidth, scale=0.7]{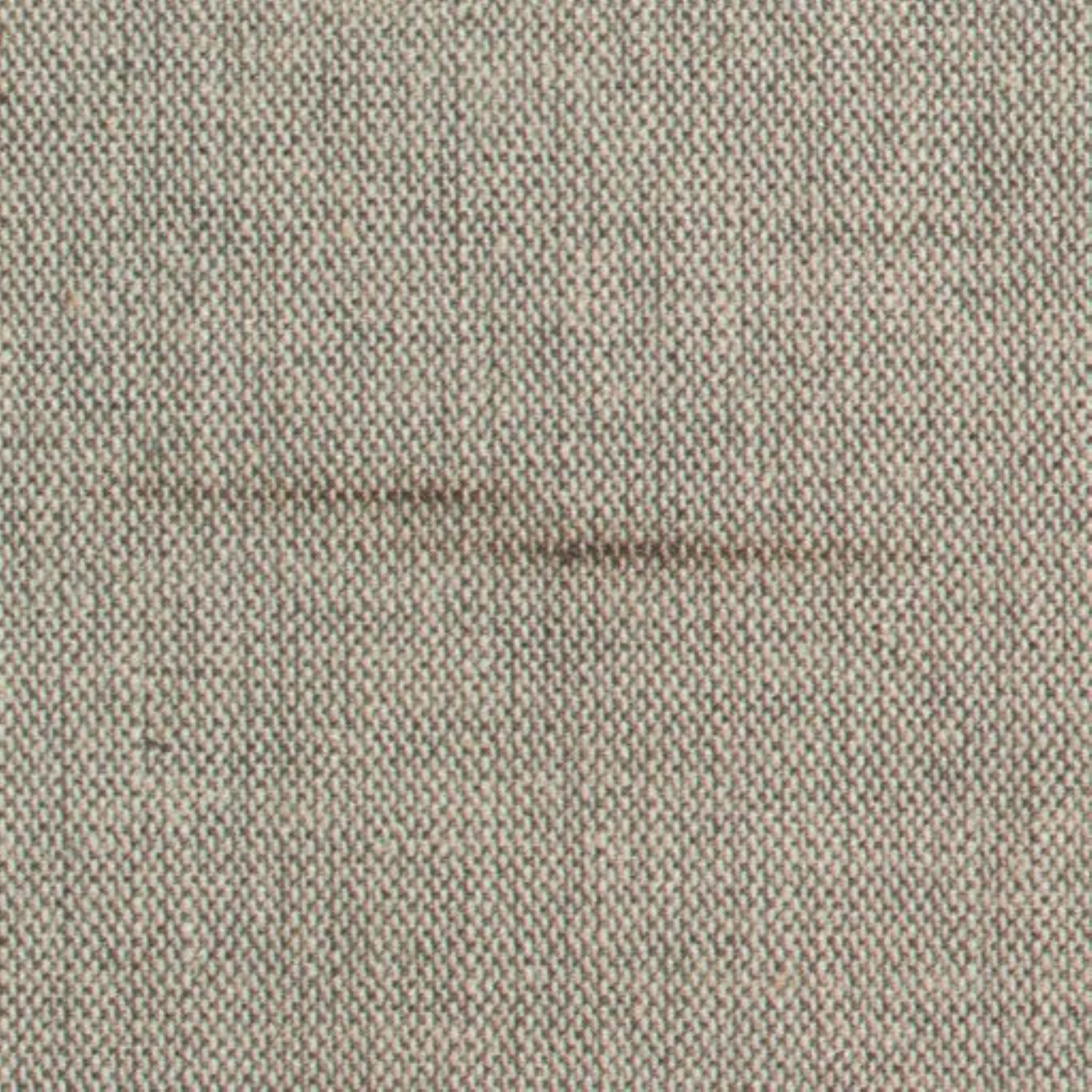}} &
		\subfloat{\includegraphics[width=0.15\textwidth, scale=0.7]{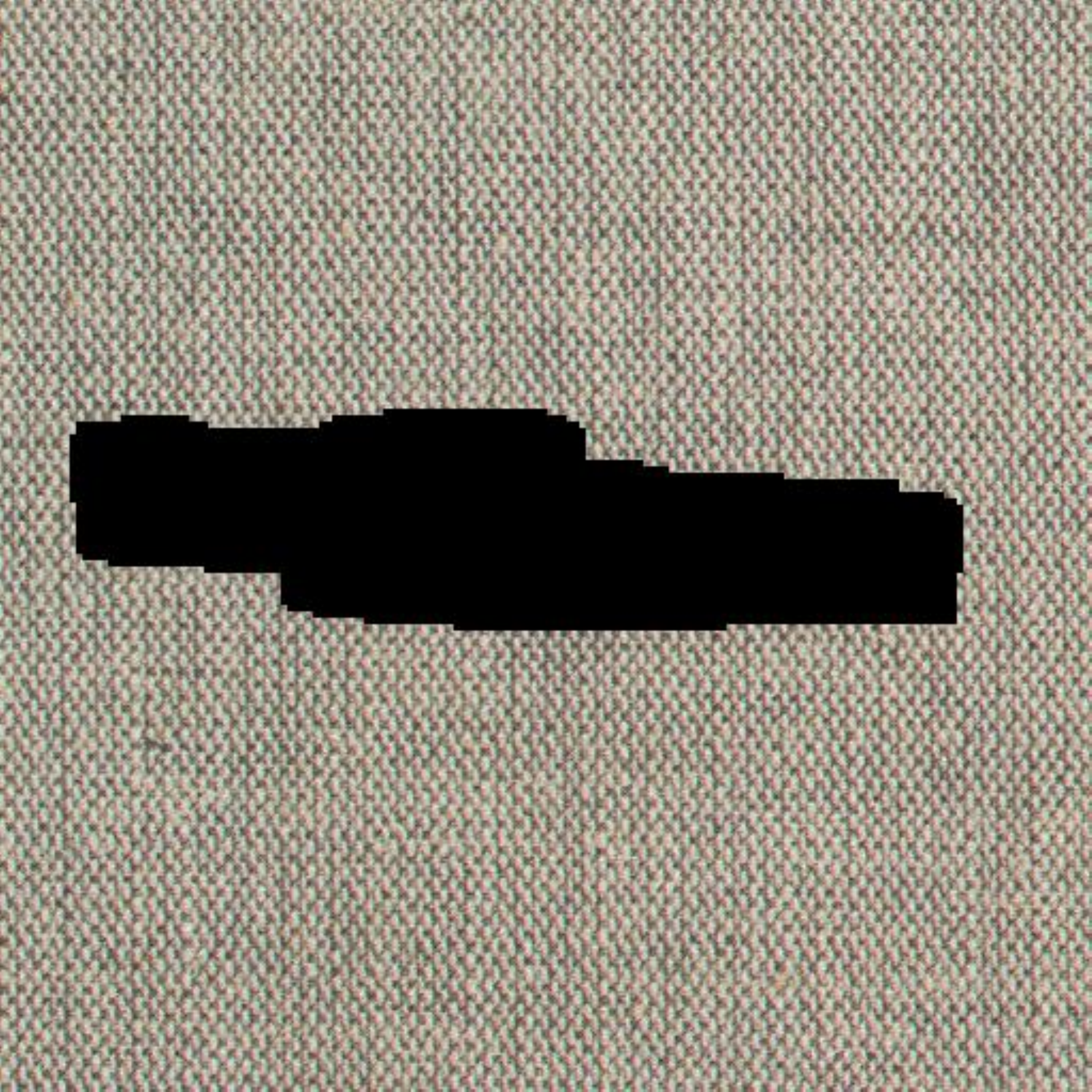}} &
		\subfloat{\includegraphics[width=0.15\textwidth, scale=0.7]{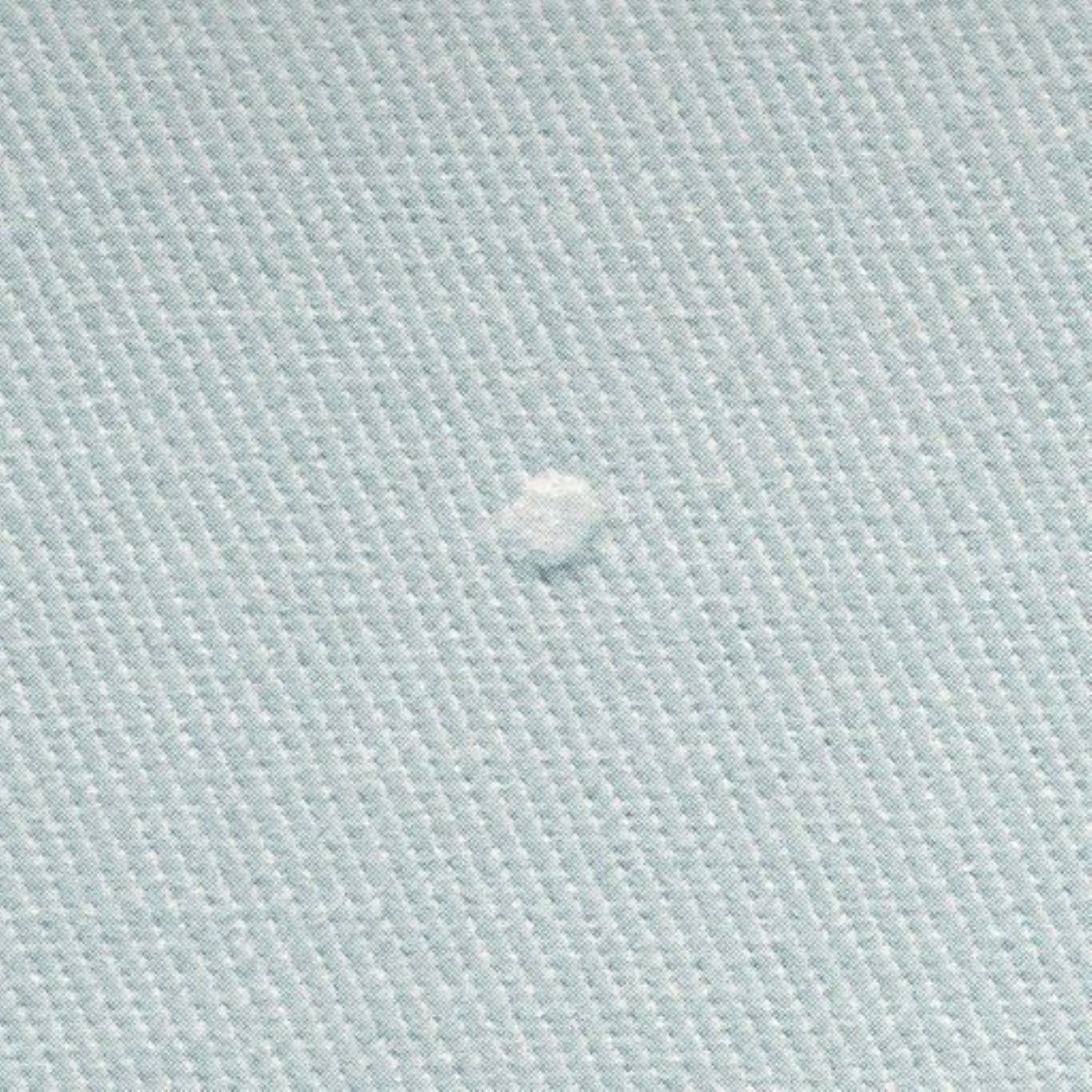}} &
		\subfloat{\includegraphics[width=0.15\textwidth, scale=0.7]{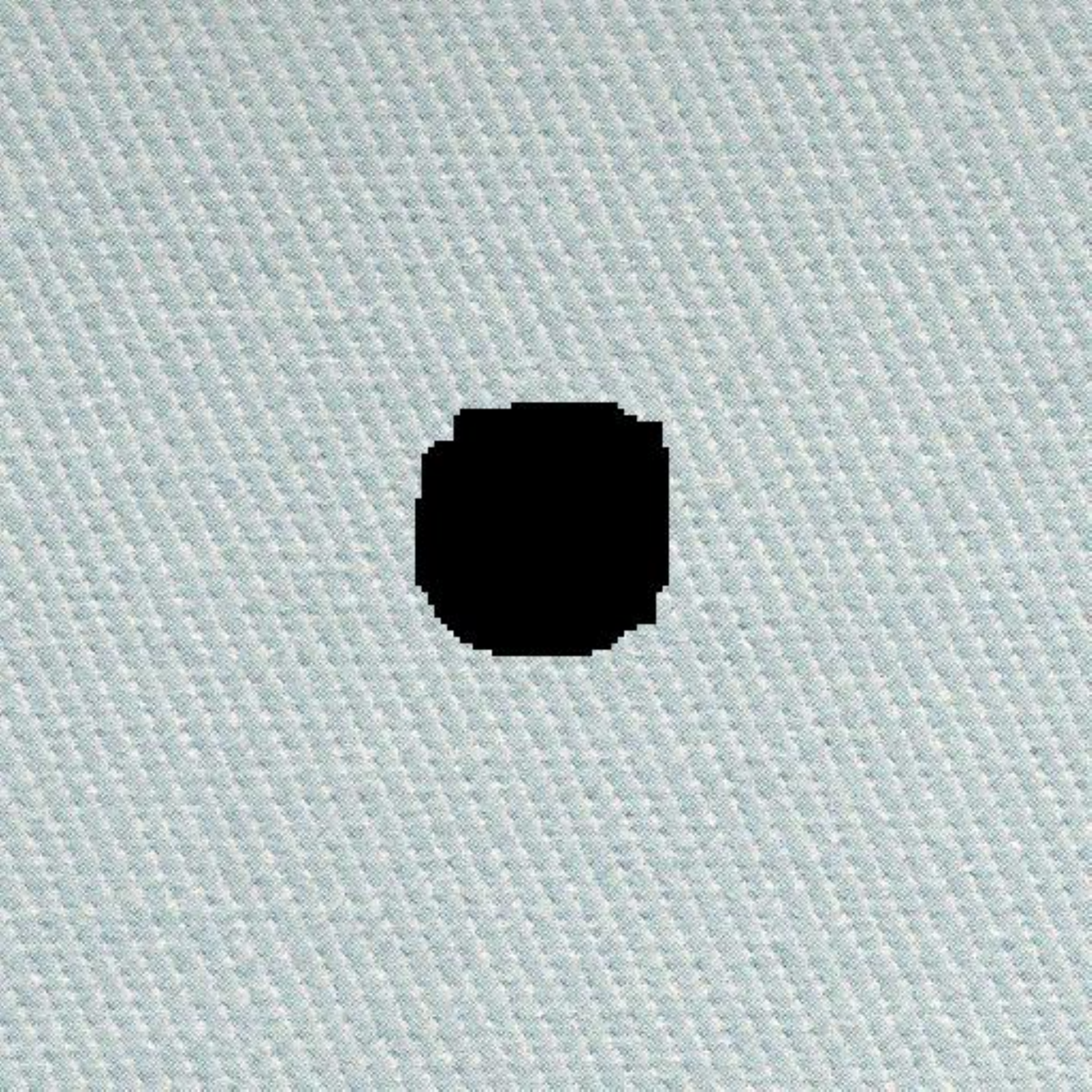}} \\
		
		\subfloat{\includegraphics[width=0.15\textwidth, scale=0.7]{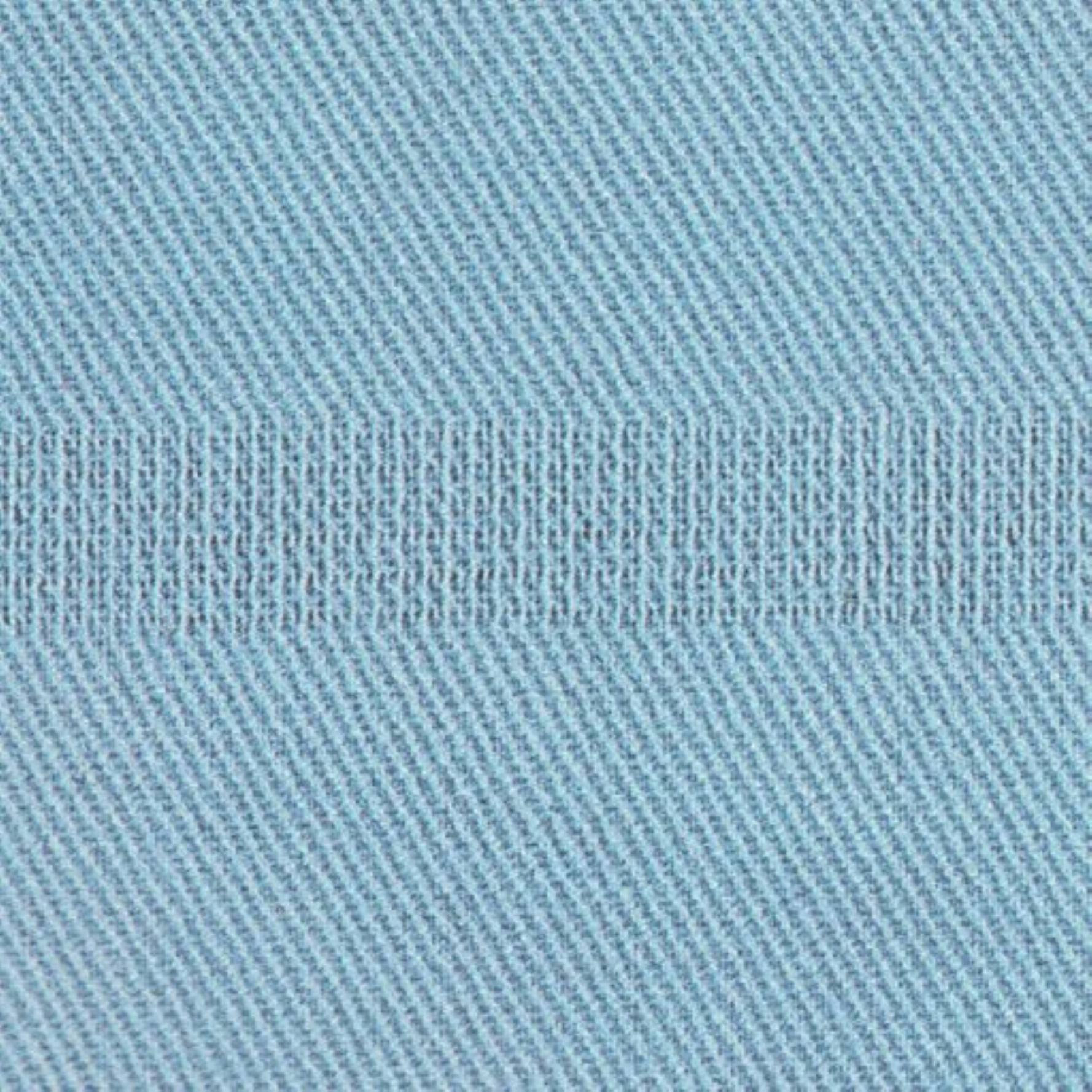}} &
		\subfloat{\includegraphics[width=0.15\textwidth, scale=0.7]{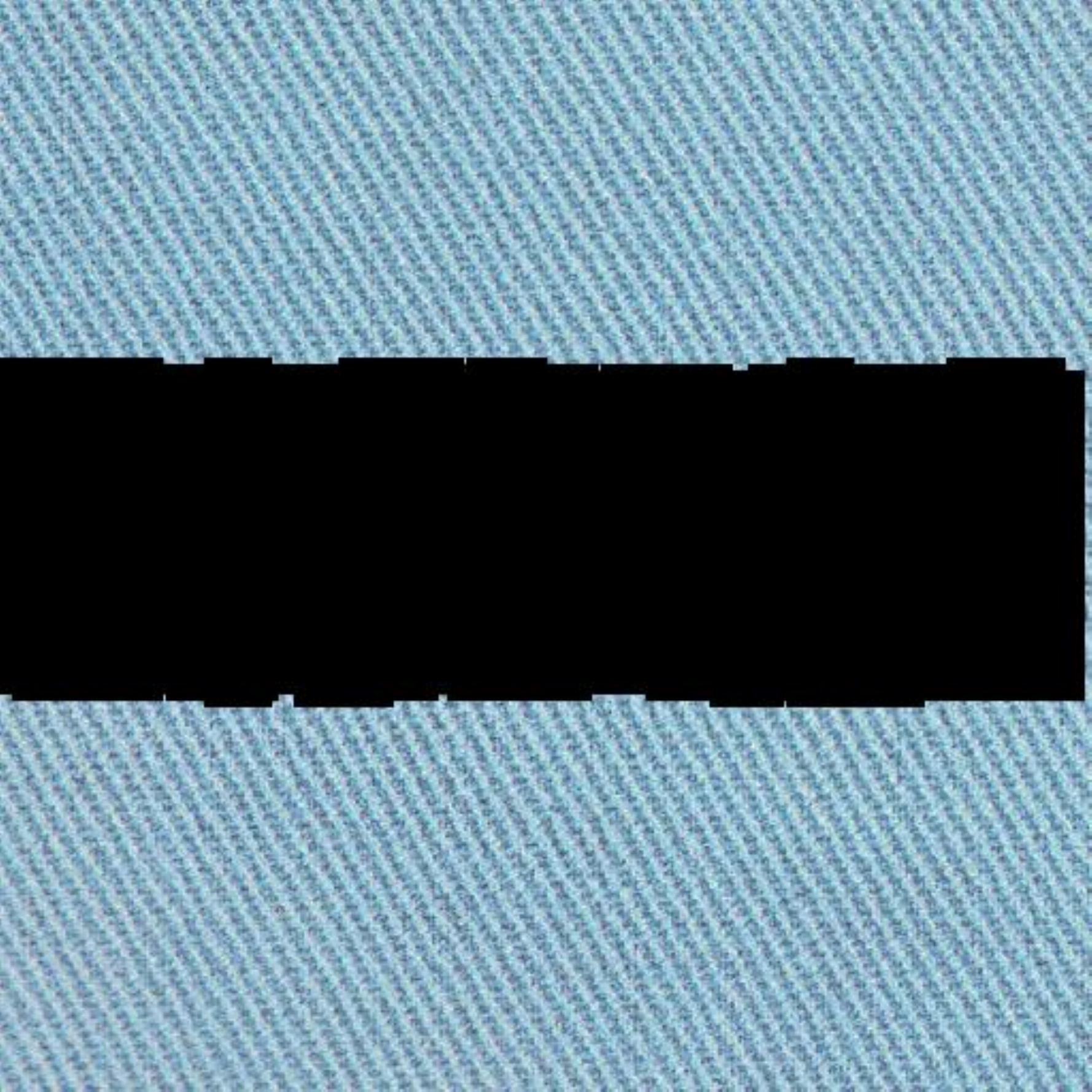}} &
		\subfloat{\includegraphics[width=0.15\textwidth, scale=0.7]{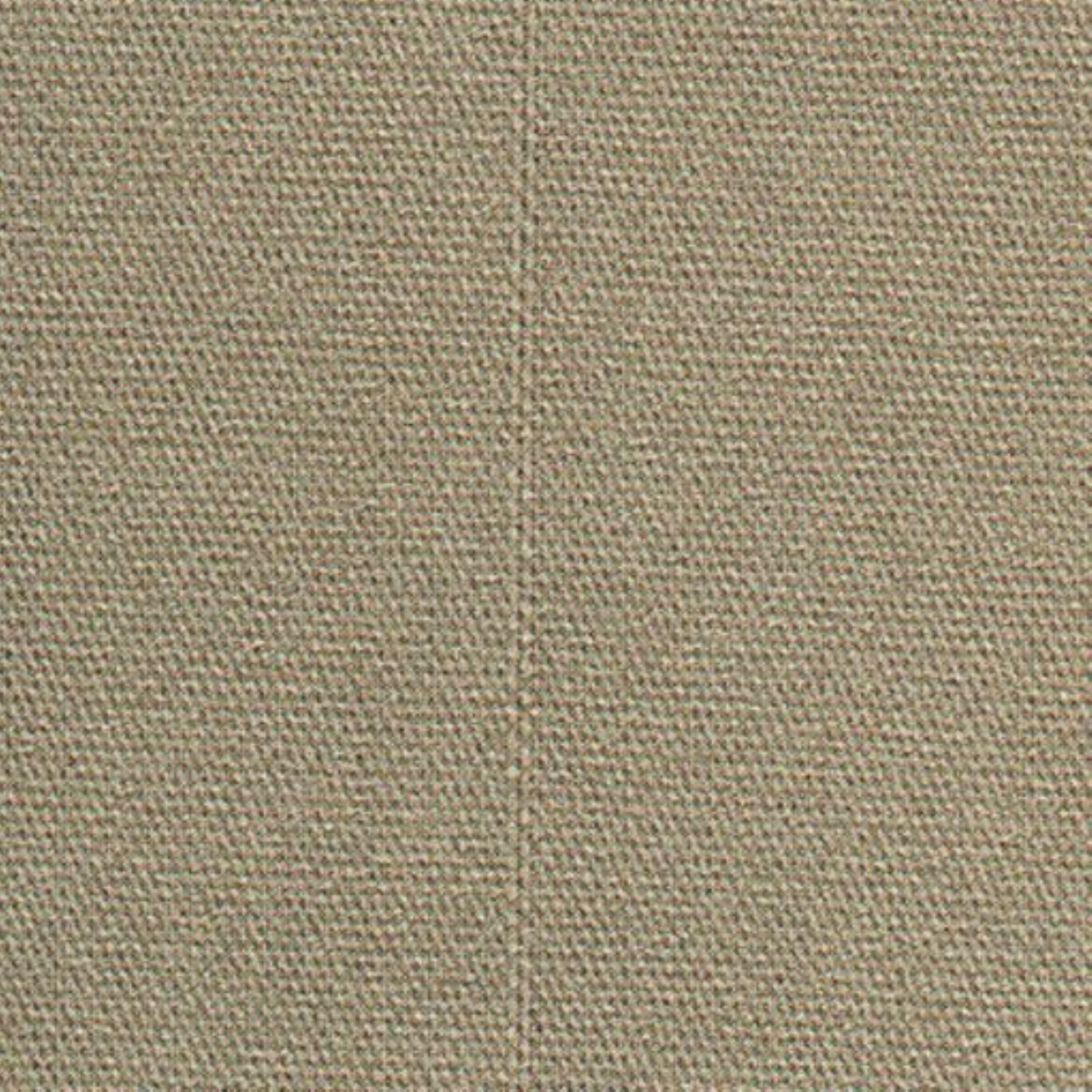}} &
		\subfloat{\includegraphics[width=0.15\textwidth, scale=0.7]{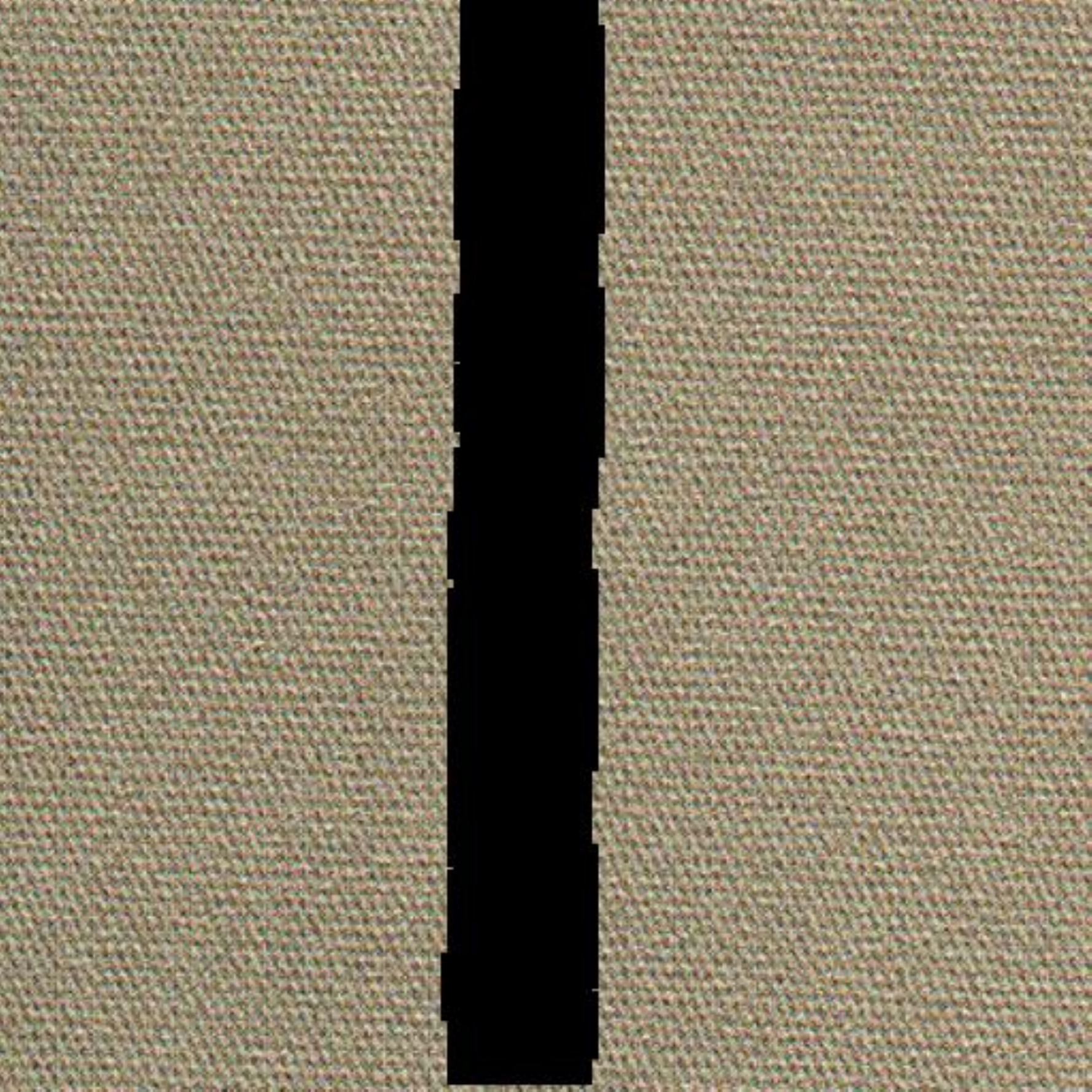}} &
		\subfloat{\includegraphics[width=0.15\textwidth, scale=0.7]{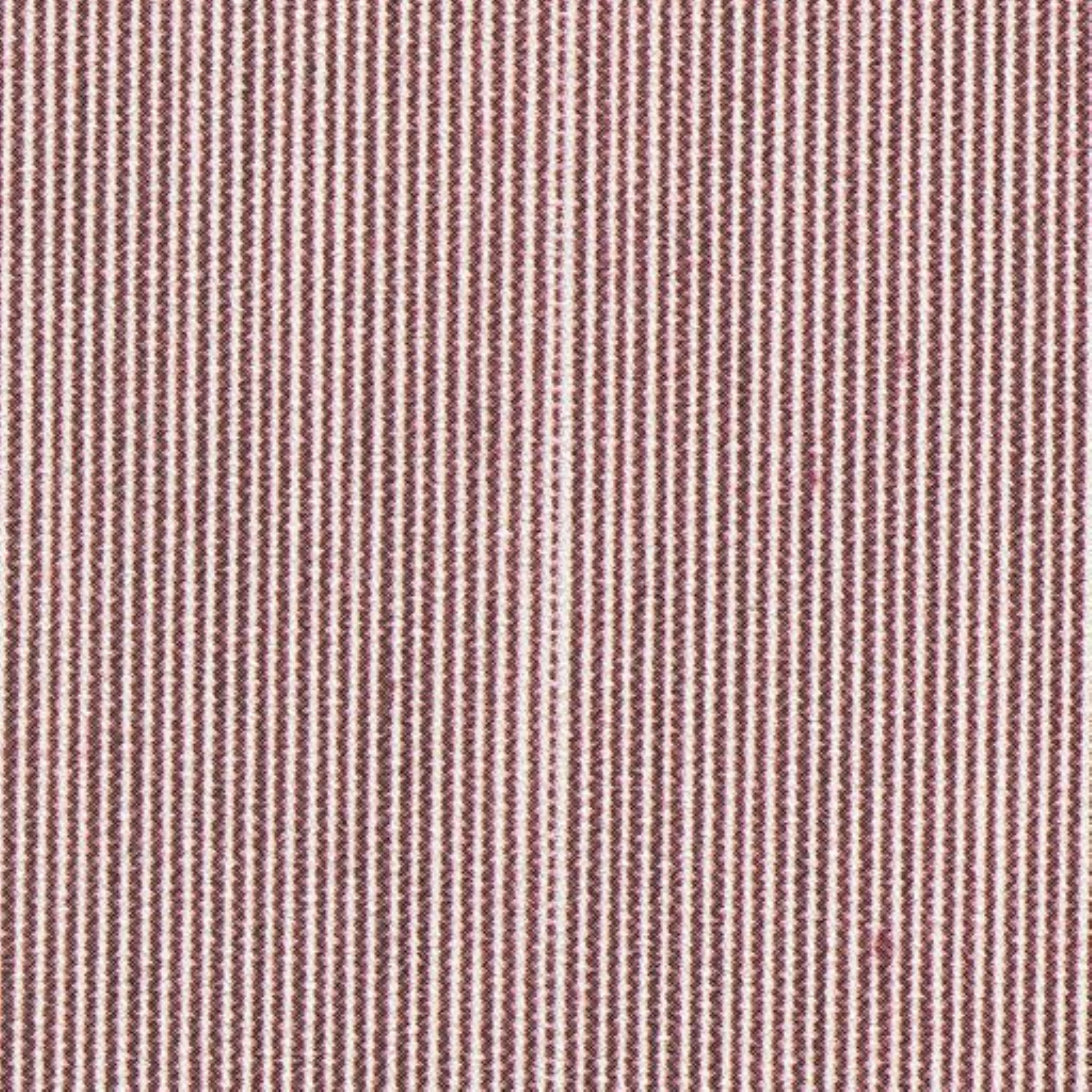}} &
		\subfloat{\includegraphics[width=0.15\textwidth, scale=0.7]{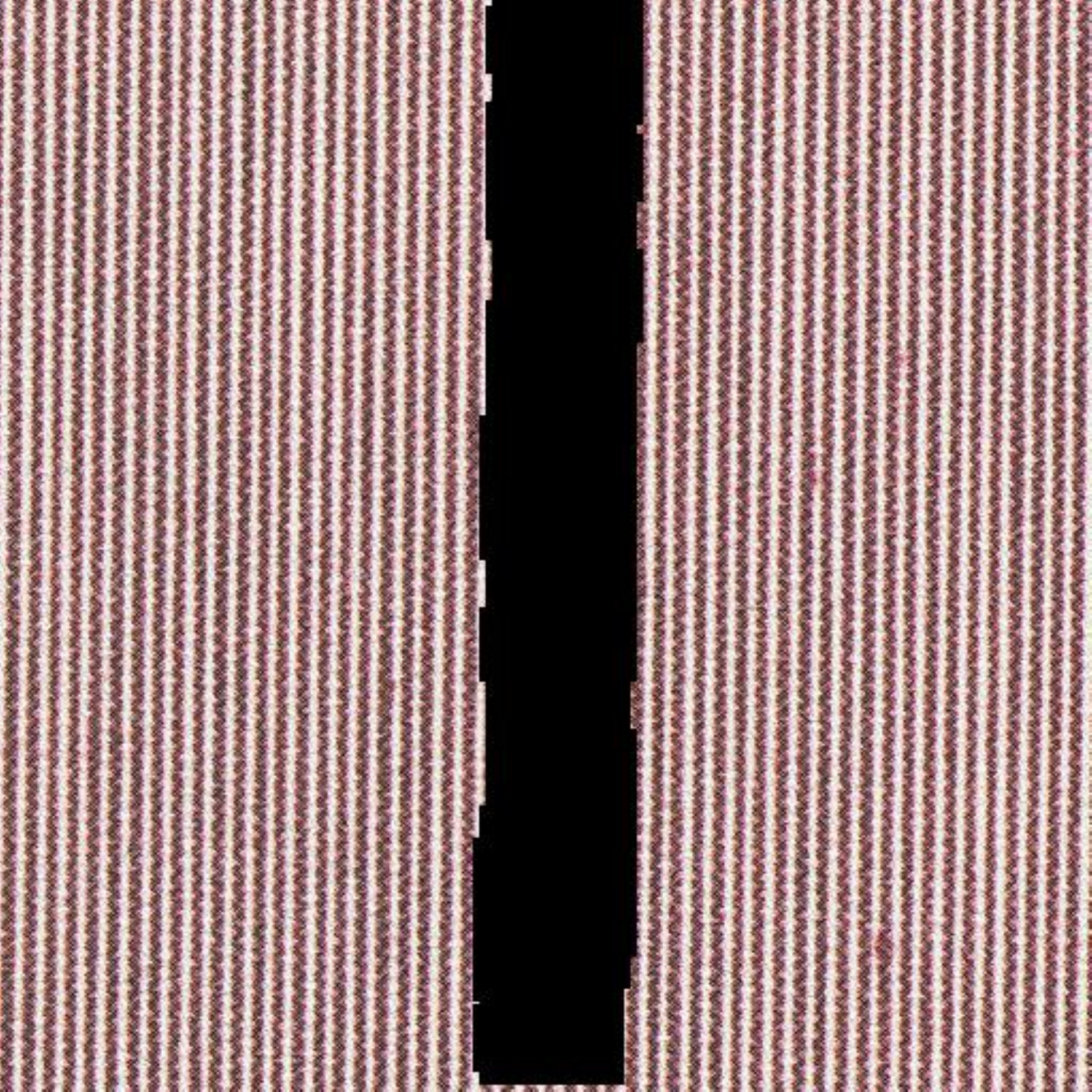}} \\
		
		\subfloat{\includegraphics[width=0.15\textwidth, scale=0.7]{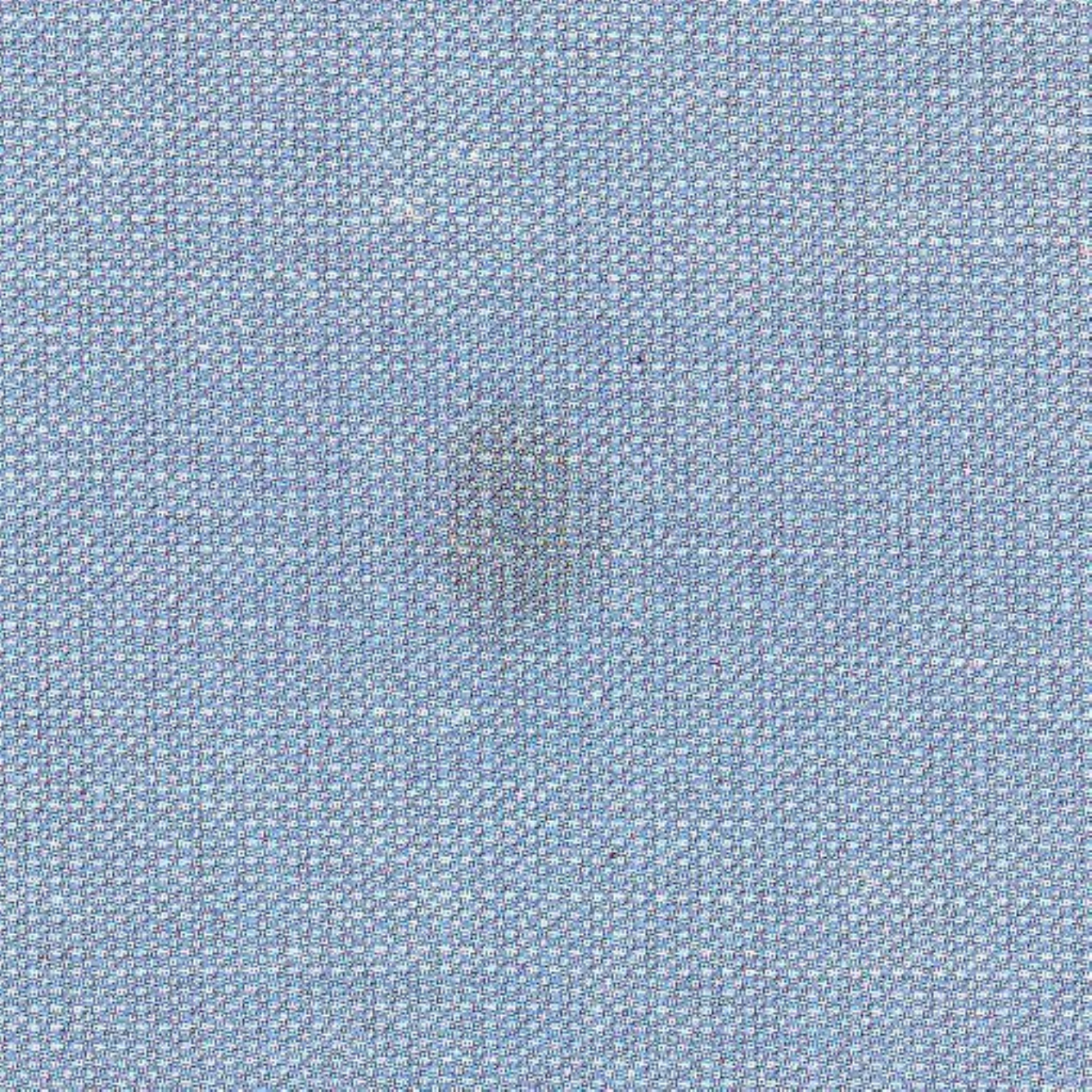}} &
		\subfloat{\includegraphics[width=0.15\textwidth, scale=0.7]{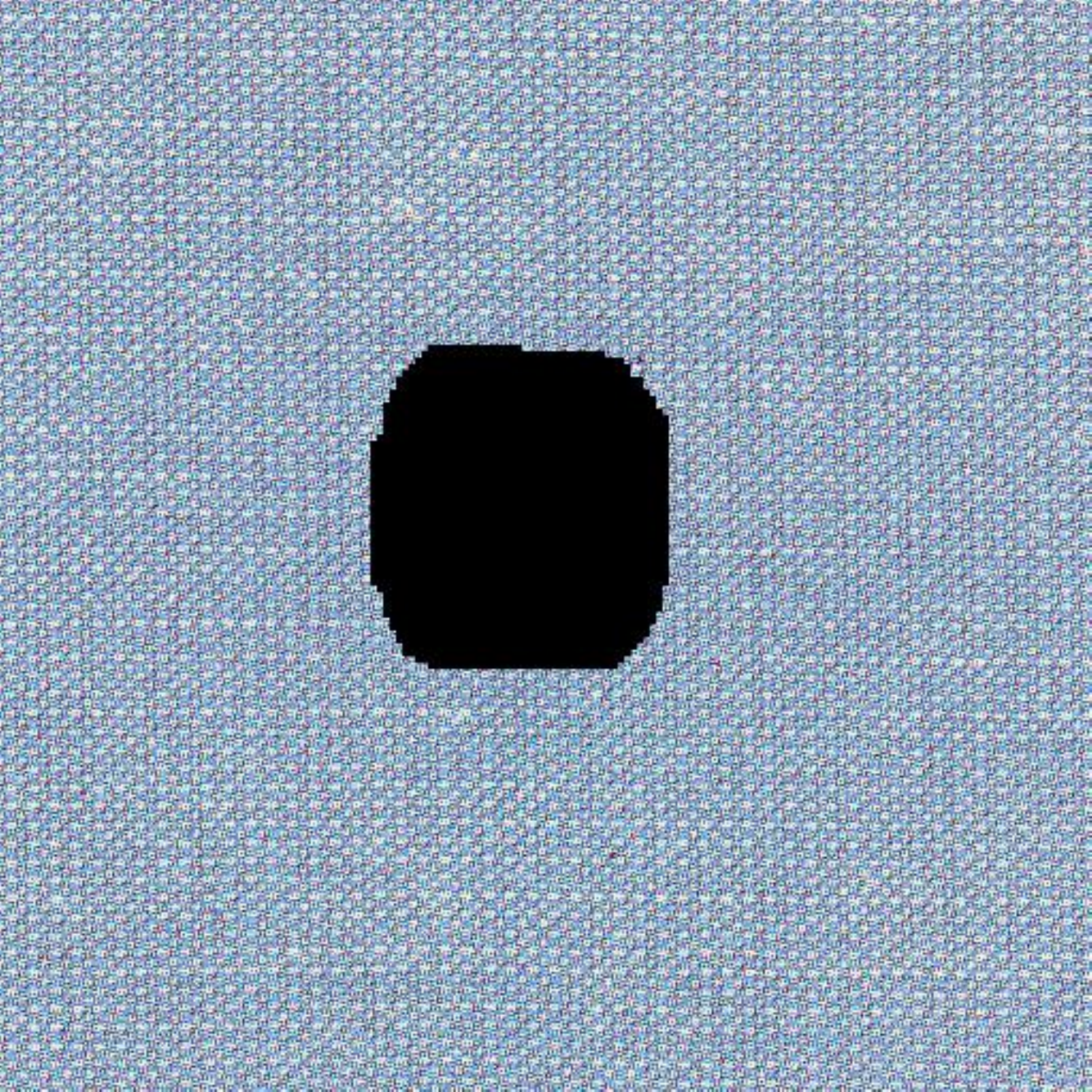}} &
		\subfloat{\includegraphics[width=0.15\textwidth, scale=0.7]{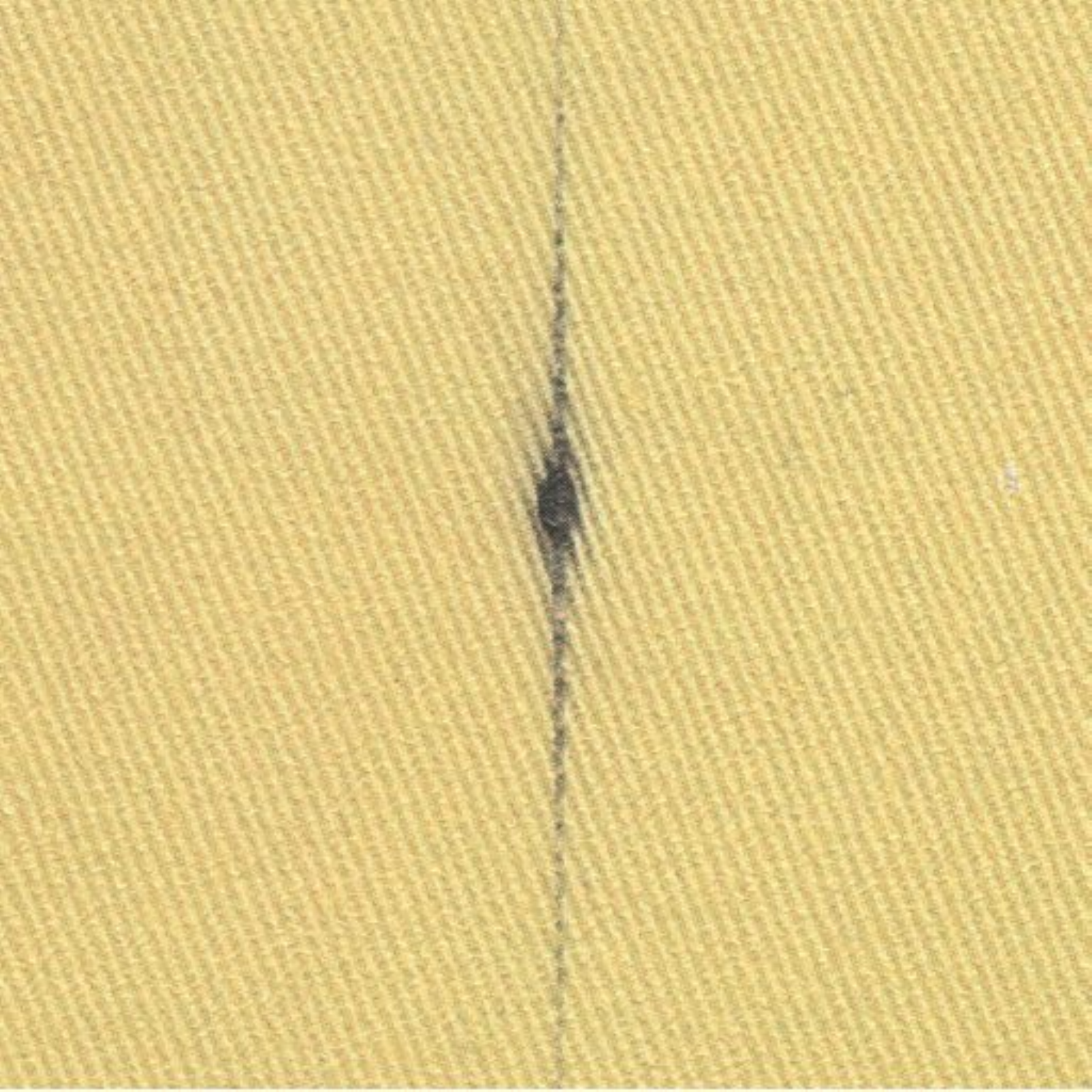}} &
		\subfloat{\includegraphics[width=0.15\textwidth, scale=0.7]{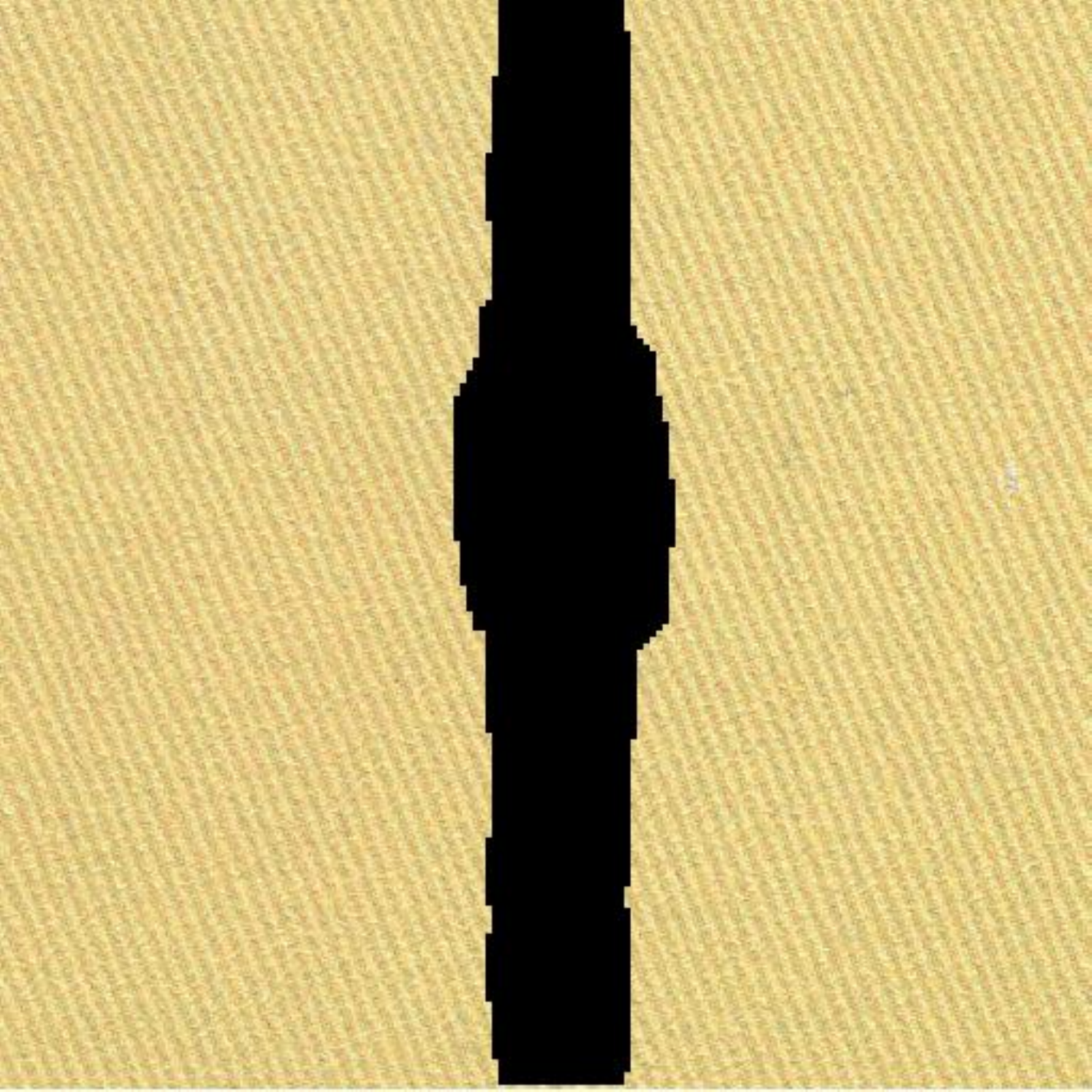}} &
		\subfloat{\includegraphics[width=0.15\textwidth, scale=0.7]{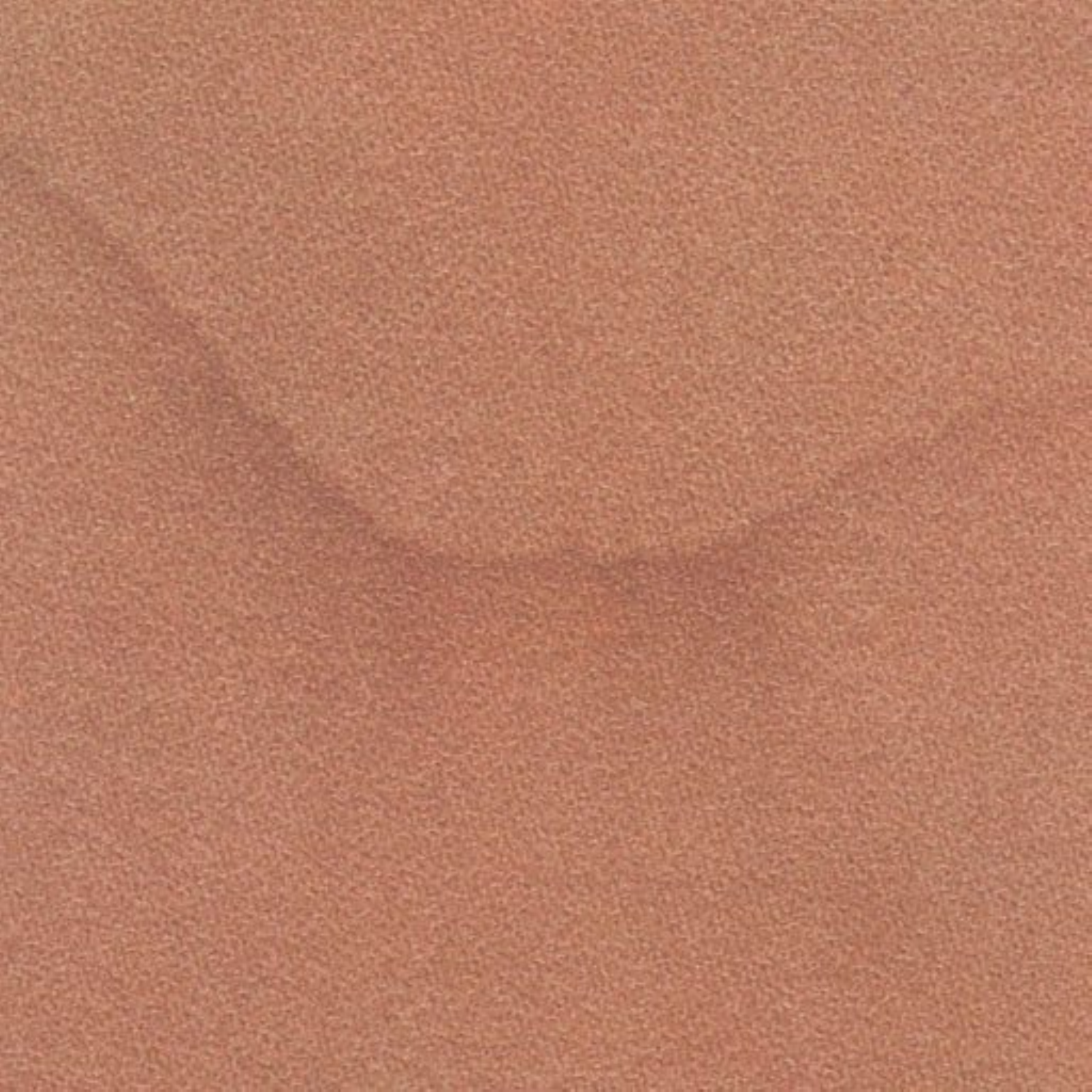}} &
		\subfloat{\includegraphics[width=0.15\textwidth, scale=0.7]{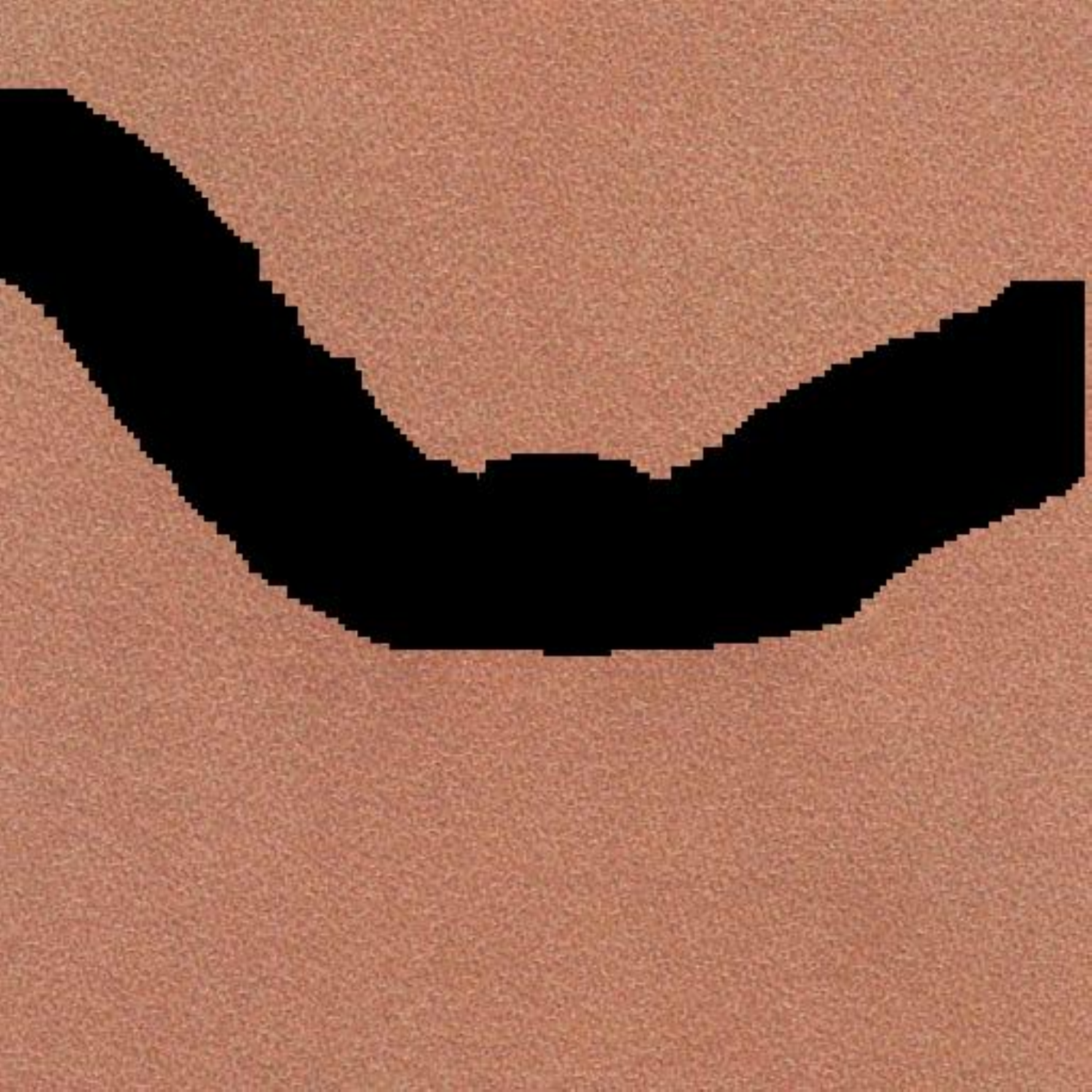}} \\

		Input image & Predicted image & Input image & Predicted image & Input image & Predicted image
	\end{tabular}
	\caption{Visual outputs of plain fabric defect detection (1st, 3rd, 5th columns present input images and 2nd, 4th, 6th columns show defects detected in black).}
	\label{fig:plain}
\end{figure*}

\begin{figure*}
	\centering
	\begin{tabular}{c c c c c c}
		
		\subfloat{\includegraphics[width=0.15\textwidth, scale=0.7]{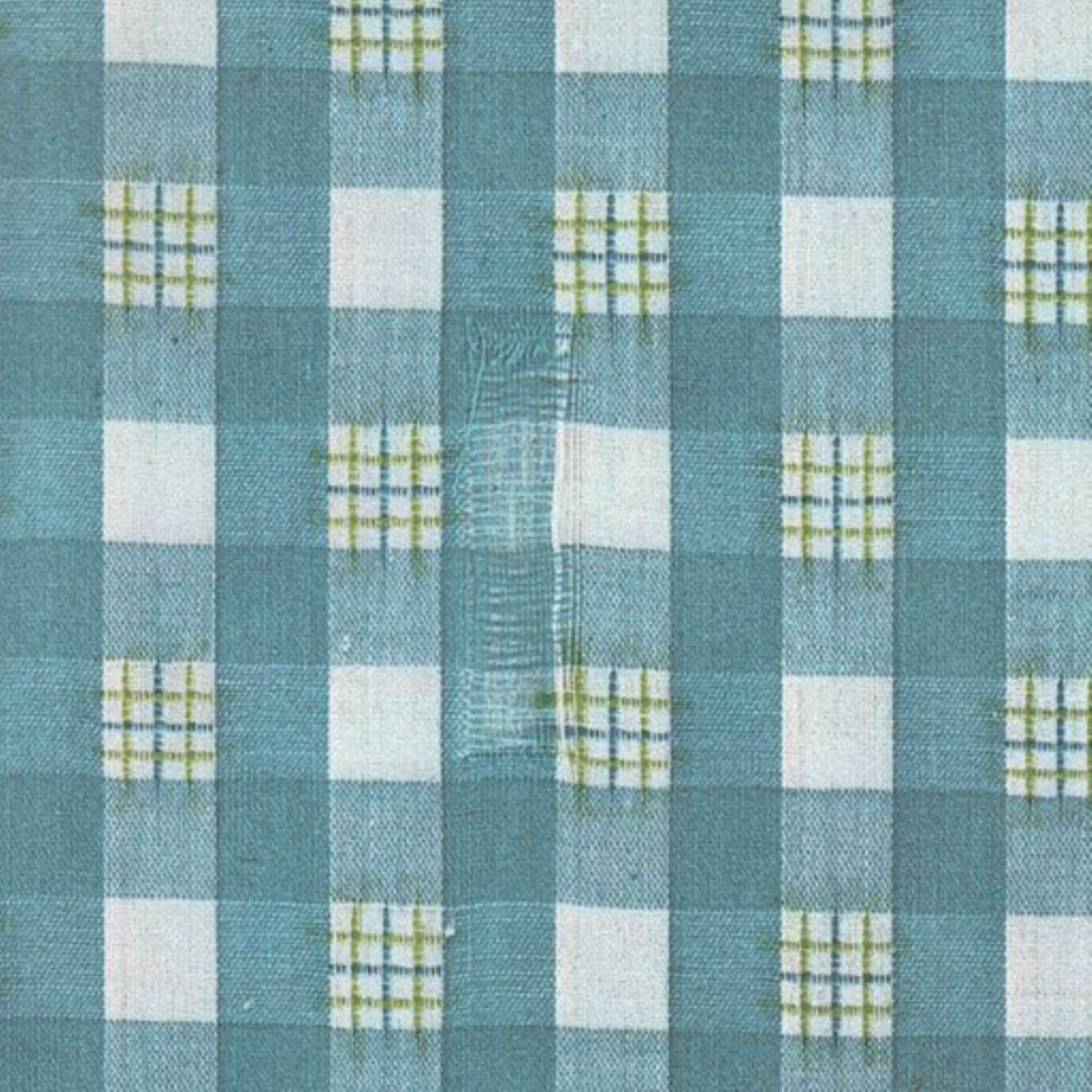}} &
		\subfloat{\includegraphics[width=0.15\textwidth, scale=0.7]{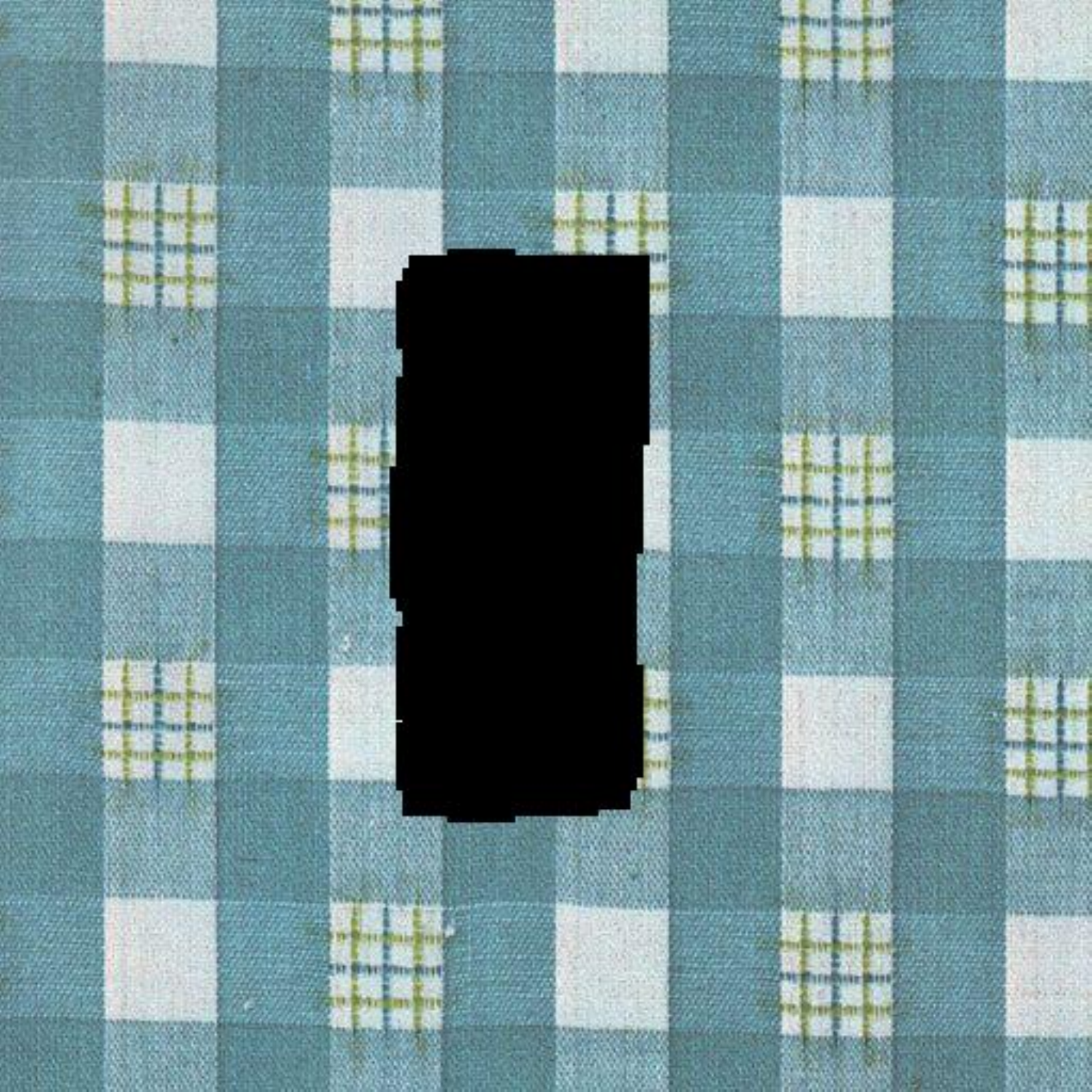}} &
		\subfloat{\includegraphics[width=0.15\textwidth, scale=0.7]{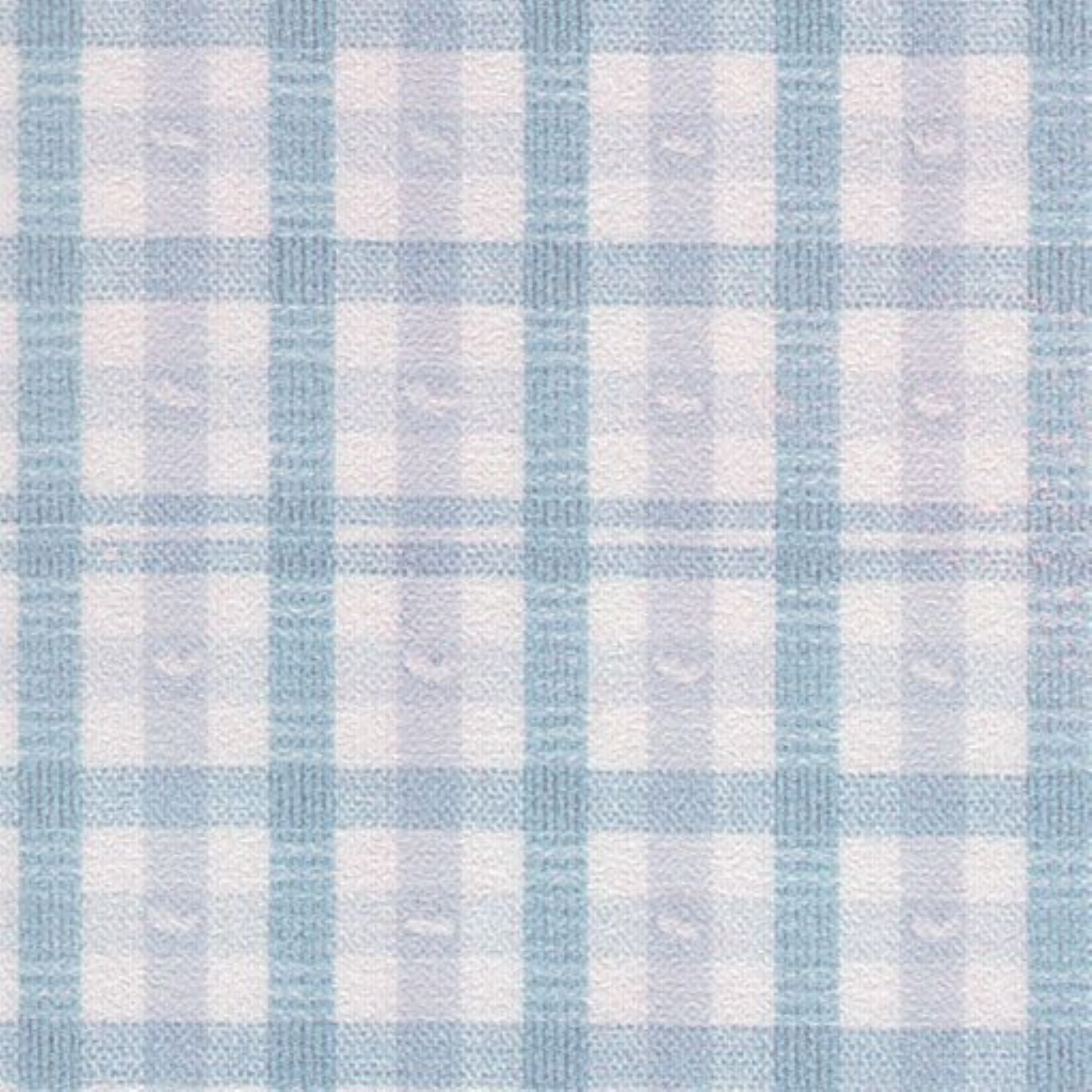}} &
		\subfloat{\includegraphics[width=0.15\textwidth, scale=0.7]{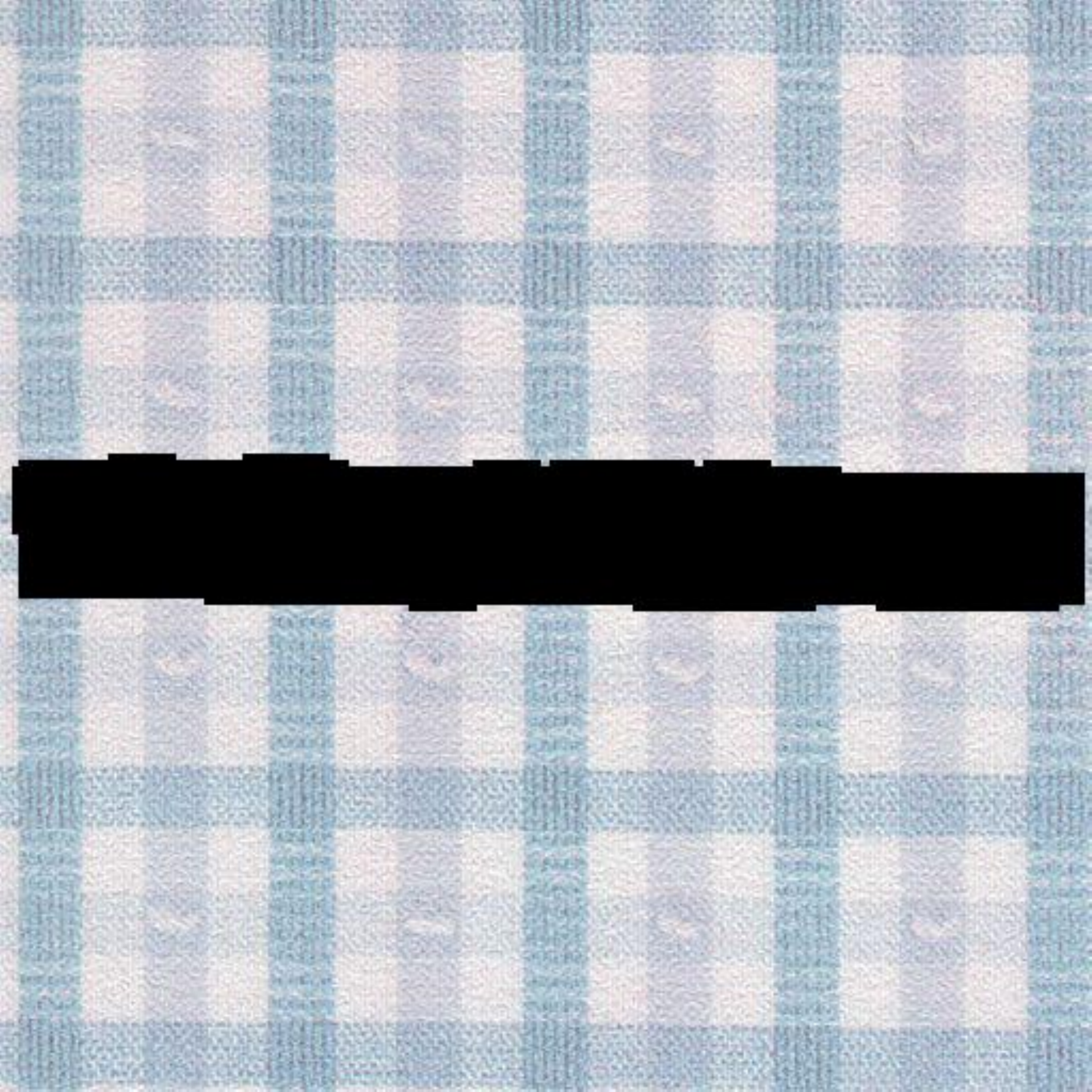}} &
		\subfloat{\includegraphics[width=0.15\textwidth, scale=0.7]{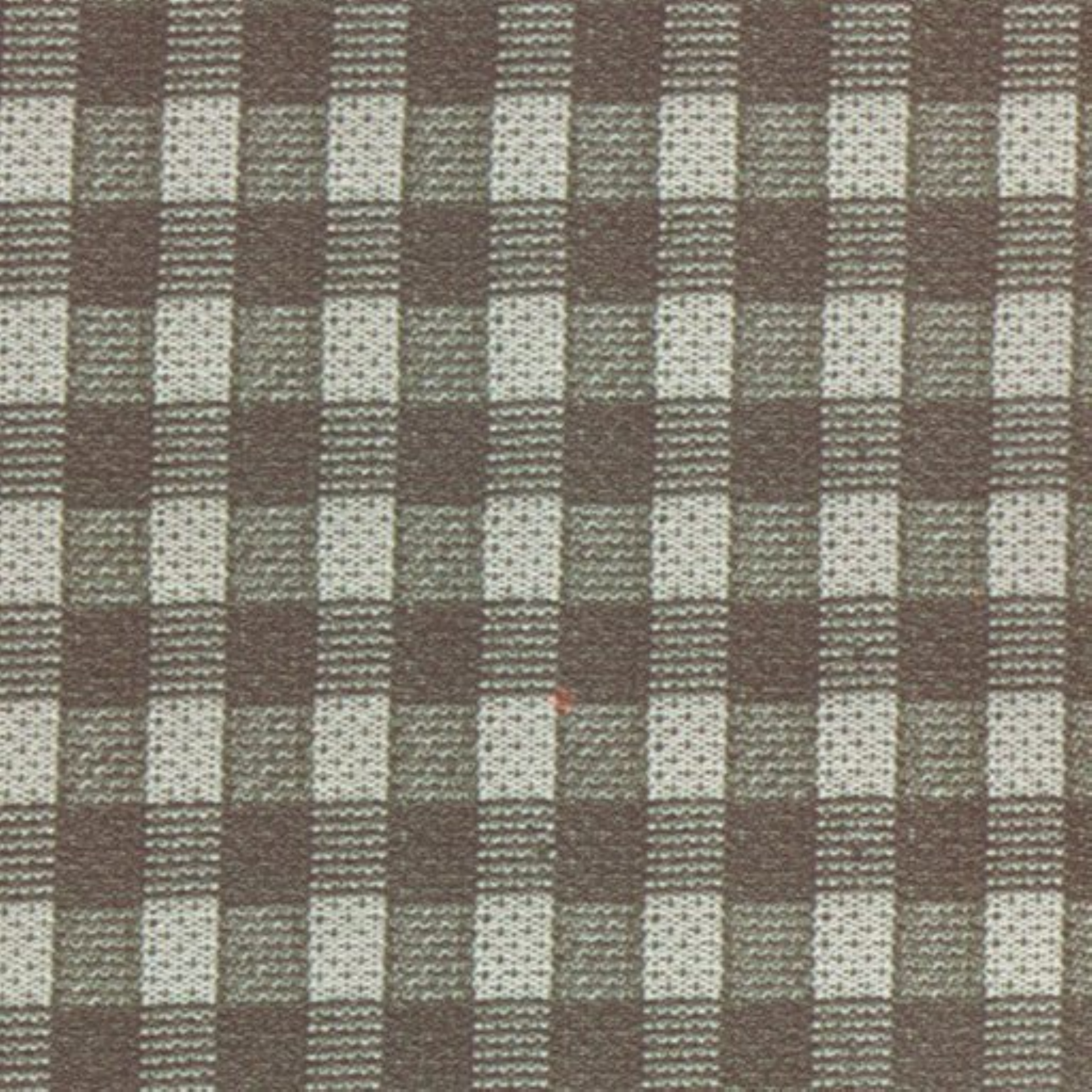}} &
		\subfloat{\includegraphics[width=0.15\textwidth, scale=0.7]{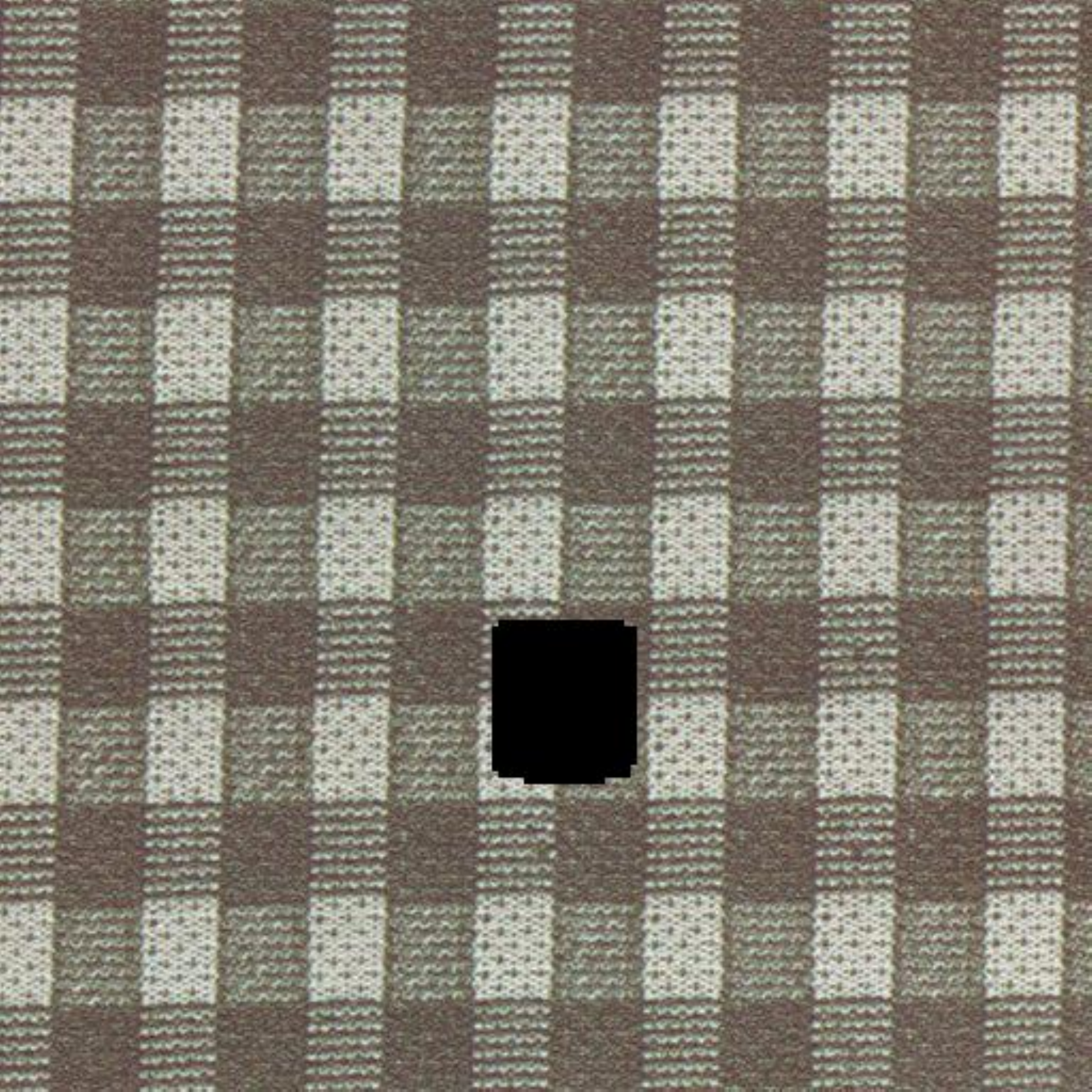}} \\

		\subfloat{\includegraphics[width=0.15\textwidth, scale=0.7]{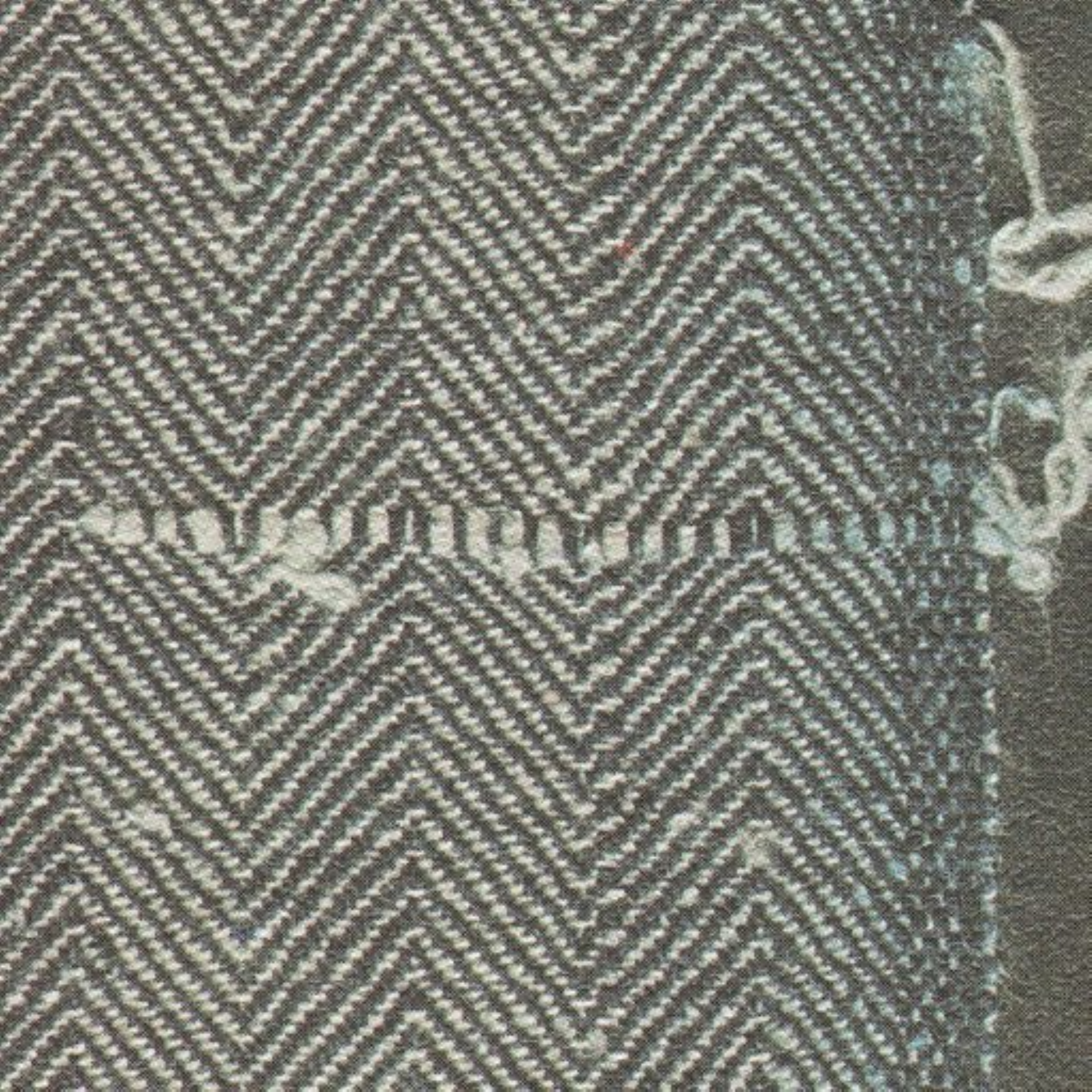}} &
		\subfloat{\includegraphics[width=0.15\textwidth, scale=0.7]{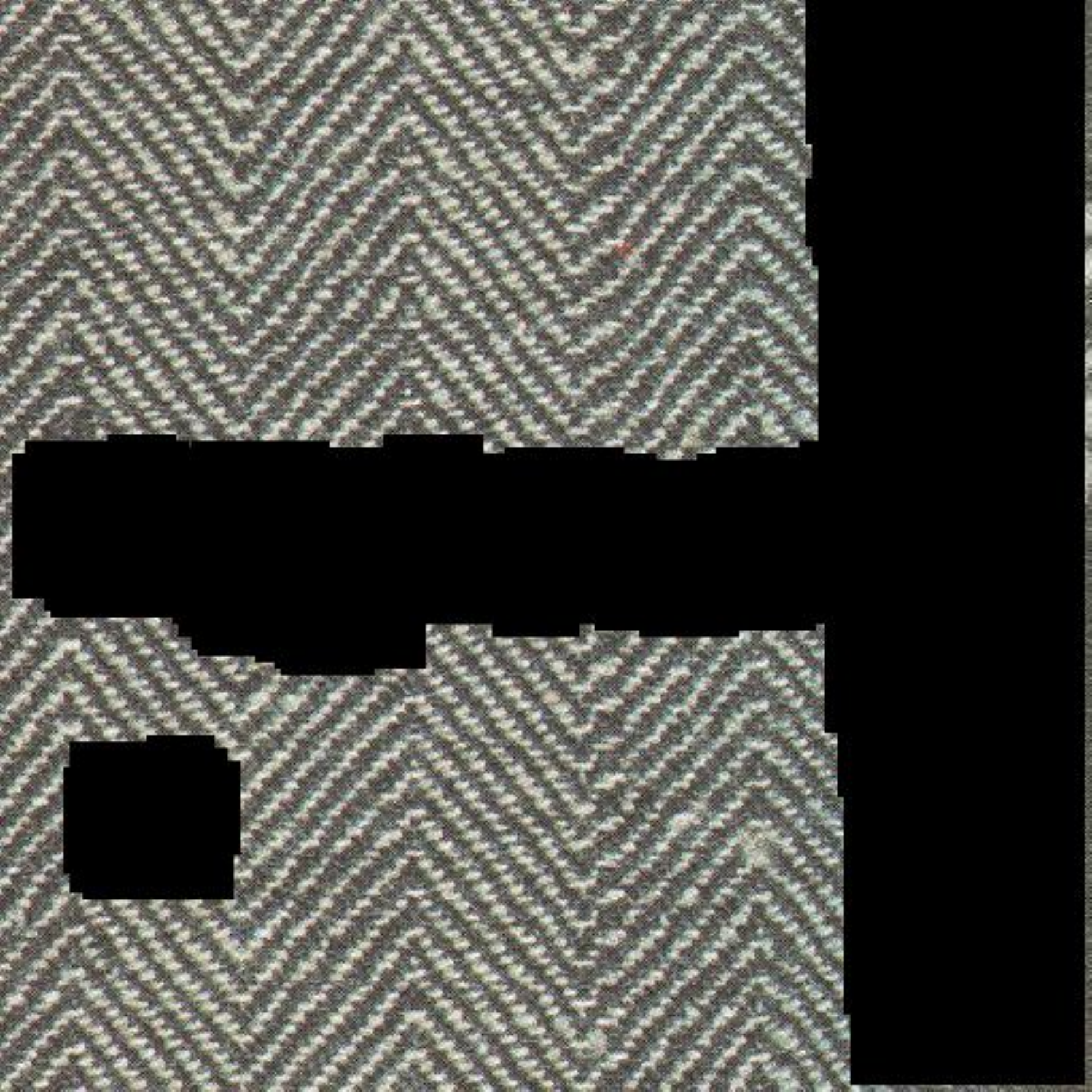}} &
		\subfloat{\includegraphics[width=0.15\textwidth, scale=0.7]{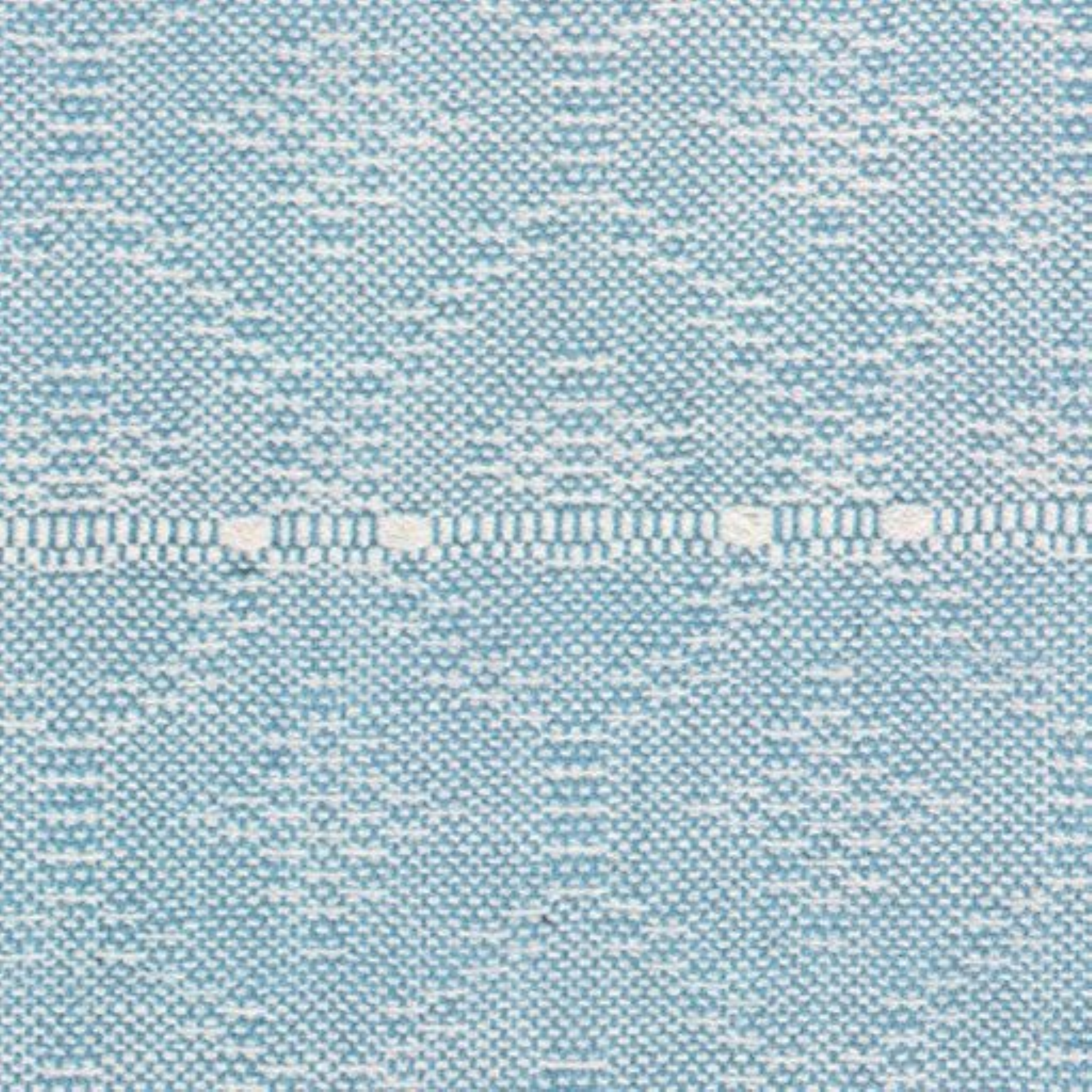}} &
		\subfloat{\includegraphics[width=0.15\textwidth, scale=0.7]{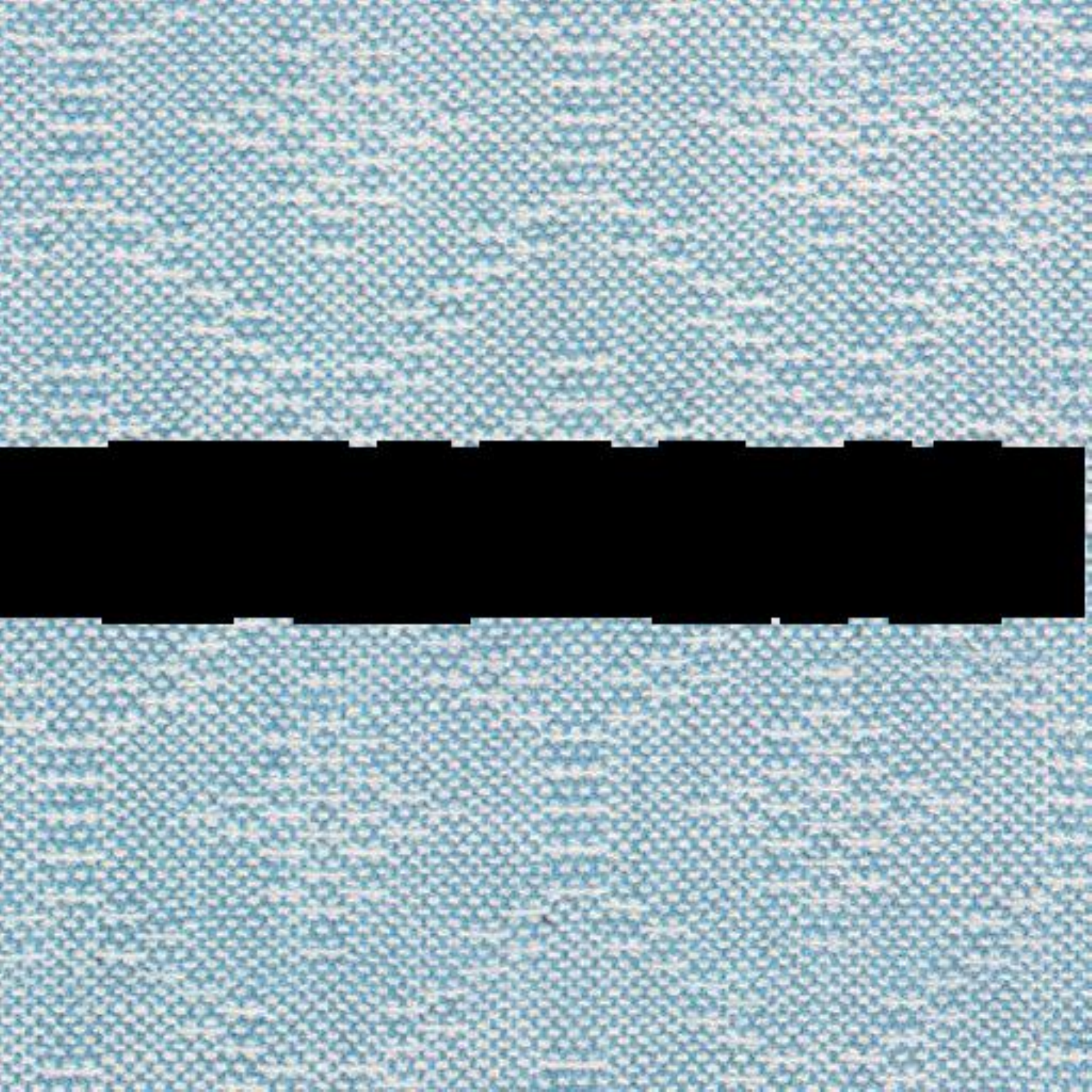}} &
		\subfloat{\includegraphics[width=0.15\textwidth, scale=0.7]{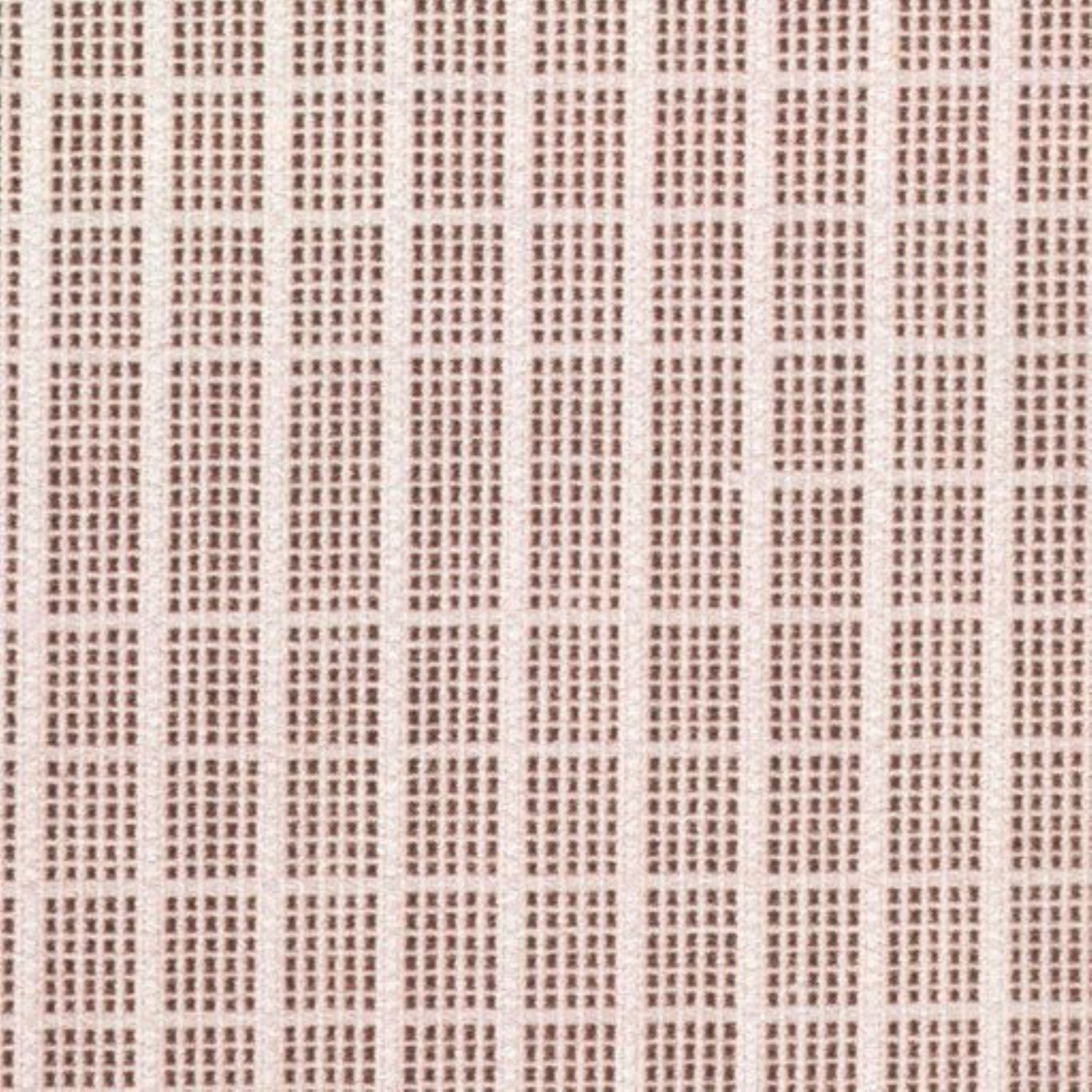}} &
		\subfloat{\includegraphics[width=0.15\textwidth, scale=0.7]{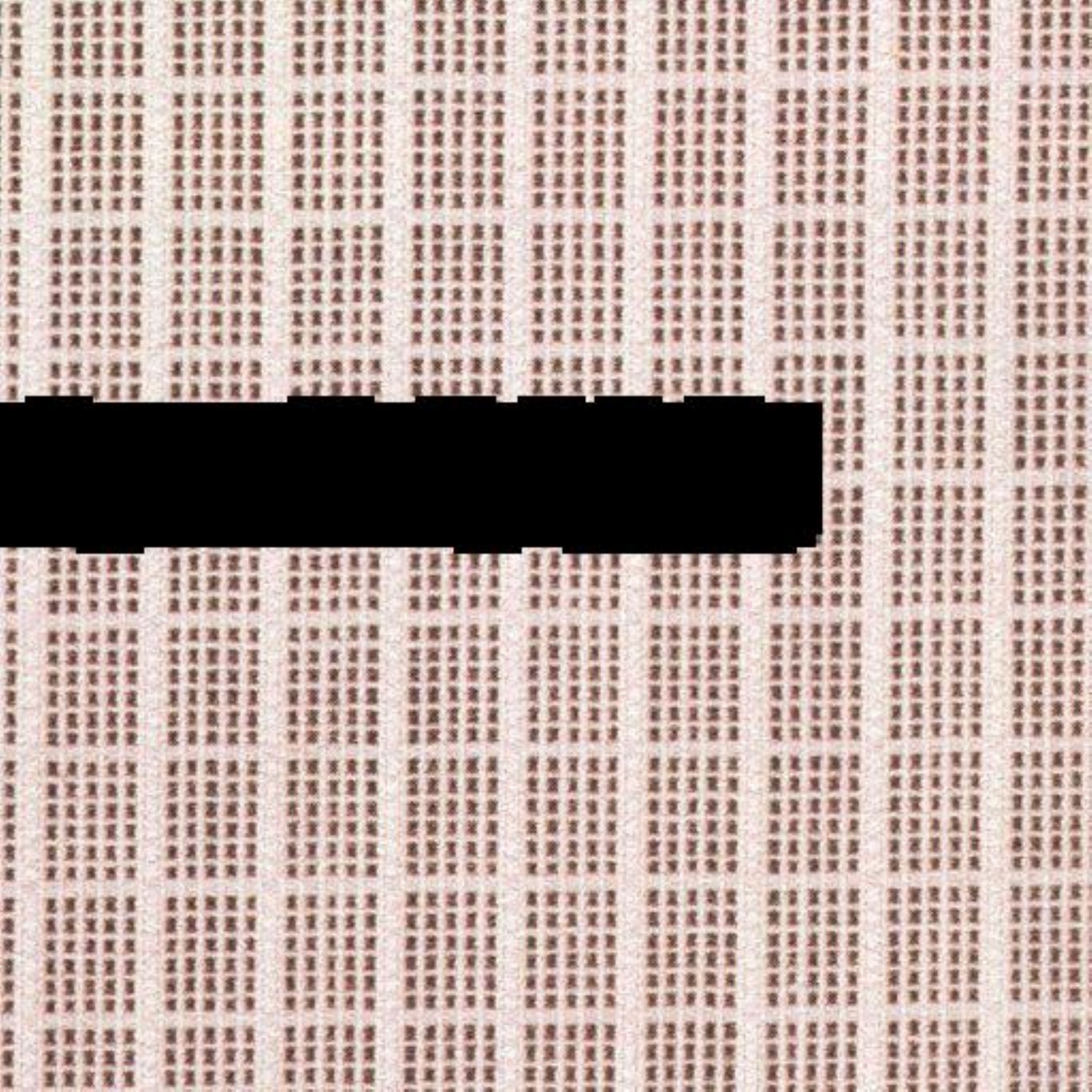}} \\
		
		\subfloat{\includegraphics[width=0.15\textwidth, scale=0.7]{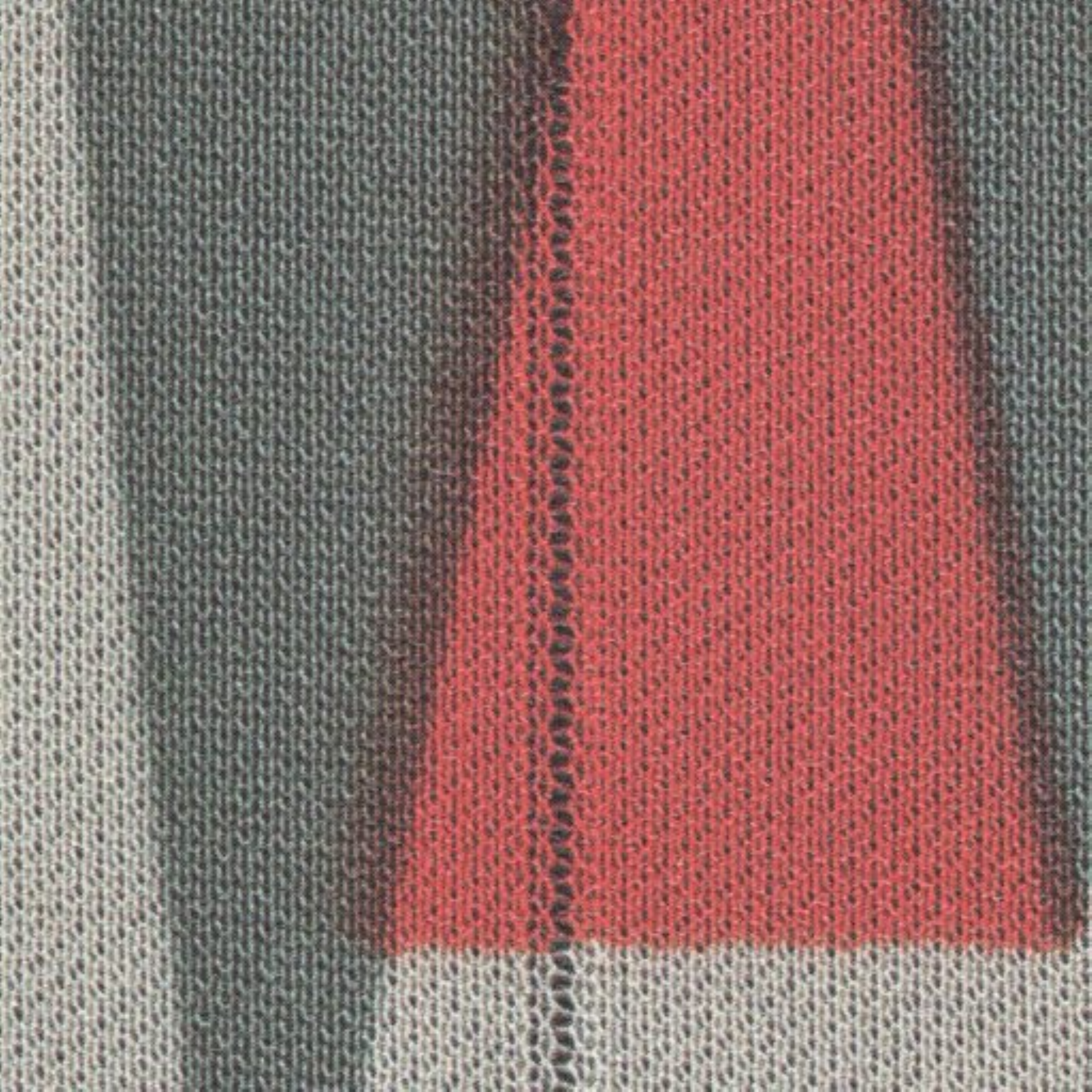}} &
		\subfloat{\includegraphics[width=0.15\textwidth, scale=0.7]{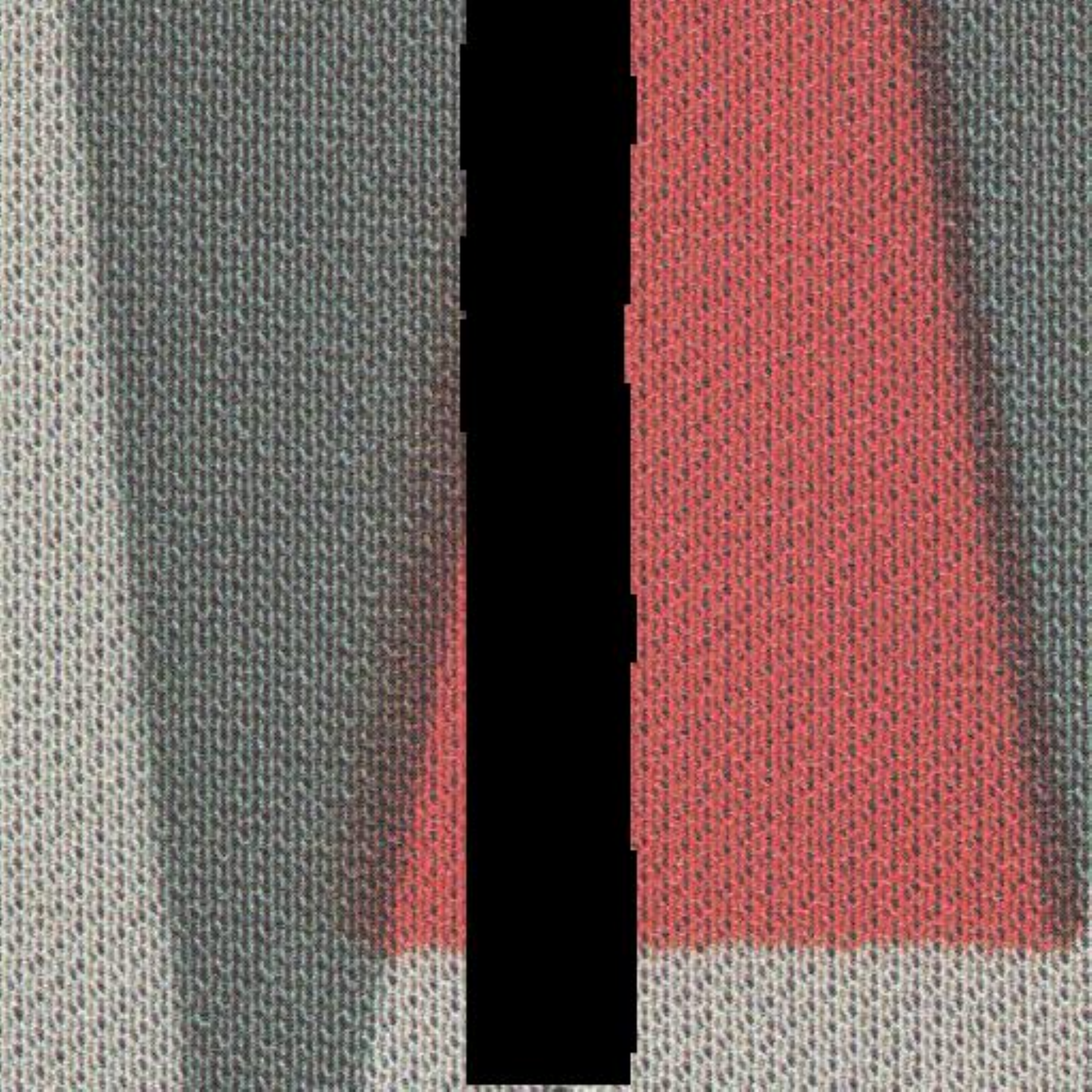}} &
		\subfloat{\includegraphics[width=0.15\textwidth, scale=0.7]{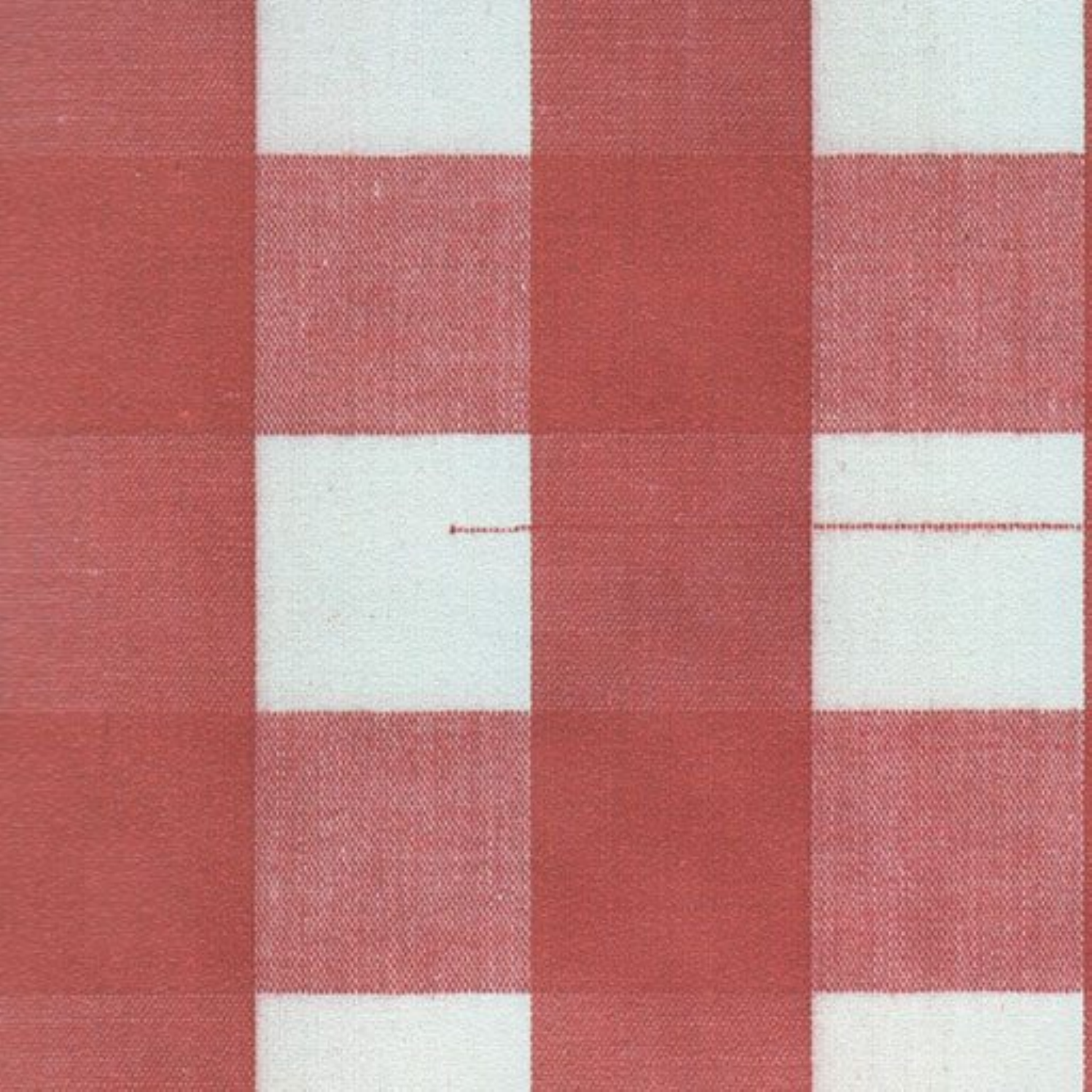}} &
		\subfloat{\includegraphics[width=0.15\textwidth, scale=0.7]{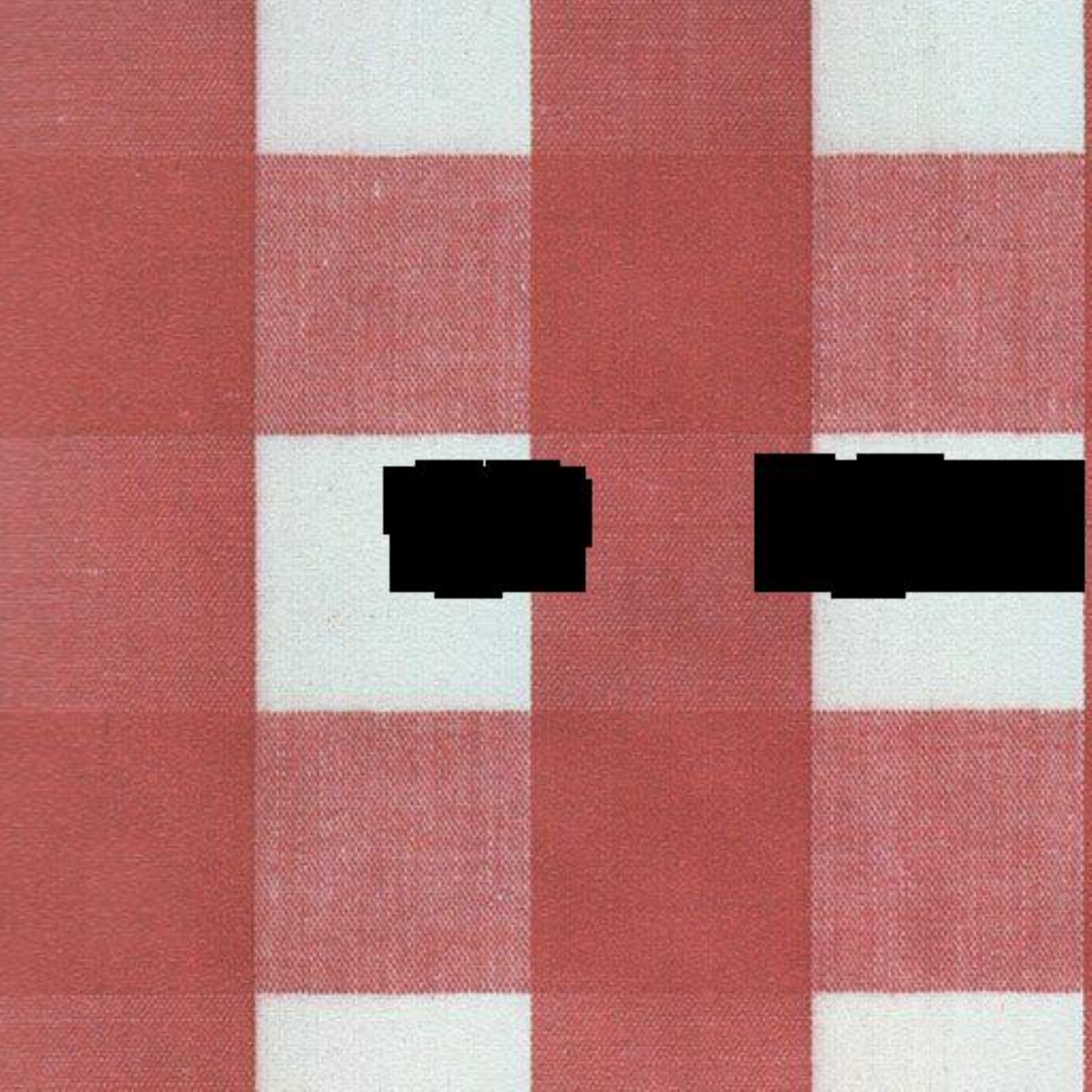}} &
		\subfloat{\includegraphics[width=0.15\textwidth, scale=0.7]{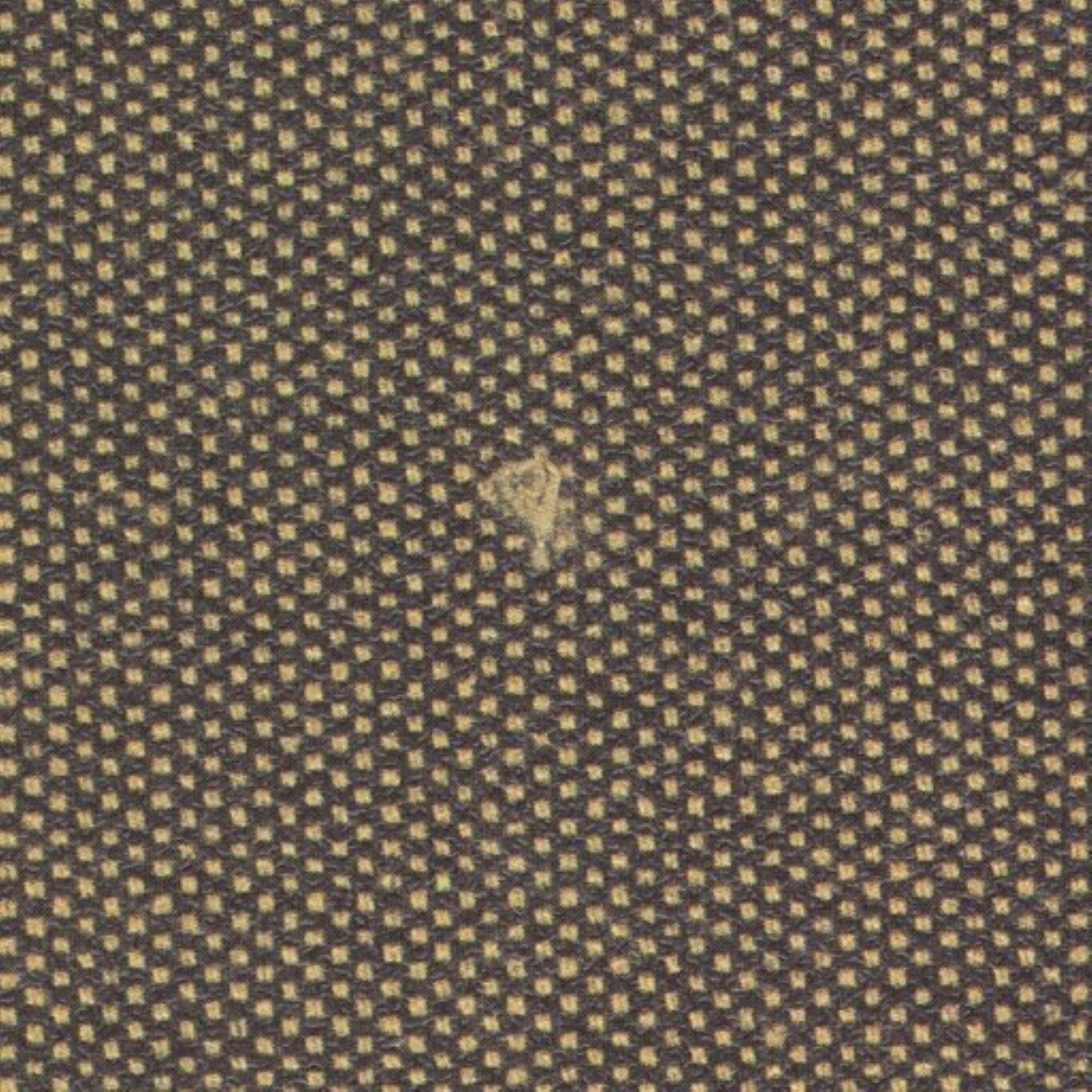}} &
		\subfloat{\includegraphics[width=0.15\textwidth, scale=0.7]{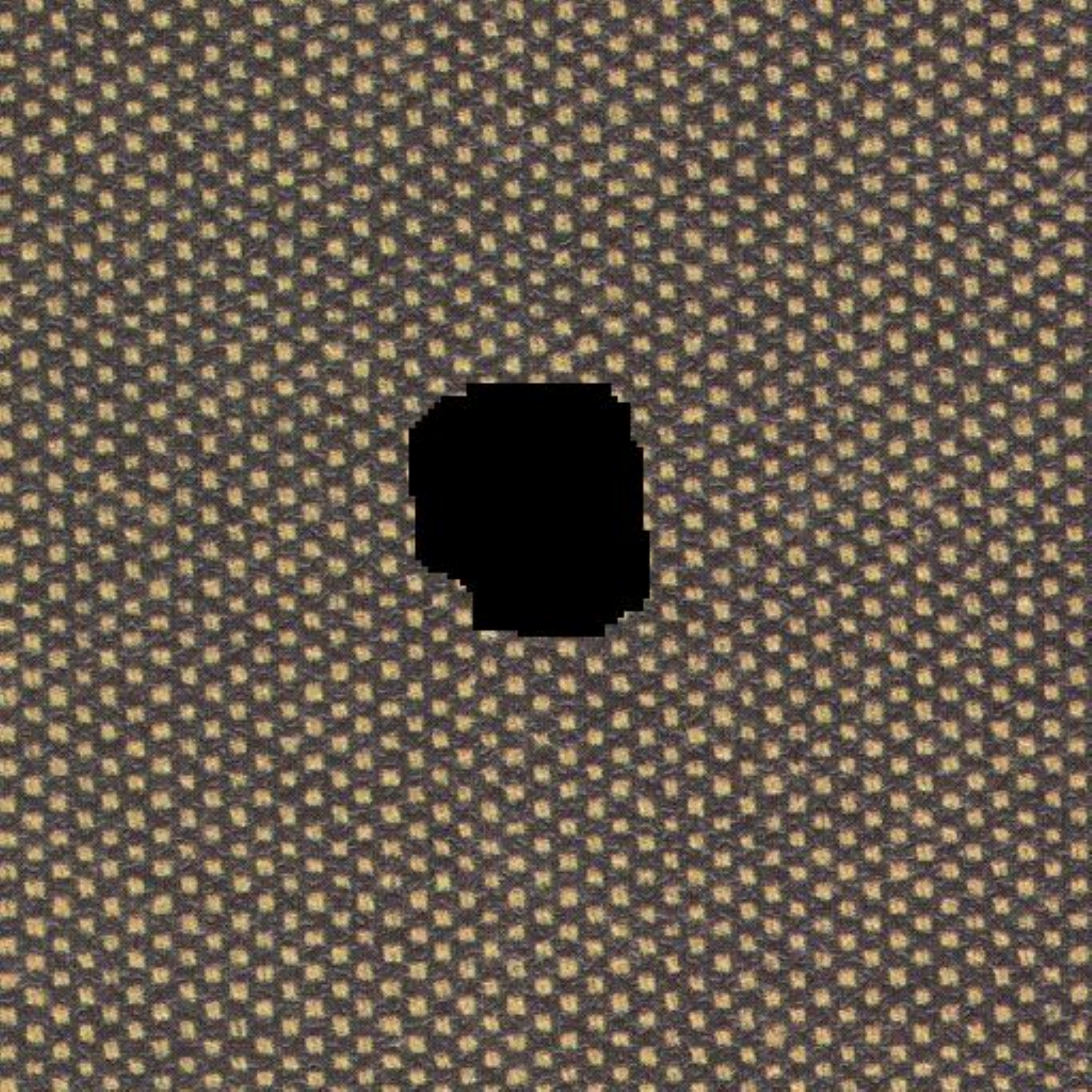}} \\

		Input image & Predicted image & Input image & Predicted image & Input image & Predicted image
	\end{tabular}
	\caption{Visual outputs of patterned fabric defect detection (1st, 3rd, 5th columns present input images and 2nd, 4th, 6th columns show defects detected in black).}
	\label{fig:pattern}
\end{figure*}

We visualize the defect detection results of plain and patterned fabric images selected from the dataset in Fig.~\ref{fig:plain} and Fig.~\ref{fig:pattern} respectively. Our proposed one-class model successfully detects all defects successfully regardless of different fabric textures and shapes/sizes of defects. Fig.~\ref{fig:pattern} shows our model works for patterned fabric as well as plain fabric. We analyze the reasons as follows. First, our Gabor filter bank is capable of extract texture features from different dimensions, which makes the model less insensitive to the changes in motif size. Second, our feature learning autoencoder makes the features extracted from the Gabor filter bank outputs more versatile. It makes the model effective in locating different kinds of defects.

Note that the defects detected are slightly larger in size than actual defects. This is because our proposed model predicts defects on the basis of image patches instead of pixels. Therefore, as the patches are larger in size, the defects marked tend to get larger. We don't see this level of discrepancy as a problem in real world as defects would be cut off with some margins for safe quality control when found.

\subsection{Comparison of different feature learning techniques}
We compare the performance of our proposed autoencoder with the other two traditional feature selection methods, the hand-crafted~\cite{chetverikov2002finding} and PCA~\cite{beirao2004defect}, to demonstrate the effectiveness of our proposed autoencoder. Regardless of the kinds of fabrics, our model achieves a higher value of cTPR and AUC (area under ROC curve) as shown in Table~\ref{table:feature}. Specifically, our autoencoder improves overall cTPR and AUC by 12.9\% and 2.9\% respectively. This suggests that by utilizing an autoencoder, our model can learn more representative and general features from the magnitude responses of the designed Gabor filter bank. The defective feature vectors, therefore, are more easily detected by the nearest neighbor density estimator.


One may notice that AUCs don't change much. Indeed, AUC measures the performance of the model across all different FPR values. Stable AUCs suggest that our model improves cTPR a lot by allowing some positive samples to be predicted as negative samples (FPR only increases a bit). This also implies some defects are very close to defective fabrics, making them very difficult to classify correctly (e.g., the fabric located at the $2^{nd}$ row and the $5^{th}$ column in Fig.~\ref{fig:plain}). Our proposed model can still predict defects very well on all kinds of fabrics when there are no any false alarms arise.


	

\begin{table}[h!]
	\caption{Comparison between different feature selections. Note that the numbers are averaged over all cTPRs and AUCs of 31 fabric images respectively.}

	\centering
	\begin{tabular}{|c c c c|} 
		\hline
		& Hand-crafted~\cite{chetverikov2002finding} & PCA~\cite{beirao2004defect} &  Ours \\
		\hline
		cTPR for plain fabrics   & 0.807 & 0.803 & \textbf{0.907} \\
		AUC for plain fabrics   & 0.953 & 0.951 & \textbf{0.980}\\
		cTPR for patterned fabrics & 0.779 & 0.774 & \textbf{0.882}\\
		AUC for patterned fabrics & 0.945 & 0.944 & \textbf{0.974}\\ 
		cTPR for all fabrics     & 0.793 & 0.788 & \textbf{0.895}\\
		AUC for all fabrics     & 0.949 & 0.947 & \textbf{0.977}\\
		\hline
	\end{tabular}
	\label{table:feature}
\end{table}

\subsection{Comparison against deep learning algorithms}
Besides traditional feature learning methods, we also compare our autoencoder with broadly-used residual neural networks (ResNet)~\cite{he2016deep}. By end-to-end training a ResNet and a binary classifier using all 31 fabrics in our dataset, the ResNet model is able to perform fabric defect detection on all 31 fabric with a 93.5\% recall and an 83.6\% precision. However, when coming to cTPR (i.e., no false alarms exist), the cTPR of the ResNet model is only about 0.217, which is far worse than ours (0.895). We conclude three reasons for it. Firstly, it is no doubt that the ResNet and its variants have gained lots of success in all kinds of computer vision tasks due to the powerful representation learning. It is also a fact that the bedrock for these successes is datasets where the number of samples and accurate labels are superior to our dataset (hundreds of thousands vs. 31). The quality of representation learning, thus, cannot be ensured. For example, unlike our Gabor filter bank and simple autoencoder, the convolutional filters of ResNet are learnable and these large number of parameters are unlikely to learn well without sufficient samples. Secondly, the sizes of defects are usually small compared with 512$\times$512 images. Without having special attention to defects, ResNet models could just learn the representations of fabrics instead of defects. Also, defects have different types and to our best knowledge, there is no concrete number that says how many types of fabric defects exist. This makes training a deep learning classifier hard since current models are still learning from existing data and cannot infer well outside of the scope. However, our one-class model shows its strength by only learning from fabrics patterns. Lastly, the number of normal samples is way larger than that of defective samples, a unbalanced dataset makes training even harder as trained models can be misleading by just predicting all samples to be normal, which is also the main reason the cTPR is low.

\begin{figure*}
	\centering
	\begin{tabular}{c c c c c c}
		
		\subfloat{\includegraphics[width=0.15\textwidth, scale=0.7]{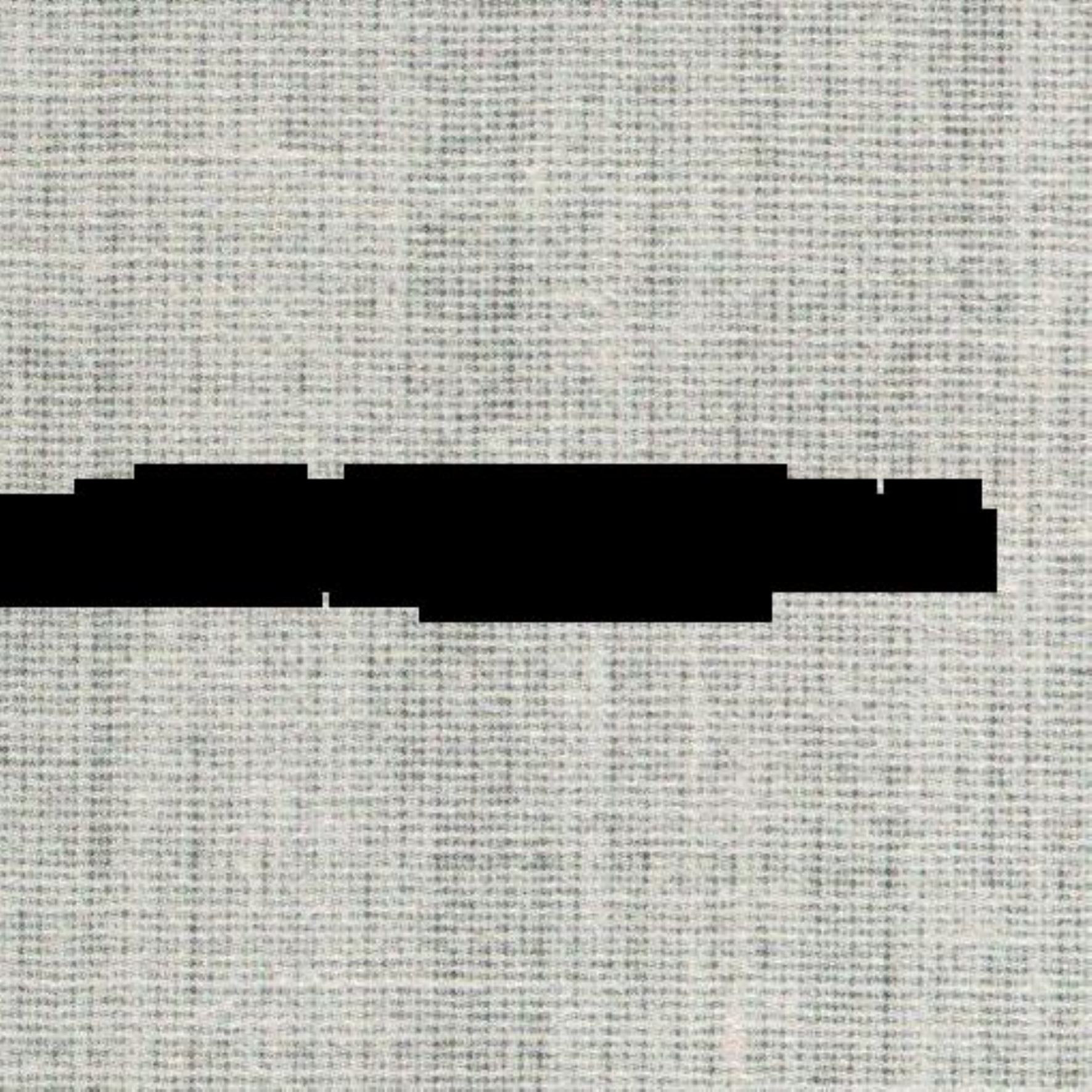}} &
		\subfloat{\includegraphics[width=0.15\textwidth, scale=0.7]{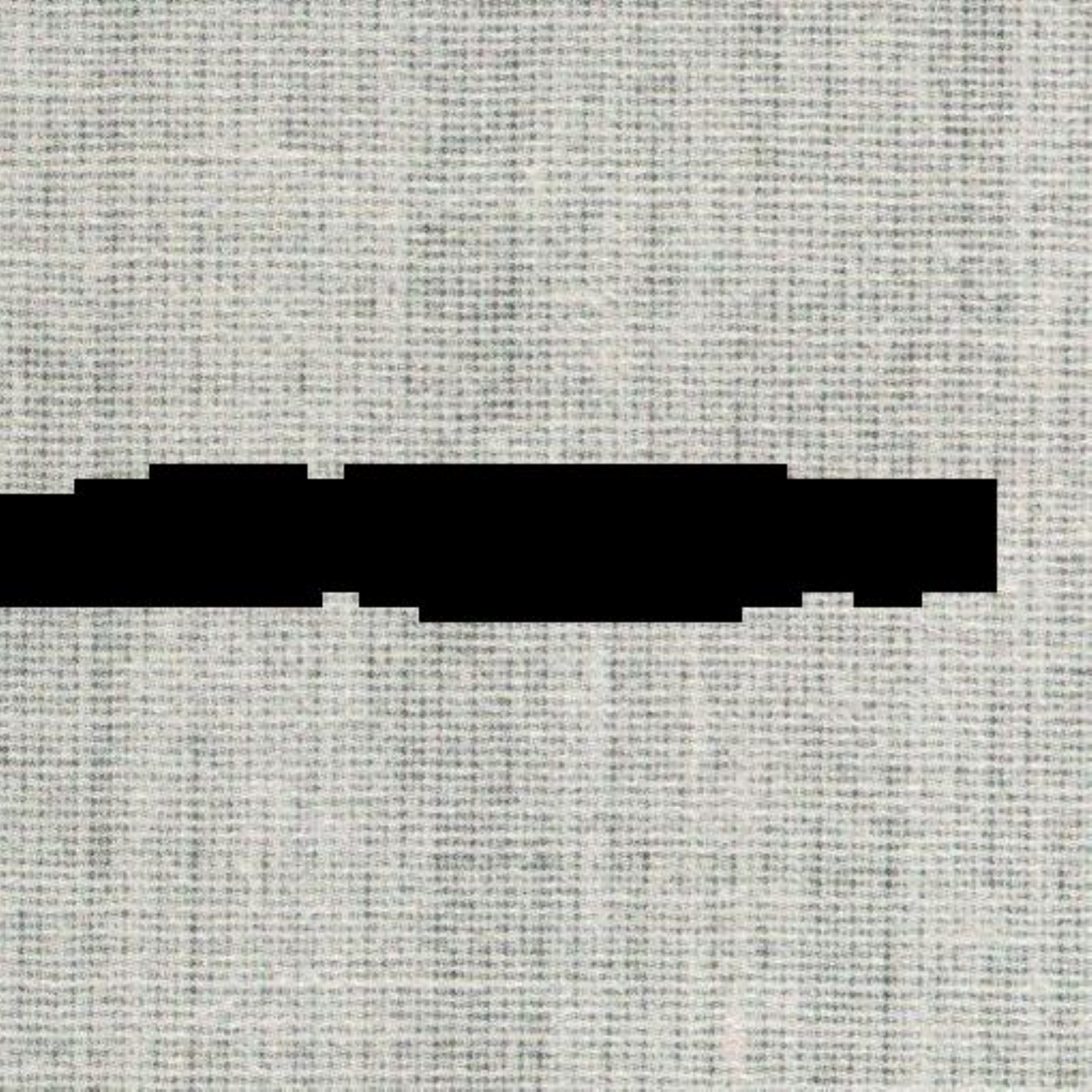}} &
		\subfloat{\includegraphics[width=0.15\textwidth, scale=0.7]{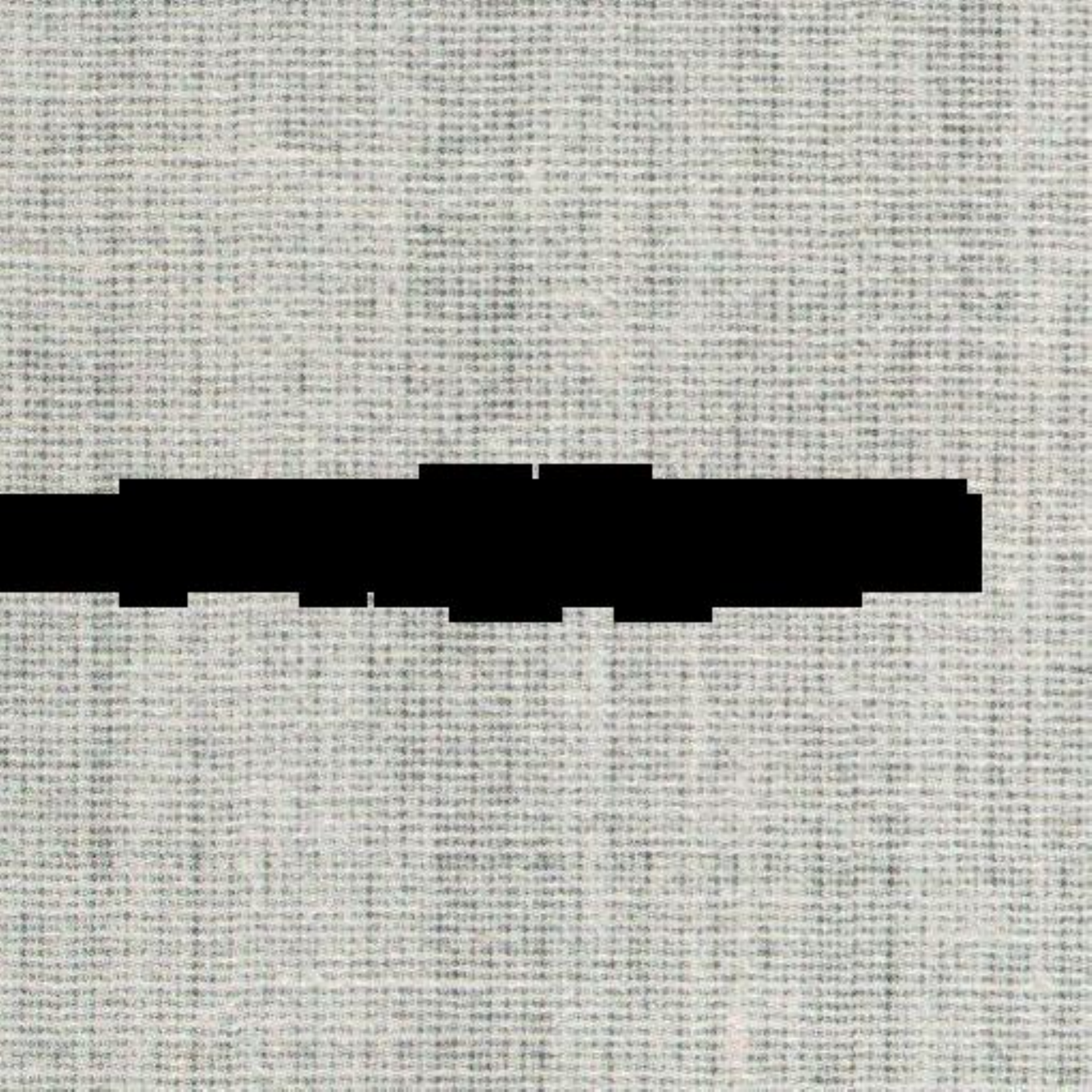}} &
		\subfloat{\includegraphics[width=0.15\textwidth, scale=0.7]{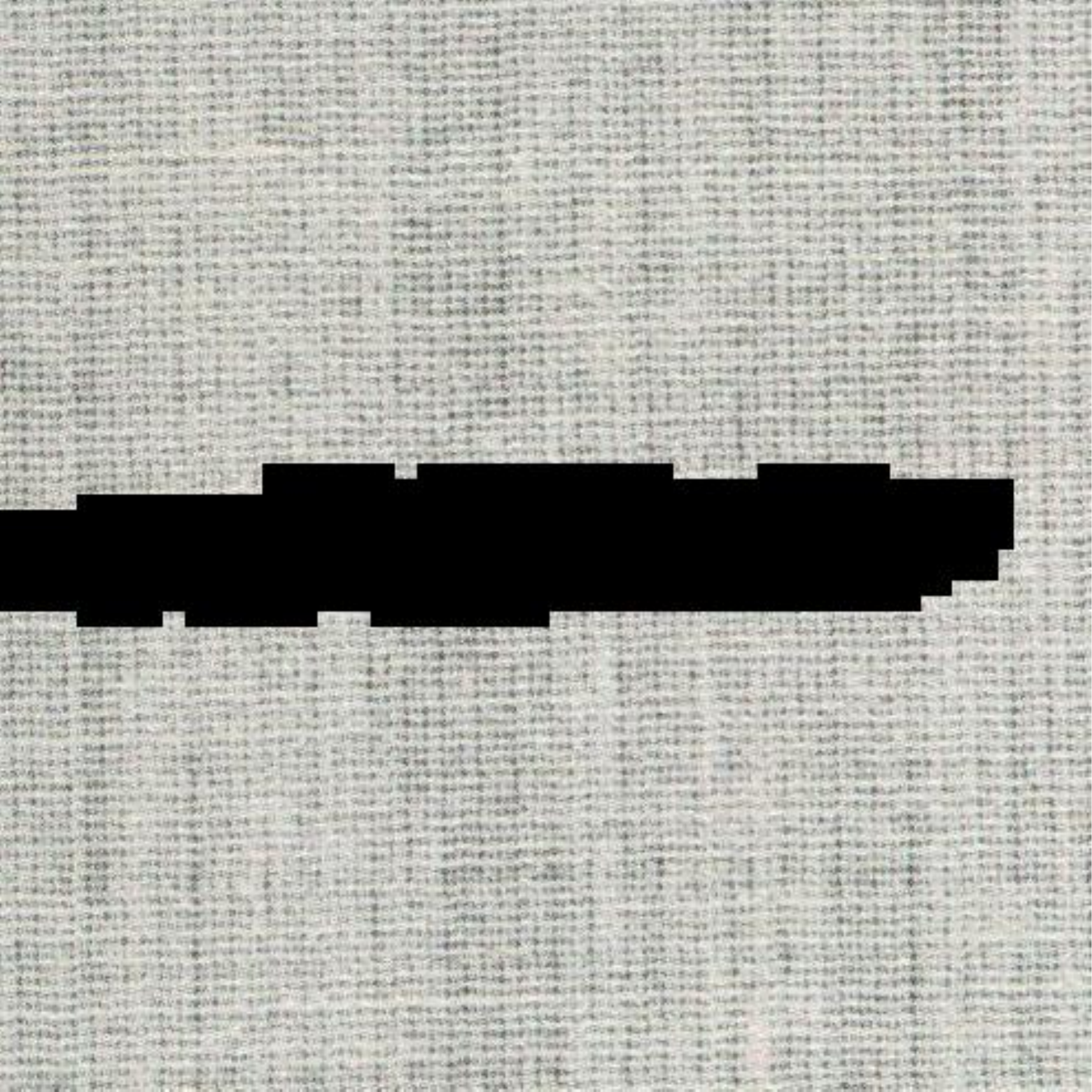}} &
		\subfloat{\includegraphics[width=0.15\textwidth, scale=0.7]{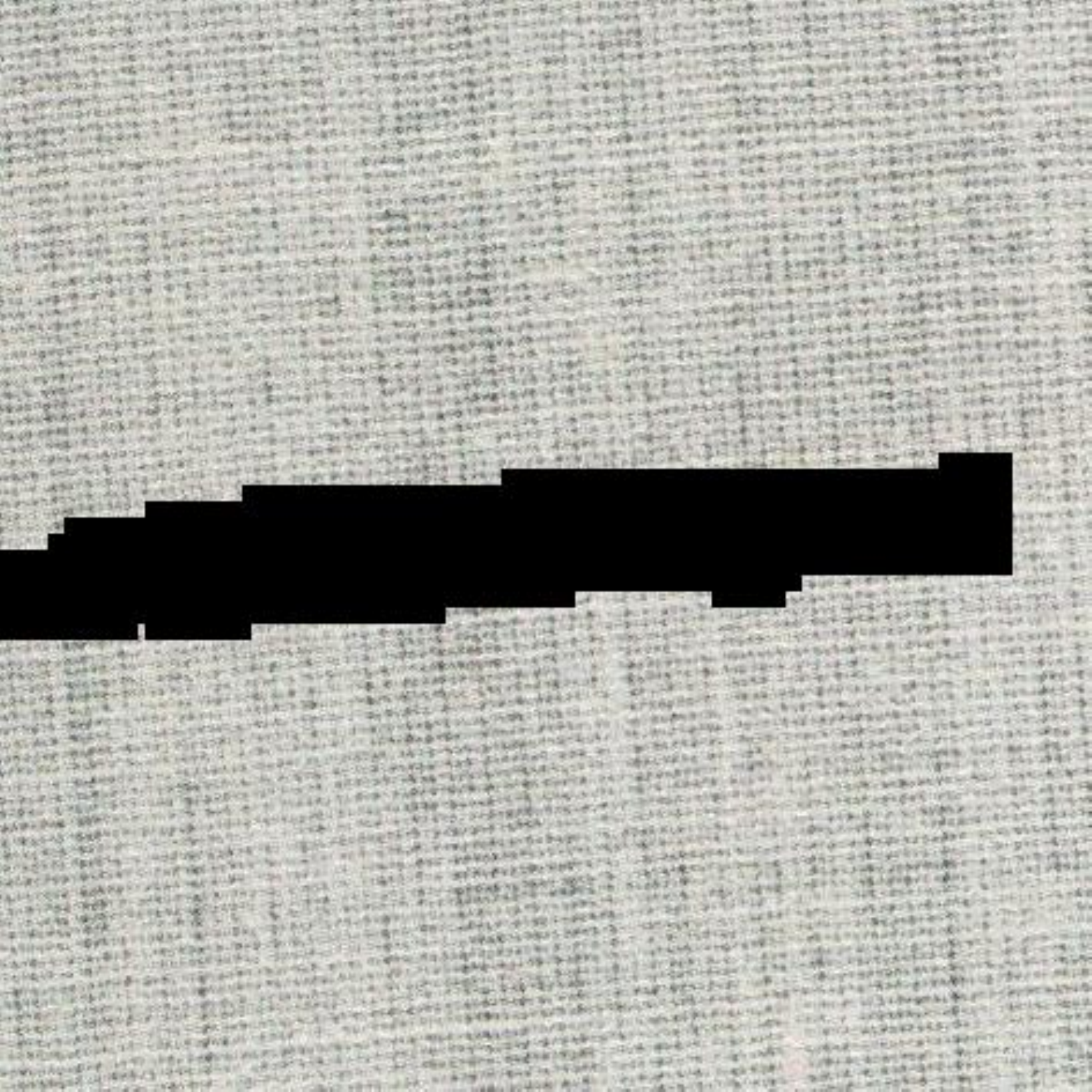}} \\


		\subfloat{\includegraphics[width=0.15\textwidth, scale=0.7]{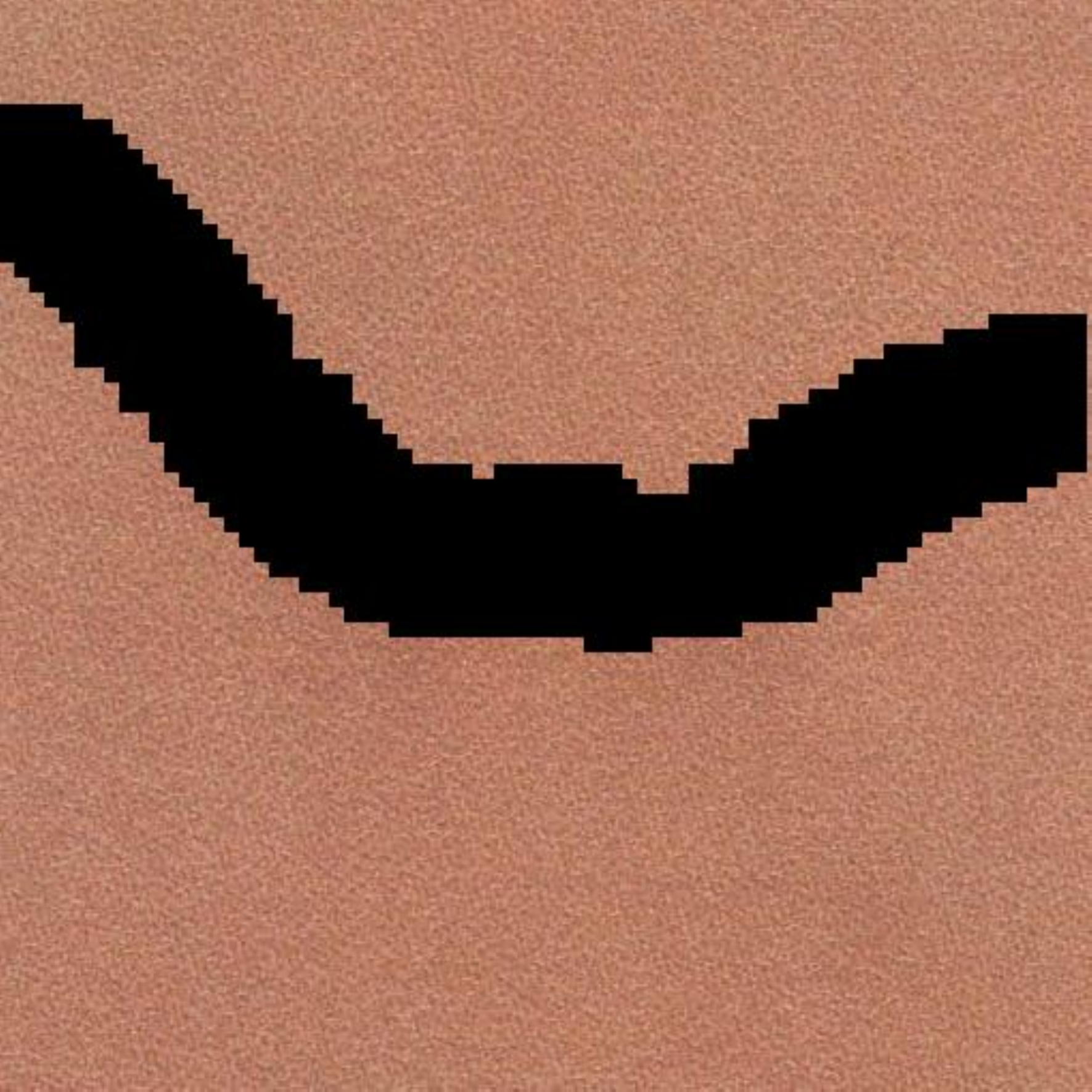}} &
		\subfloat{\includegraphics[width=0.15\textwidth, scale=0.7]{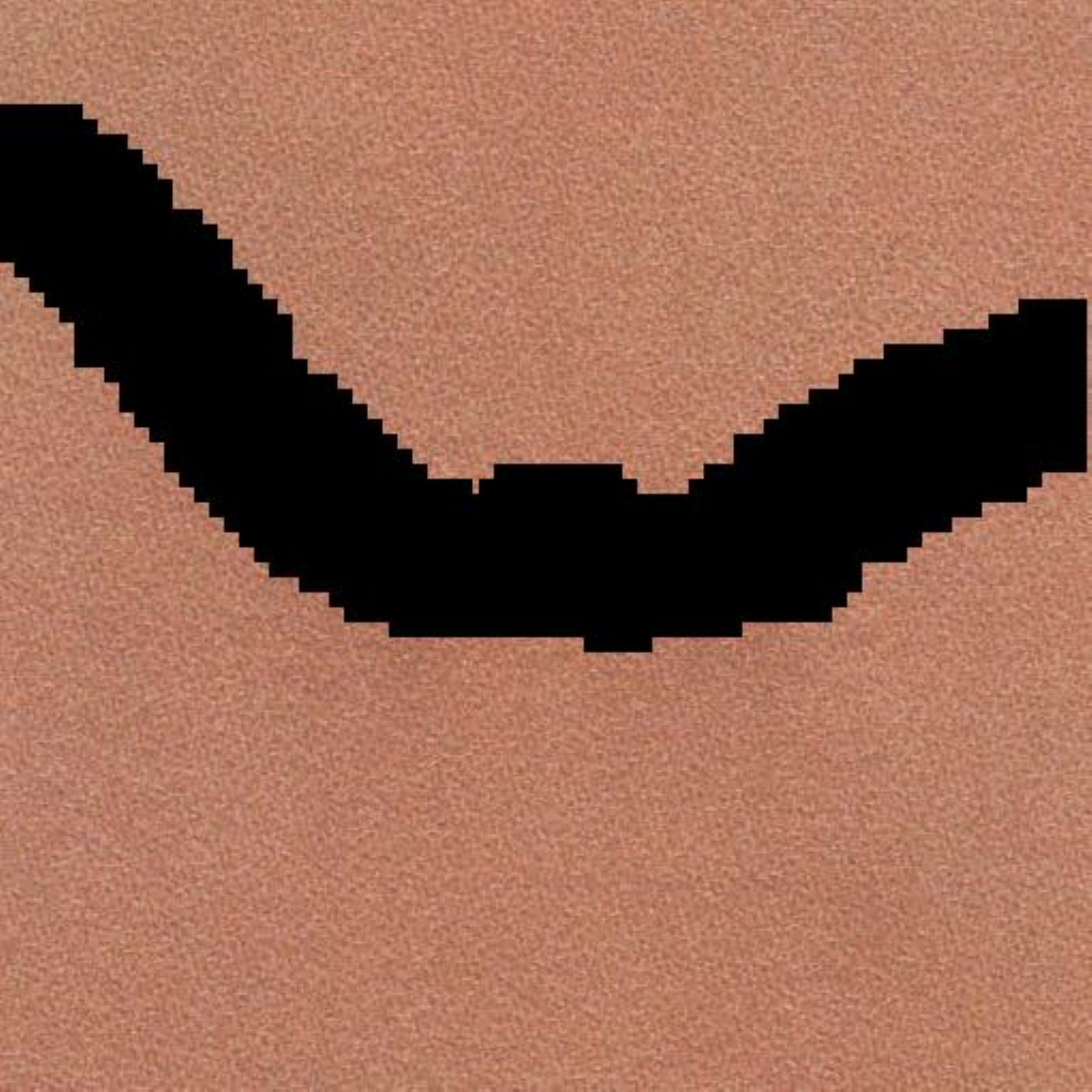}} &
		\subfloat{\includegraphics[width=0.15\textwidth, scale=0.7]{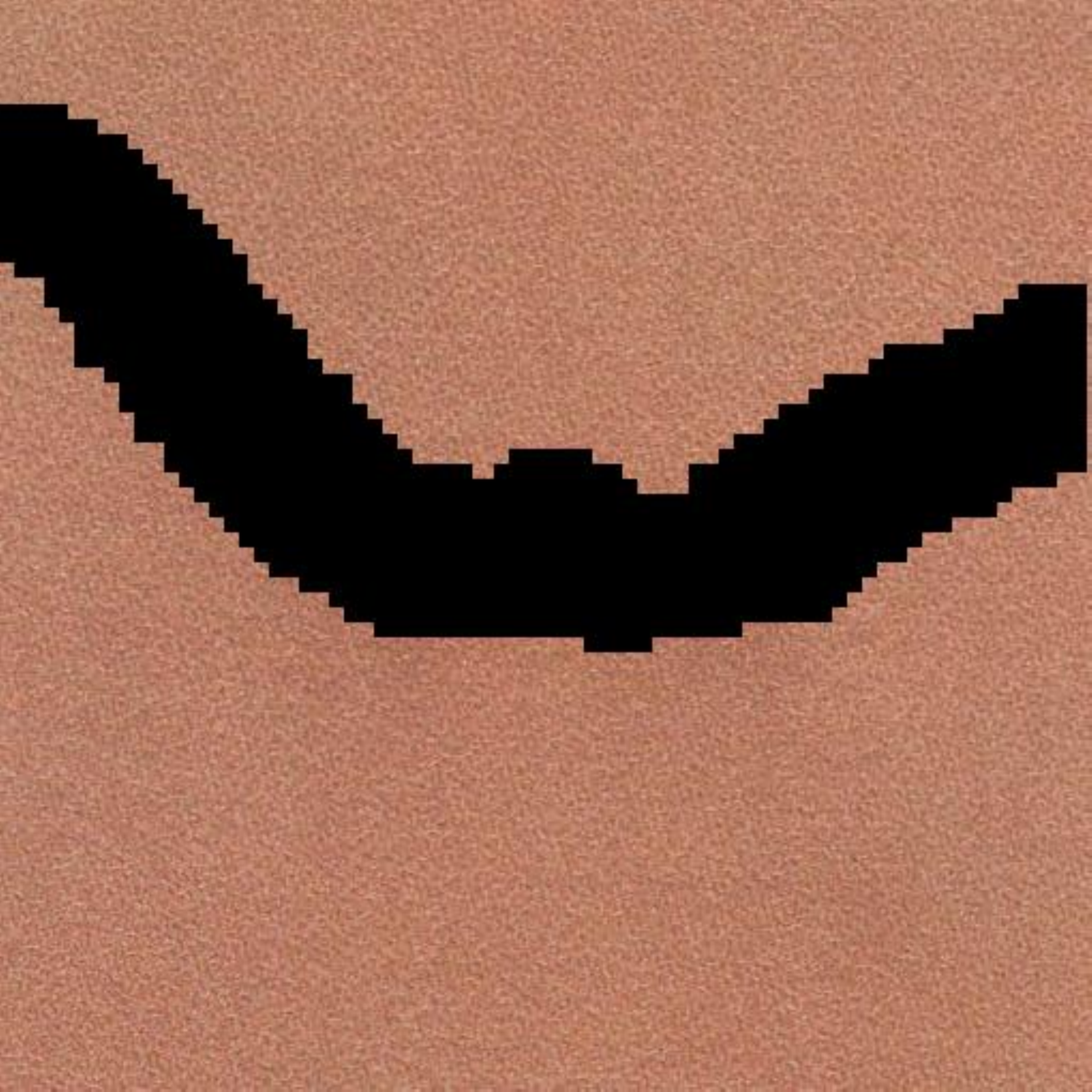}} &
		\subfloat{\includegraphics[width=0.15\textwidth, scale=0.7]{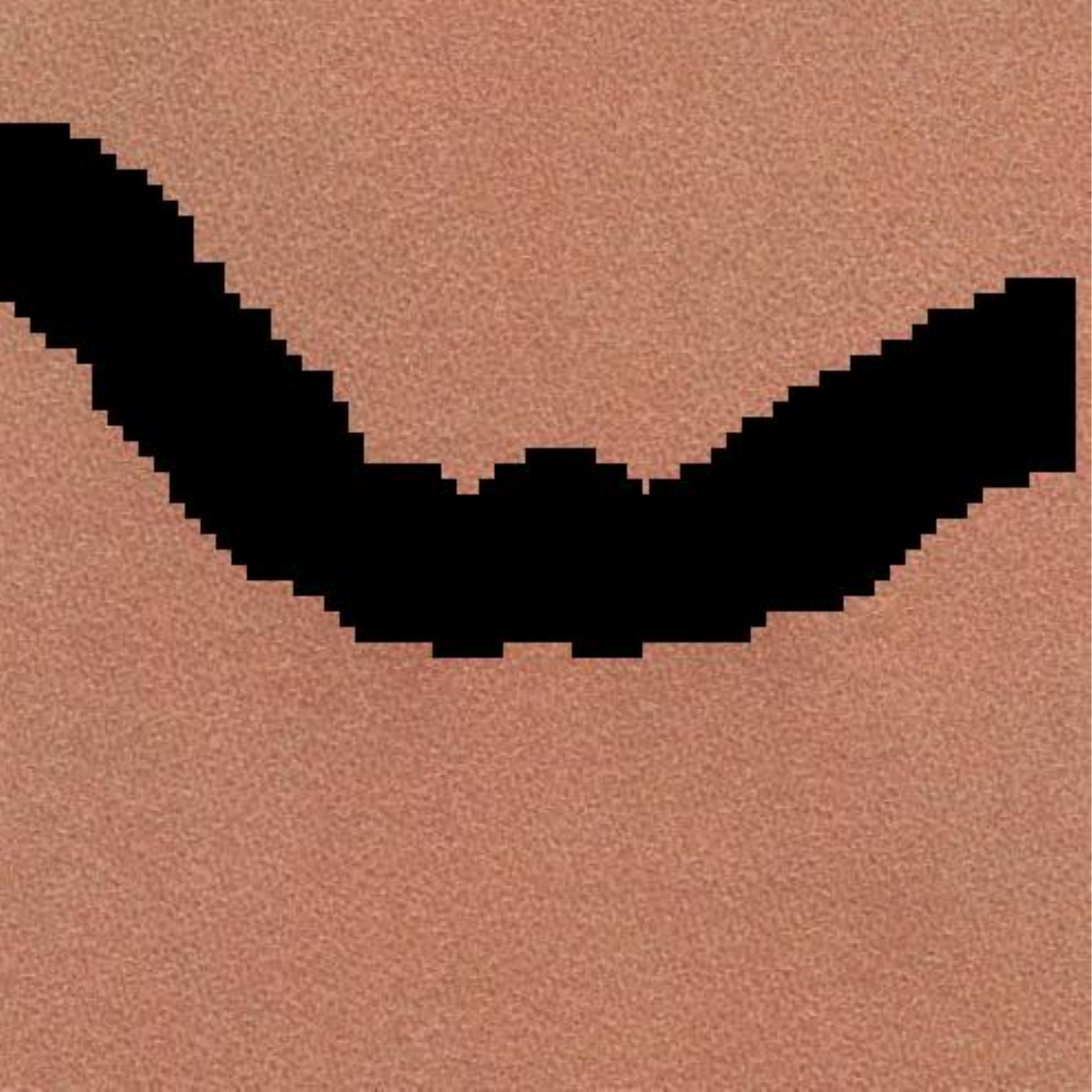}} &
		\subfloat{\includegraphics[width=0.15\textwidth, scale=0.7]{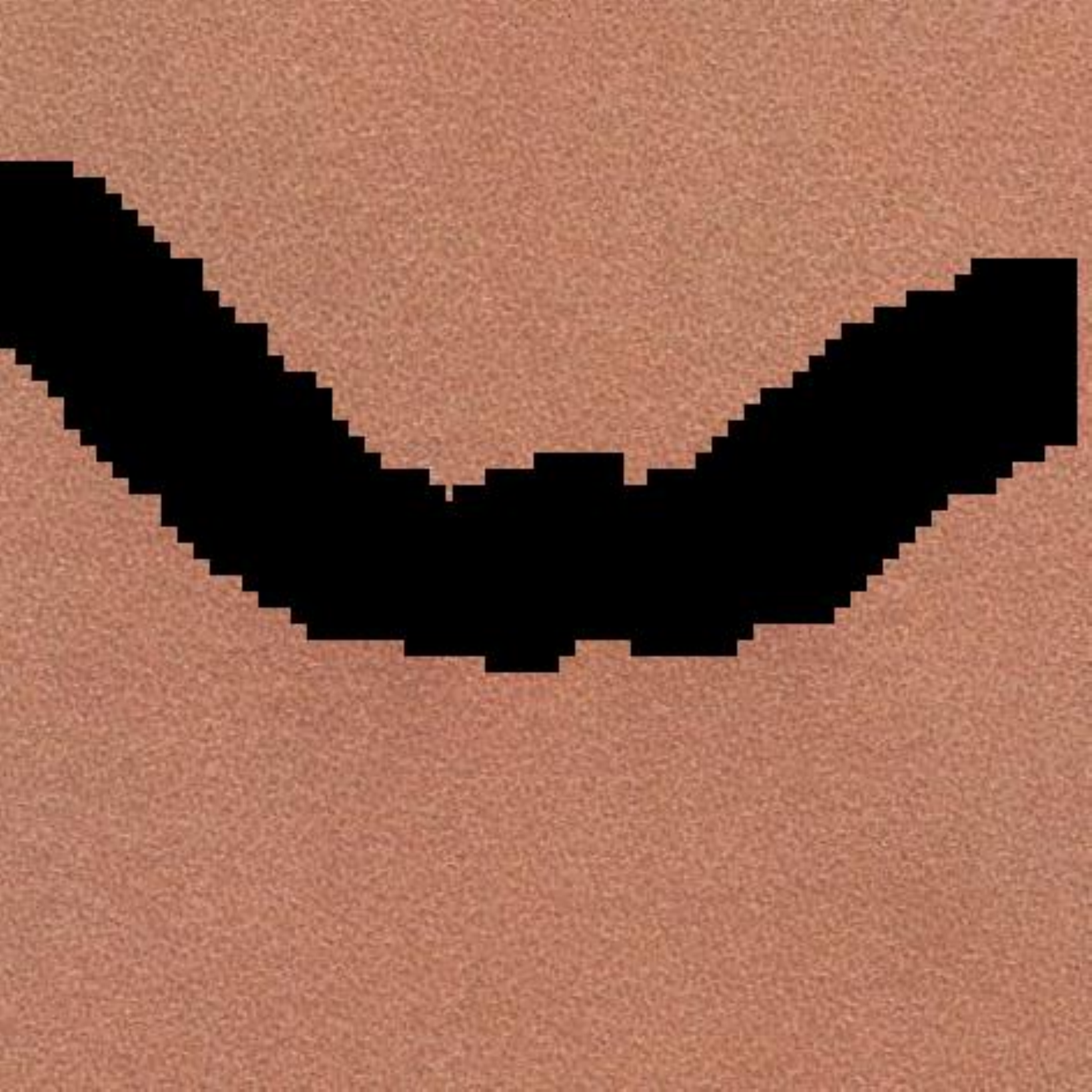}} \\	
		
		\subfloat{\includegraphics[width=0.15\textwidth, scale=0.7]{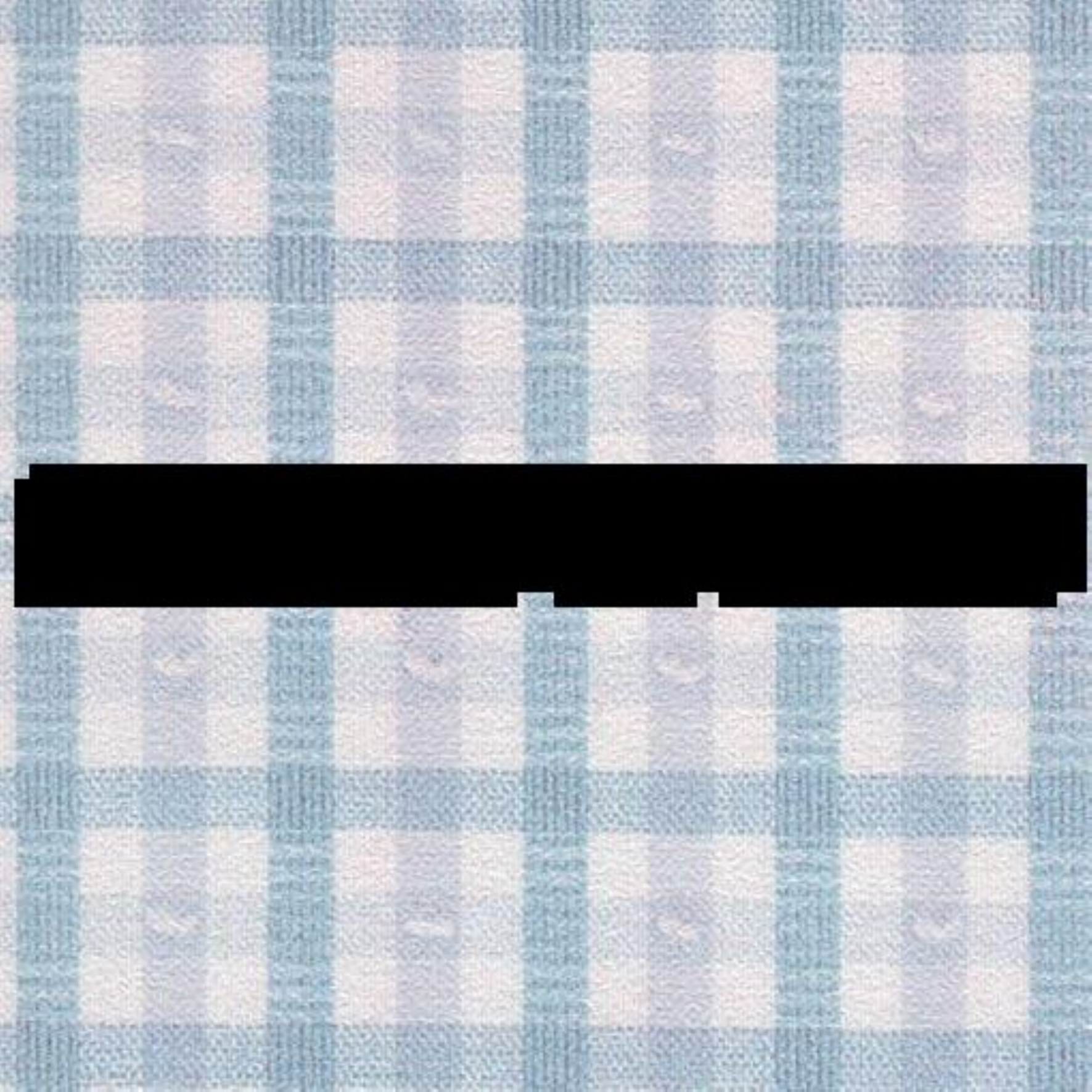}} &
		\subfloat{\includegraphics[width=0.15\textwidth, scale=0.7]{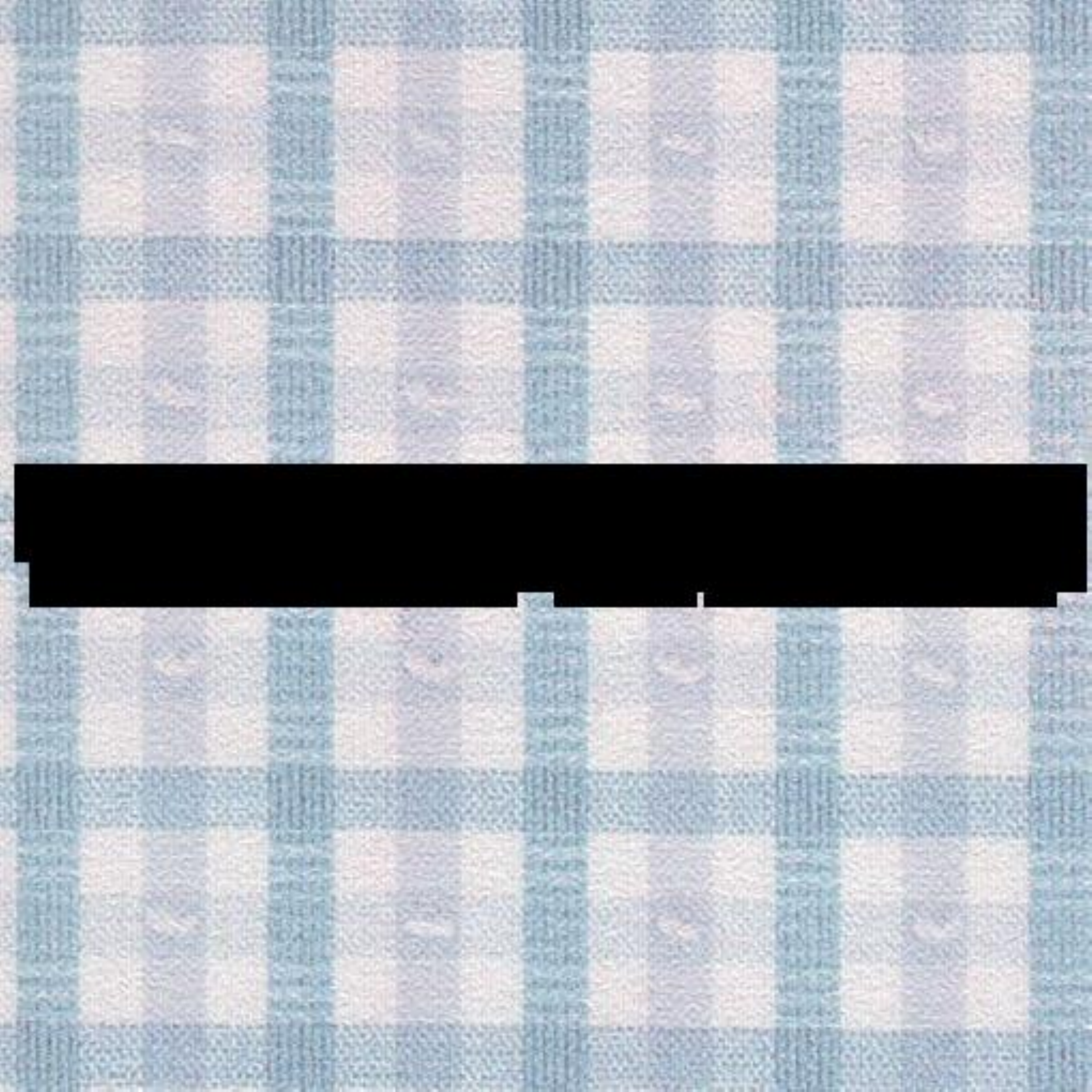}} &
		\subfloat{\includegraphics[width=0.15\textwidth, scale=0.7]{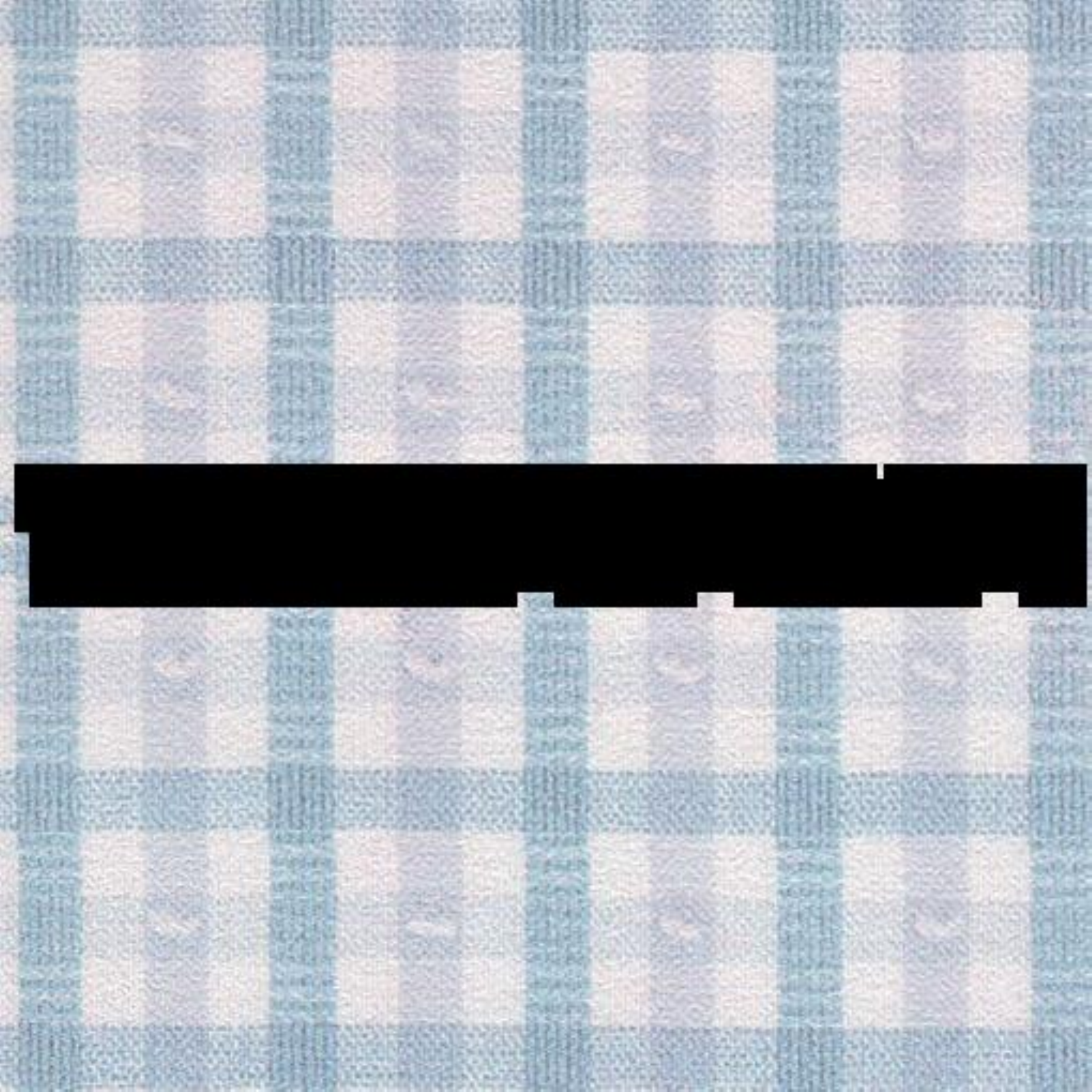}} &
		\subfloat{\includegraphics[width=0.15\textwidth, scale=0.7]{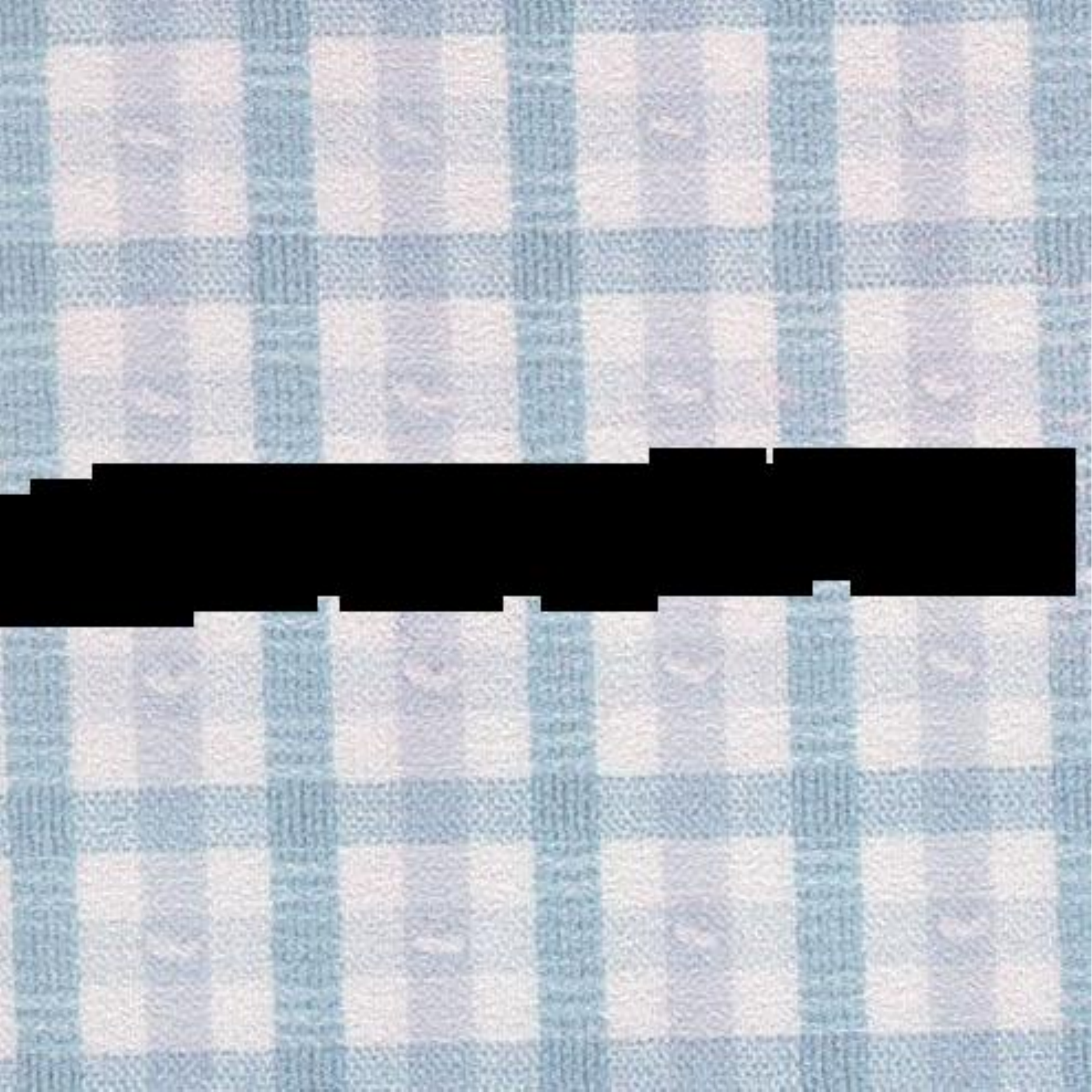}} &
		\subfloat{\includegraphics[width=0.15\textwidth, scale=0.7]{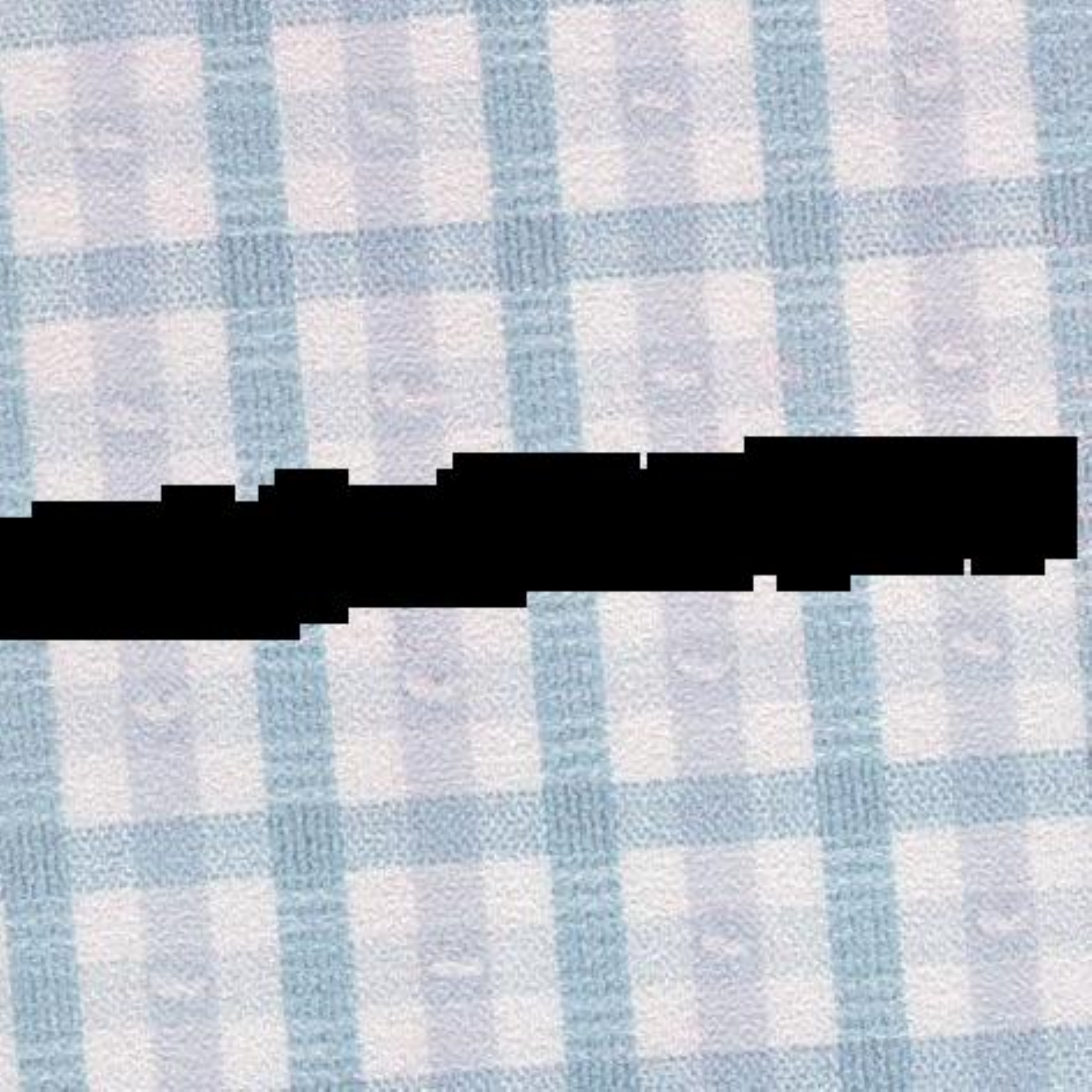}} \\

		\subfloat{\includegraphics[width=0.15\textwidth, scale=0.7]{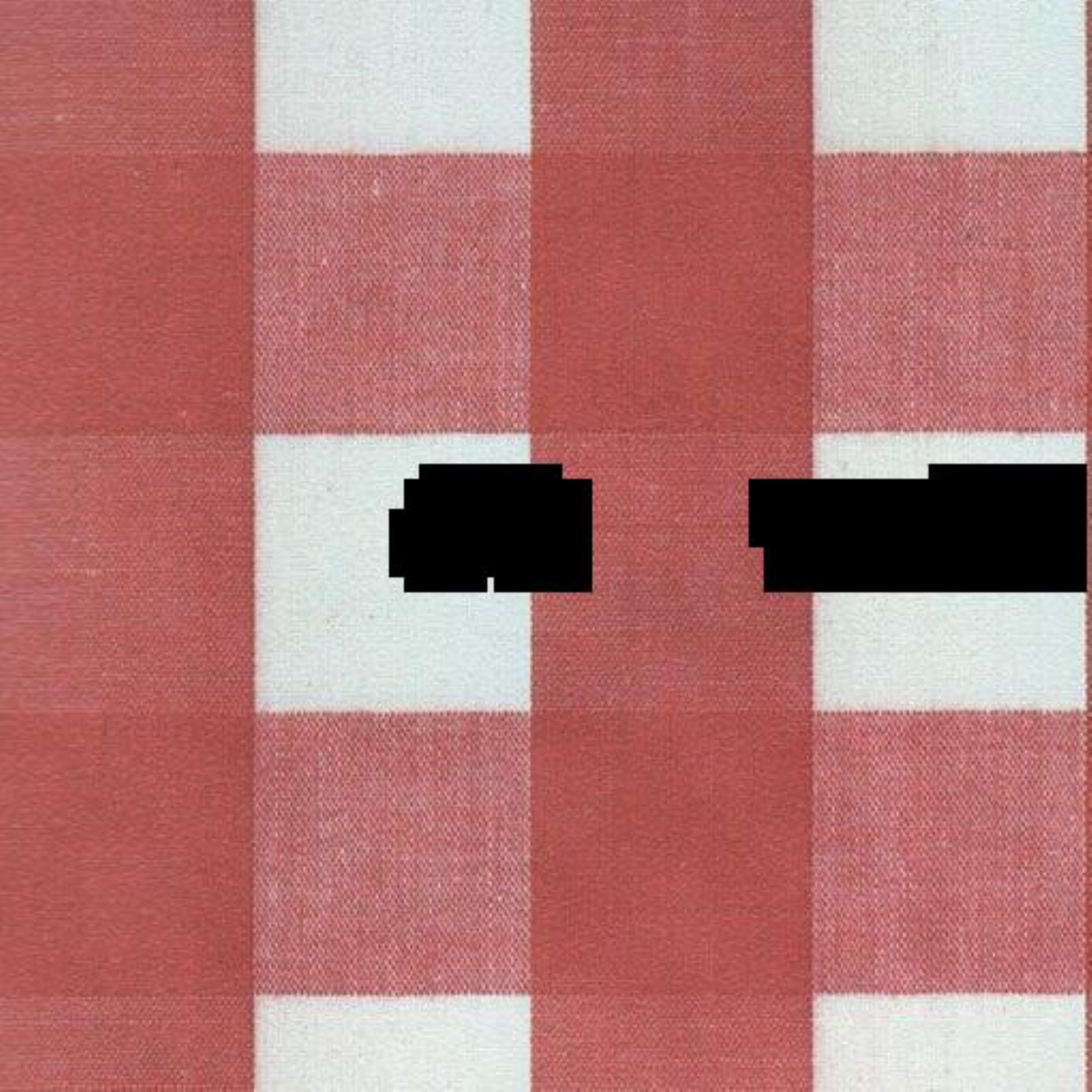}} &
		\subfloat{\includegraphics[width=0.15\textwidth, scale=0.7]{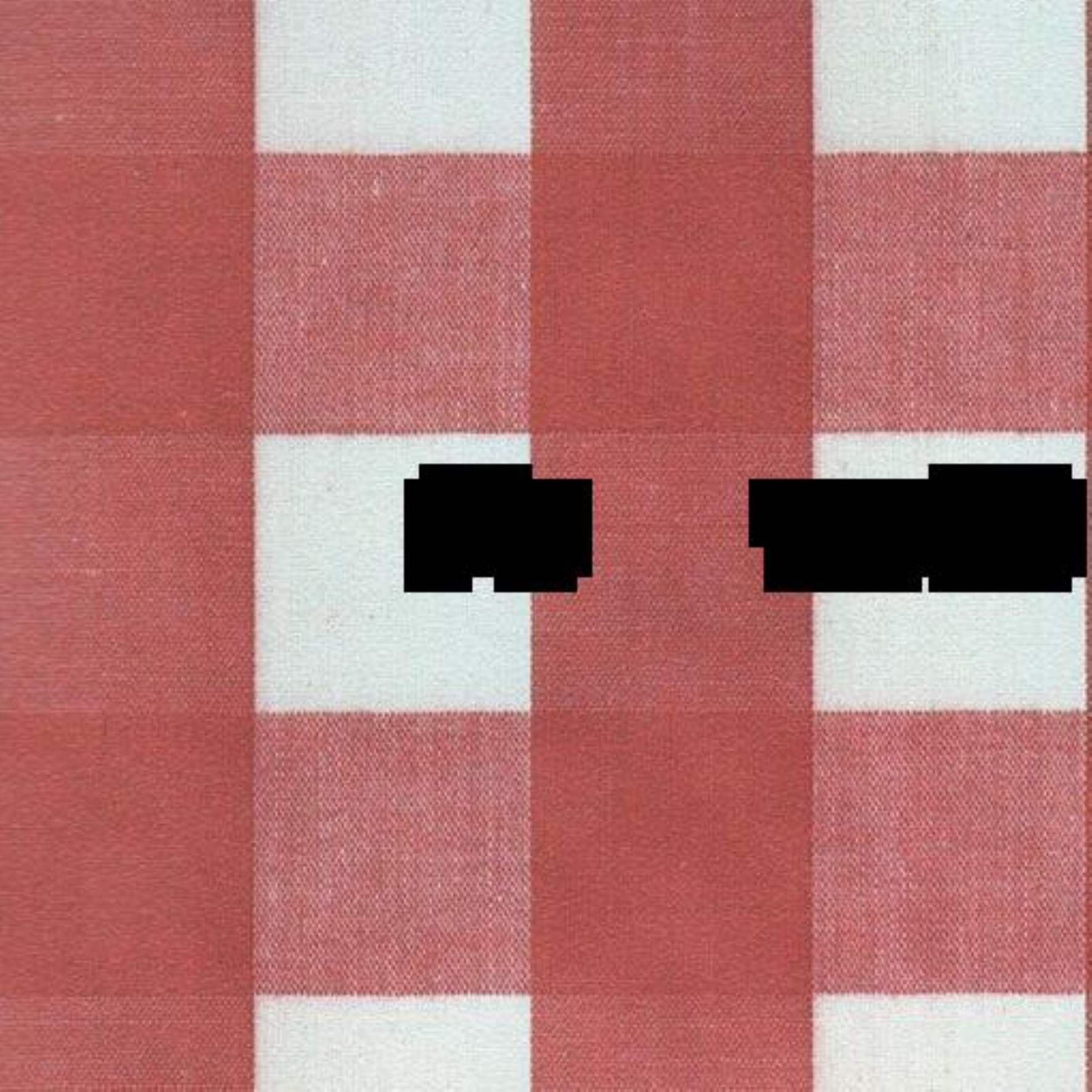}} &
		\subfloat{\includegraphics[width=0.15\textwidth, scale=0.7]{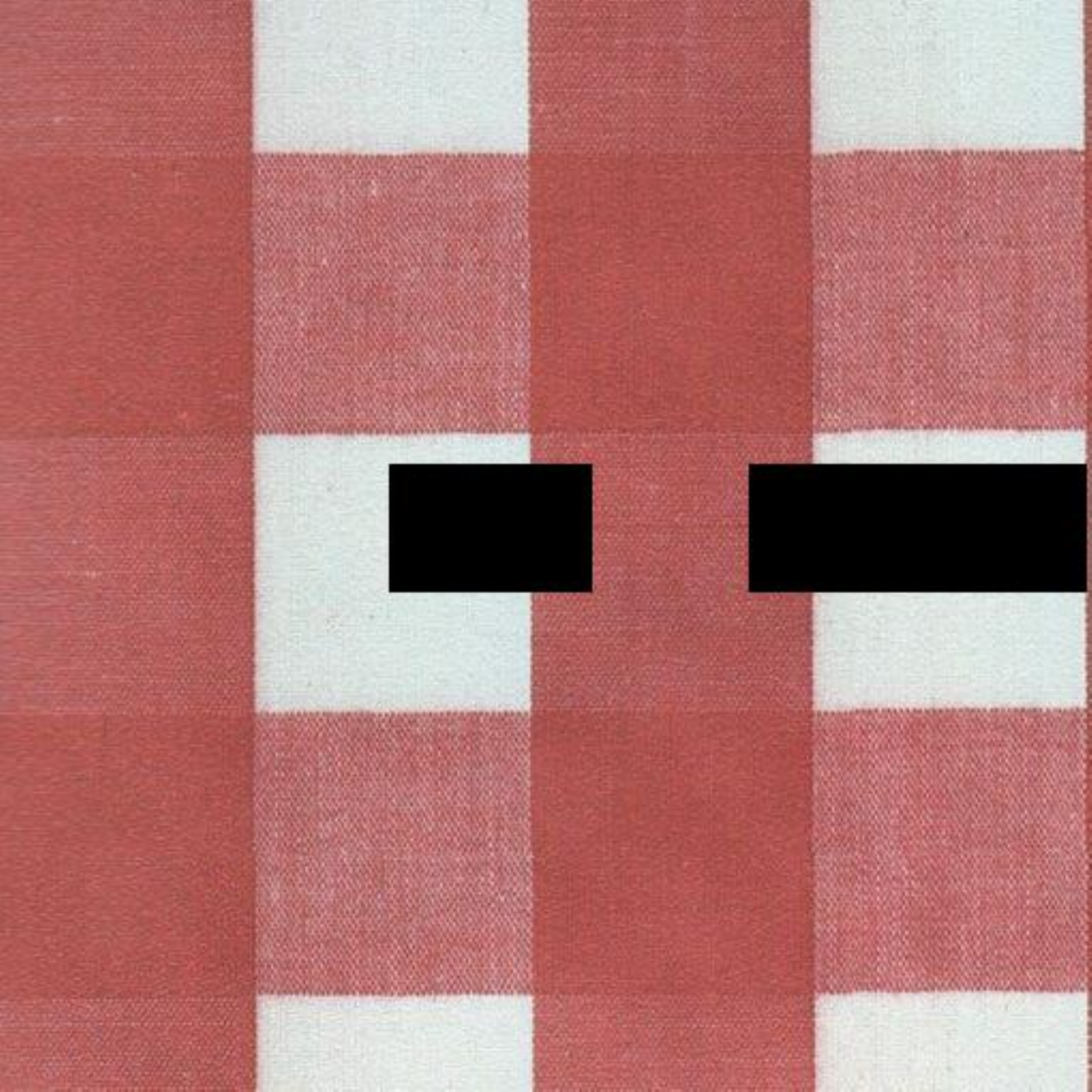}} &
		\subfloat{\includegraphics[width=0.15\textwidth, scale=0.7]{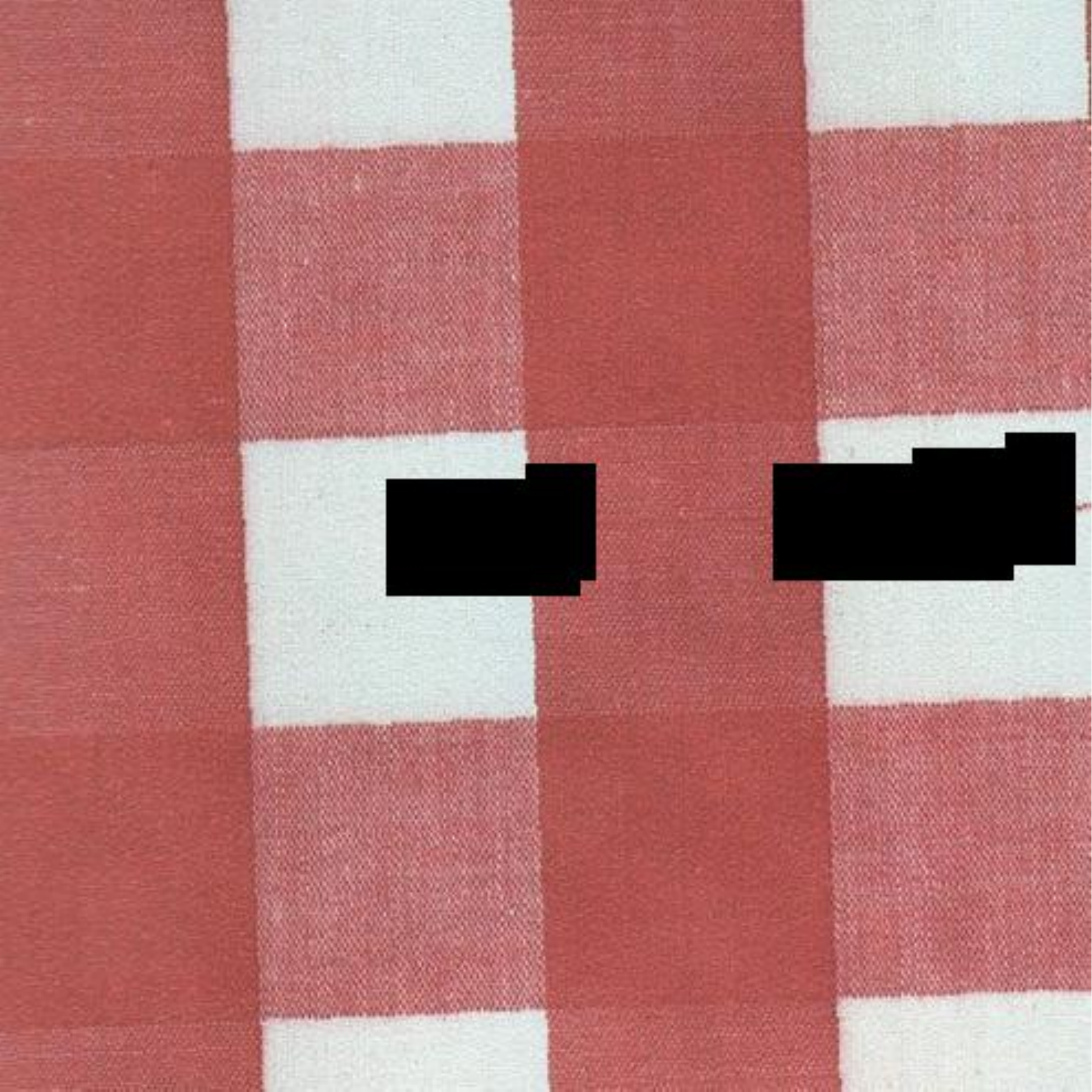}} &
		\subfloat{\includegraphics[width=0.15\textwidth, scale=0.7]{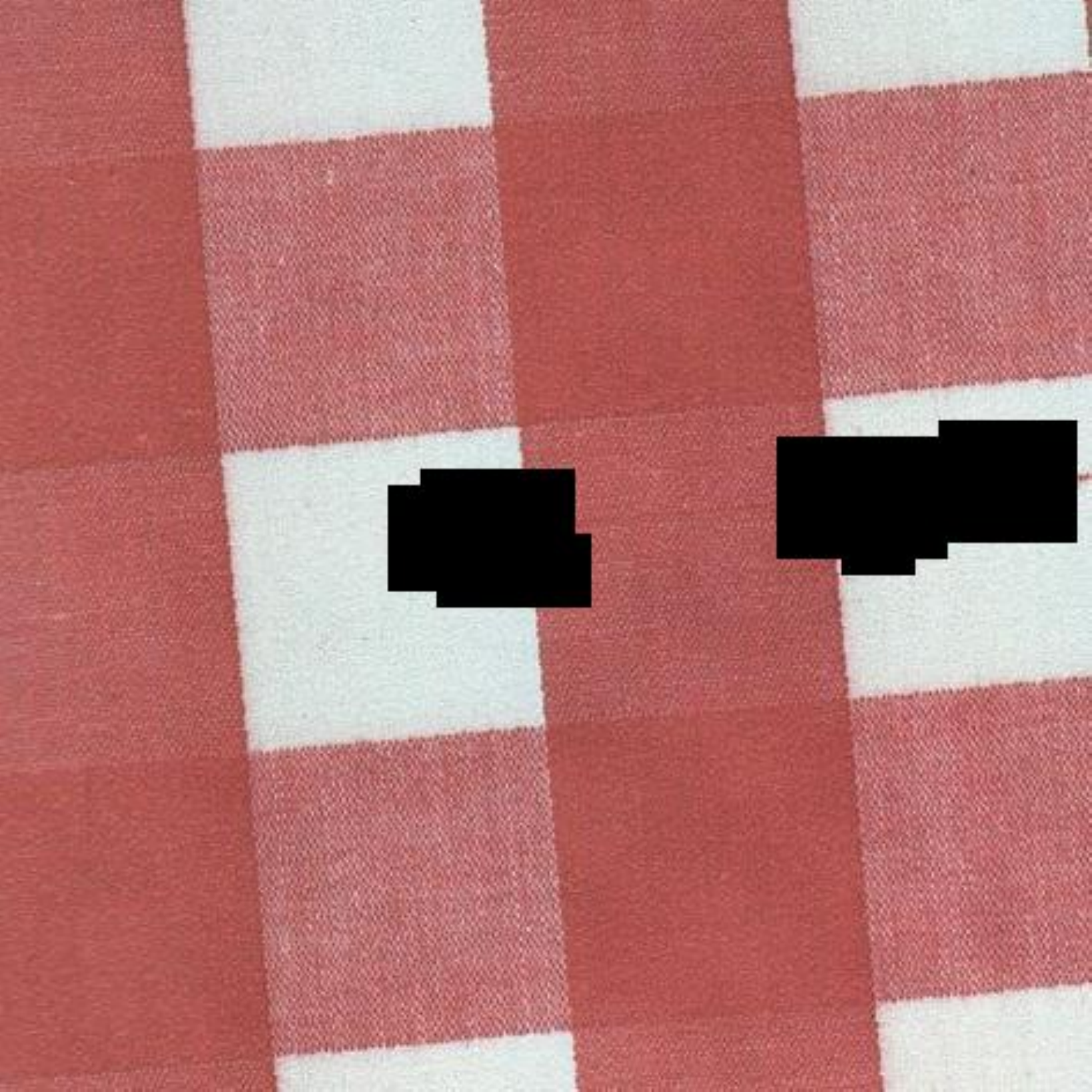}} \\


		0.02$^{\circ}$ Rotated & 0.05$^{\circ}$ Rotated & 0.2$^{\circ}$ Rotated  & 2$^{\circ}$ Rotated  & 5$^{\circ}$  Rotated
		
	\end{tabular}
	\caption{Prediction results of rotated fabric images at different angles.}
	\label{fig:rotated_vis}
\end{figure*}

\subsection{Impact on parameters}
\subsubsection{Image patch size}

\begin{figure}
	\centering
	\begin{tabular}{c}
		\subfloat{\includegraphics[width=\columnwidth, scale=1]{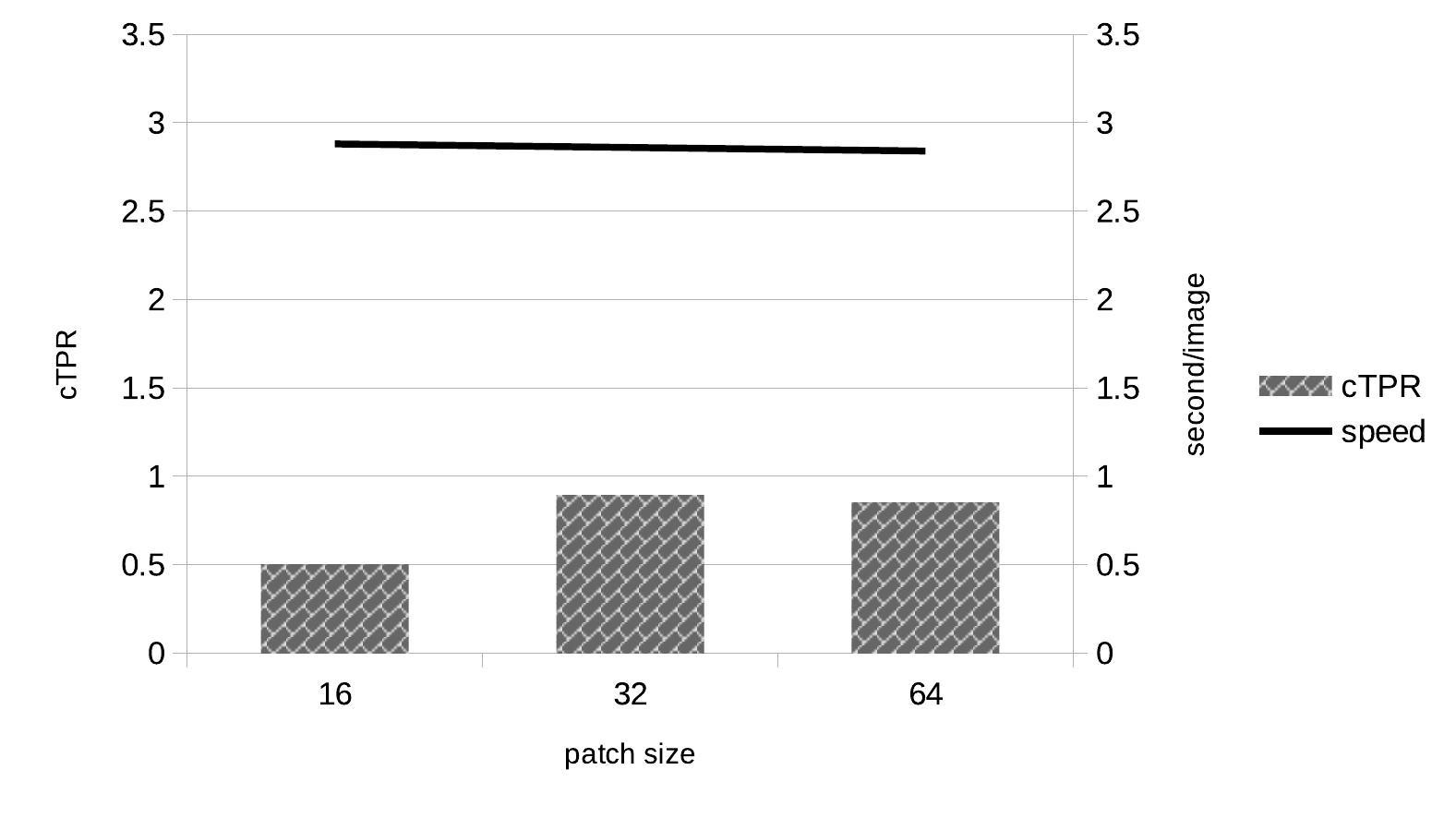}\label{fig:speed}} \\

	\end{tabular}
	\caption{The performance impact of patch size on cTPR and speed.}
	\label{fig:patch}
\end{figure}

Image patch size is important in our design. Bigger image patches contain more information than smaller image patches. To find an appropriate size, we test our model on various image patch sizes to see how it affects cTPR and speed. Fig.~\ref{fig:patch} shows that cTPR increases as the patch size increases then decreases back. We believe that bigger image patch sizes may have more information which simply gets redundant after a certain size. Our autoencoder is able to process redundant information fairly well. Therefore, there is little gain overall unless it is too small. Fig.~\ref{fig:patch} also shows the speed decreases by a very limited amount as the patch size increases. If we take the speed of patch size 32 as a baseline, the difference is only about 0.6\% (existing when cropping an image to patches and compressing by the autoencoder). So we claim that the speed is little affected by the patch size. Also bigger patch sizes make defects visualized larger than actual size. Therefore, we choose 32 as our patch size in the following experiments. However, one should know that the speed can be greatly improved if the number of patches per image decreases.

\subsubsection{Orientations and bandwidths of Gabor filter bank}
\begin{figure*}
	\centering
	\begin{tabular}{c c}
		\subfloat{\includegraphics[width=1\columnwidth,
		scale=1]{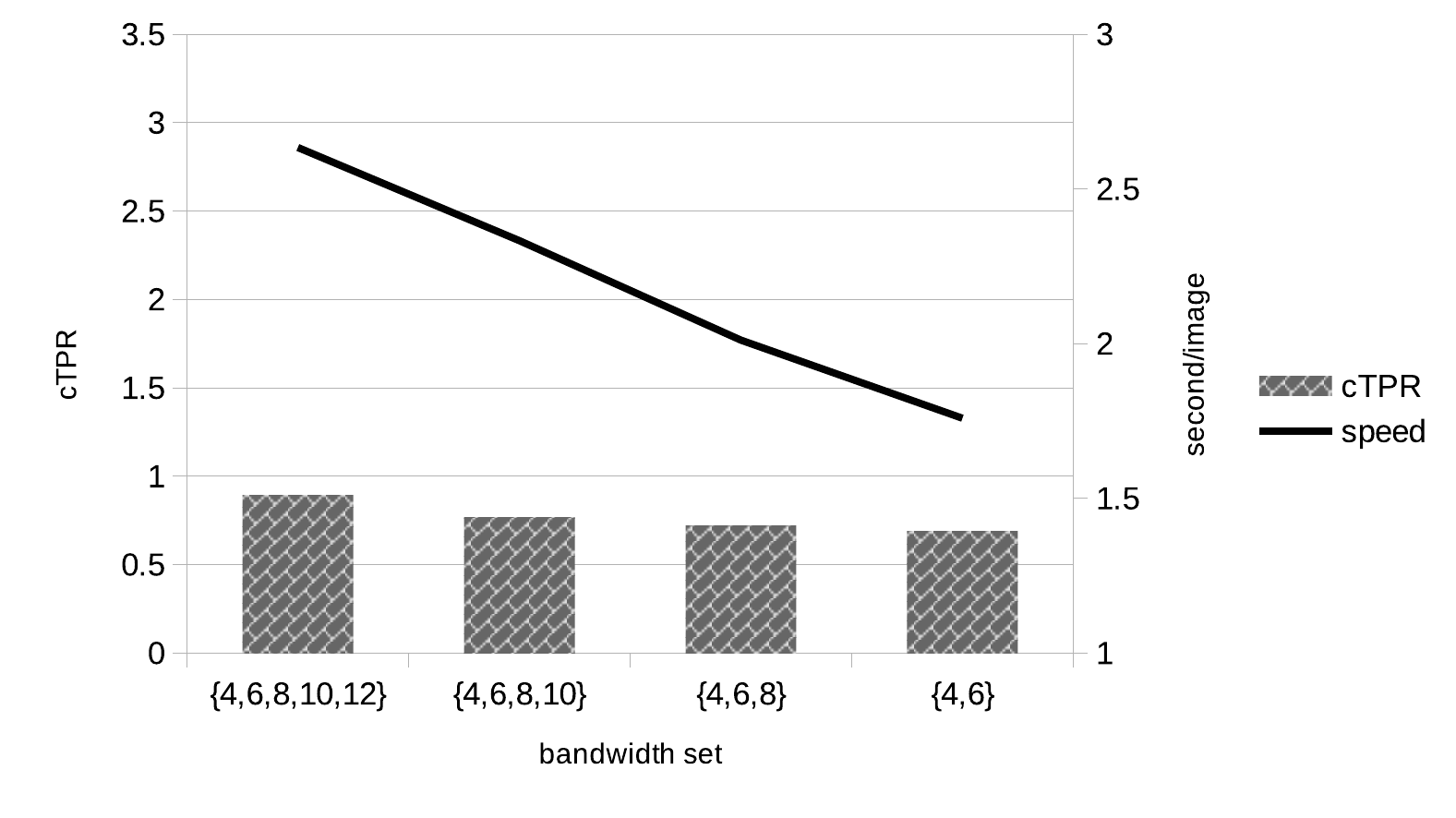}\label{fig:band}} &
		\subfloat{\includegraphics[width=1\columnwidth, scale=1]{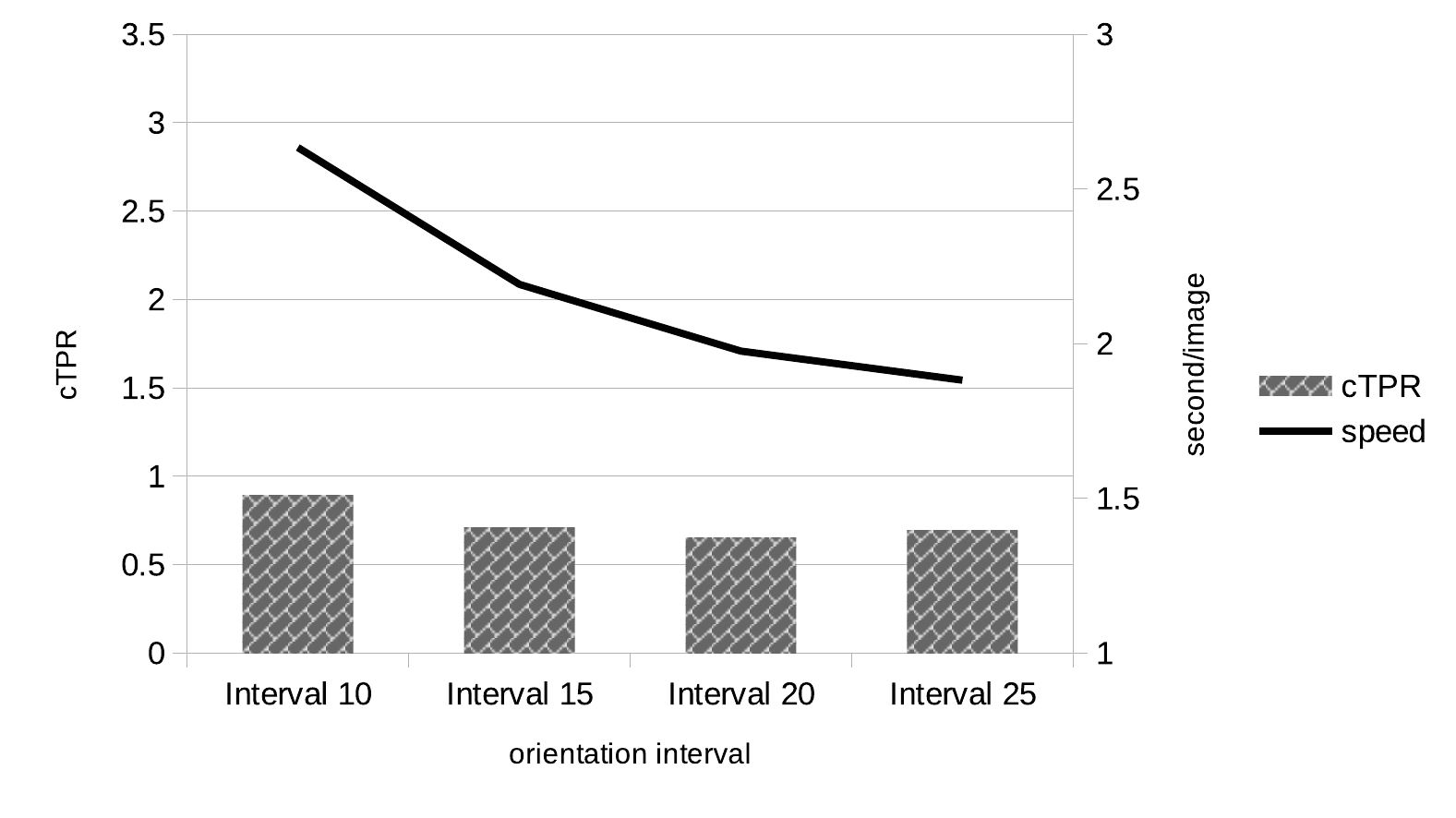}\label{fig:ori}} \\
		
		(a) Bandwidth & (b) Orientation
		
	\end{tabular}
	\caption{The performance impact of parameters of Gabor filter bank on cTPR and speed.}
	\label{fig:gaborbank}
\end{figure*}

One common issue of applying the Gabor filter bank is speed. While more orientations and bandwidths introduce more outputs of the Gabor filter bank, the processing speed increases. In order to make our model applicable and practical, we extensively tested our model on various orientations and bandwidths. In Fig.~\ref{fig:band}, we decrease the number of bandwidths by one so the total number of filters in our Gabor bank is decreased by 18 (e.g., 18 orientations in total). We also decrease the number of orientations by setting an interval from 10 to 25 as in Fig.~\ref{fig:ori}. One can see that when having more bandwidths and orientations, the metric, cTPR, increases but the processing time also increases. Since accuracy has the top priority in the problem, we keep all 90 filters in the experiments.

\subsection{Tests on fabric distortion (rotation)}
\begin{figure}
	\centering
	\begin{tabular}{c}
		\subfloat{\includegraphics[width=\columnwidth, scale=1]{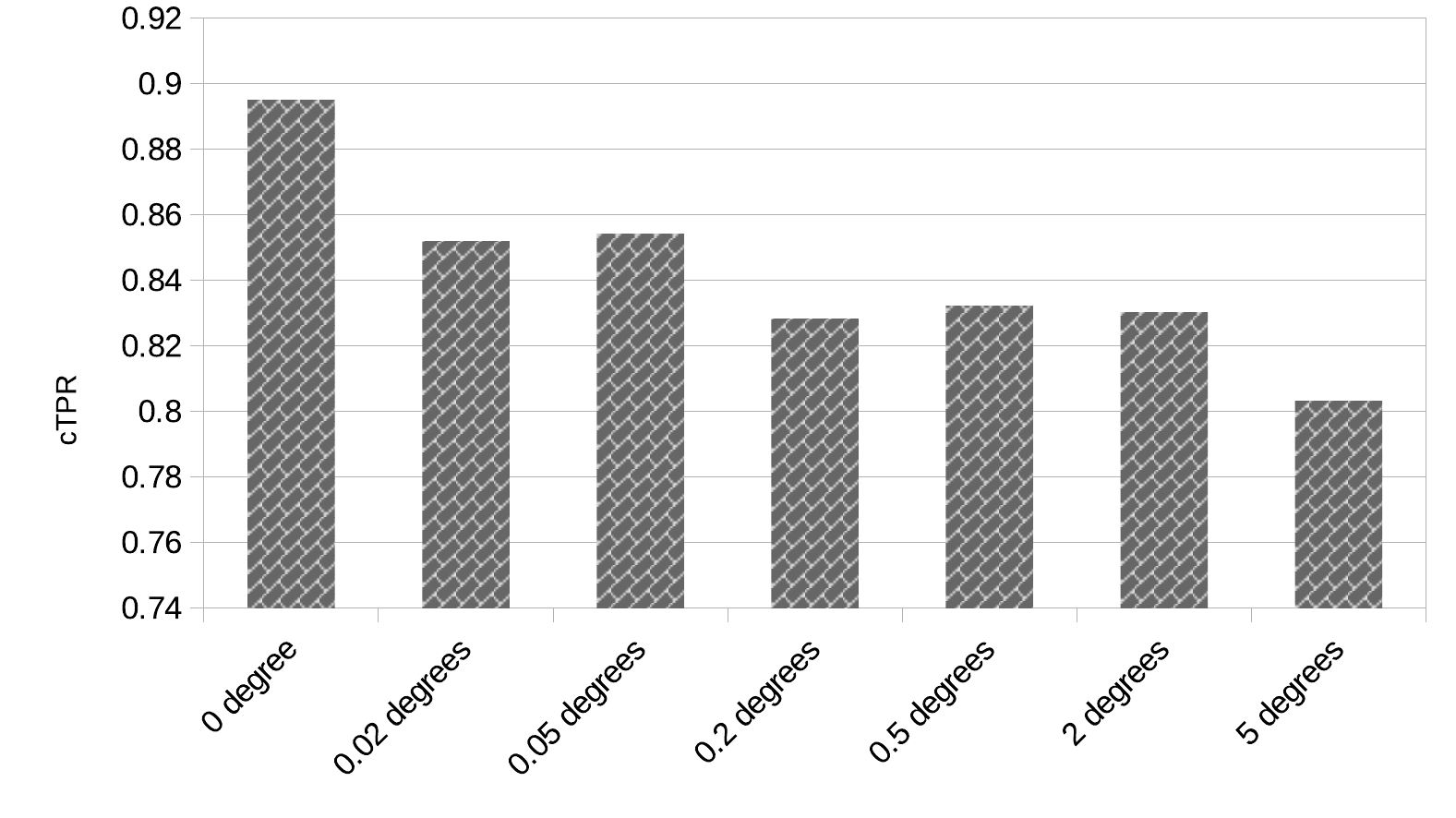}} \\
		
	\end{tabular}
	\caption{cTPR impact of rotated fabric images at different angles. Note that the images are rotated in counter-clock wise.}
	\label{fig:rotated_ctpr}
\end{figure}

Previous research tests proposed models only on fabric images without any distortion. However, in reality, fabrics are easily distorted while traveling from one process to another (In fact, distortion detection/correction machines are placed between some processes in factory). To simulate it, we test our model on manually rotated fabric images at different angles (e.g., 0.02, 0.05, 0.2, 2, and 5 degrees). Fig.~\ref{fig:rotated_ctpr} shows that the cTPR decreases as the rotation angle grows. The performance decreases since when classifying a rotated image patch, the classification boundary is not as clear as the one without rotations. Even with the decrease in performance, we think our model still holds its robustness very well as one can see from the visualized results of Fig.~\ref{fig:rotated_vis}. We believe the robustness of our model not only credits to the Gabor filter bank but also the autoencoder. The Gabor filter bank provides magnitude responses from different orientations and then the autoencoder is able to learn more general features. 


\section{Conclusions\label{sec:conc}}

In this paper, we propose a one-class model with a carefully designed Gabor filter bank, an autoencoder as a general feature learner, and the nearest neighbor density estimator, to solve the fabrics defect detection problem. Our proposed model does not need defective samples, which is a significant practical advantage. The Gabor filter bank contains 90 Gabor filters at numerous scales and orientations, which saves our efforts on optimizing Gabor filters for any specific fabric, and also show the advantage when compared with a ResNet model that was trained on a limited dataset. In order to increase the applicability of our model, we design an autoencoder to better learn general features from the outputs of the Gabor filter bank. The experiments show the proposed autoencoder improves the performance by over 12.9\% when compared against the hand-crafted and PCA feature selection methods. We also extensively test our model at different parameter settings. Moreover, we demonstrate applicability by showing our model can work well on plain, patterned, and rotated fabric images.


\bibliographystyle{IEEEtran}
\bibliography{reference}


\end{document}